\documentclass[10pt,twocolumn,letterpaper]{article}

\usepackage{cvpr}
\usepackage{times}
\usepackage{epsfig}
\usepackage{graphicx}
\usepackage{amsmath}
\usepackage{amssymb}
\usepackage{breqn}
\usepackage{booktabs} 
\usepackage{color}
\usepackage{tabularx}

\definecolor{darkgreen}{RGB}{10,150,10}

\usepackage[pagebackref=true,breaklinks=true,letterpaper=true,colorlinks,bookmarks=false,citecolor=blue,linkcolor=blue]{hyperref}

\cvprfinalcopy % *** Uncomment this line for the final submission

\def\cvprPaperID{719} % *** Enter the CVPR Paper ID here

\ifcvprfinal\fi
\begin{document}

%%%%%%%%% TITLE
\title{Im2Pencil: Controllable Pencil Illustration from Photographs}

%title1: Learning to Sketch with Pencil Strokes
%title2: Learning to Sketch with Multiple Styles
%title3: Learning to Sketch with Different Strokes

\author{Yijun Li$^1$, ~Chen Fang$^2$, ~Aaron Hertzmann$^3$, ~Eli Shechtman$^3$, ~Ming-Hsuan Yang$^{1,4}$\\
$^1$UC Merced~~~~~~$^2$ByteDance AI~~~~~$^3$Adobe Research~~~~$^4$Google Cloud\\
{\tt\small \{yli62,mhyang\}@ucmerced.edu ~~fangchen@bytedance.com~~\{elishe,hertzman\}@adobe.com}
}

\maketitle
%\thispagestyle{empty}
%%%%%%%%% ABSTRACT
\begin{abstract}

We propose a high-quality photo-to-pencil translation method with fine-grained control over the drawing style.
This is a challenging task due to multiple stroke types (e.g., outline and shading), structural complexity of pencil shading (e.g., hatching), and the lack of aligned training data pairs.
To address these challenges, we develop a two-branch model that learns separate filters for generating sketchy outlines and tonal shading from a collection of pencil drawings.
We create training data pairs by extracting clean outlines and tonal illustrations from original pencil drawings using image filtering techniques, and we manually label the drawing styles.
In addition, our model creates different pencil styles (e.g., line sketchiness and shading style) in a user-controllable manner.
Experimental results on different types of pencil drawings show that the proposed algorithm performs favorably against existing methods in terms of quality, diversity and user evaluations.

%This paper proposes a learning-based framework that automatically turns a natural image to a piece of pencil drawing, with sketchy lines and tonal shadings that contain rich pencil strokes.
%
%We formulate this task in an image-to-image translation fashion by learning such a mapping from a collection of pencil drawings.
%
%Our main contribution is to introduce non-photorealistic rendering (NPR) techniques into the translation model so that it is unnecessary to spend laborious human effort in constructing photo-pencil data pairs.
%
%Specifically, we only need to collect the pencil drawing data, employ NPR techniques to simplify them and let the model learn how to restore original pencil drawings by generating right strokes.

\end{abstract}

%%%%%%%%% BODY TEXT
\section{Introduction}

\begin{figure}[t]
\centering
\begin{tabular}{c@{\hspace{0.005\linewidth}}c@{\hspace{0.005\linewidth}}c}

\includegraphics[width = .45\linewidth]{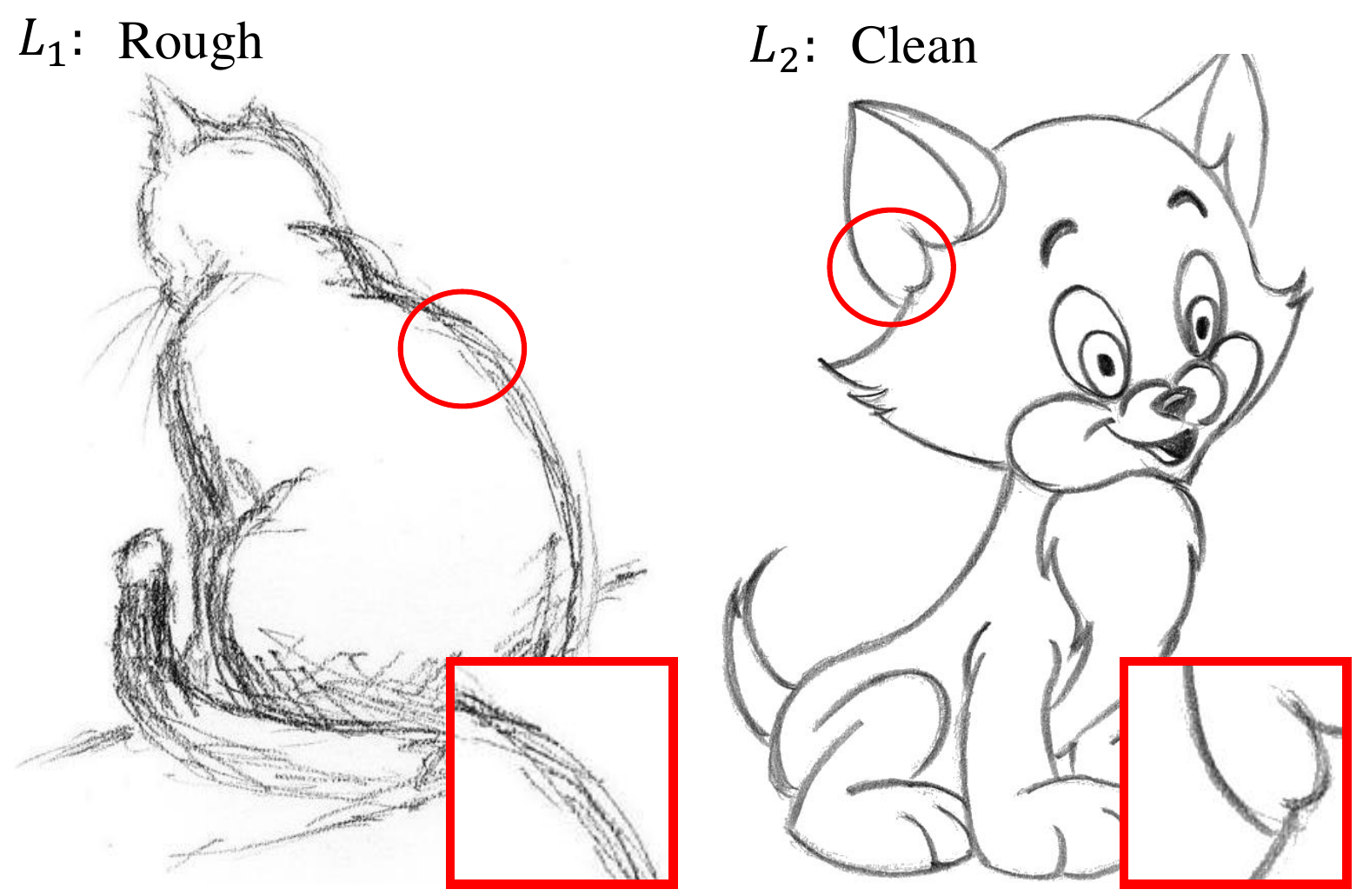} & 

\includegraphics[width = .51\linewidth]{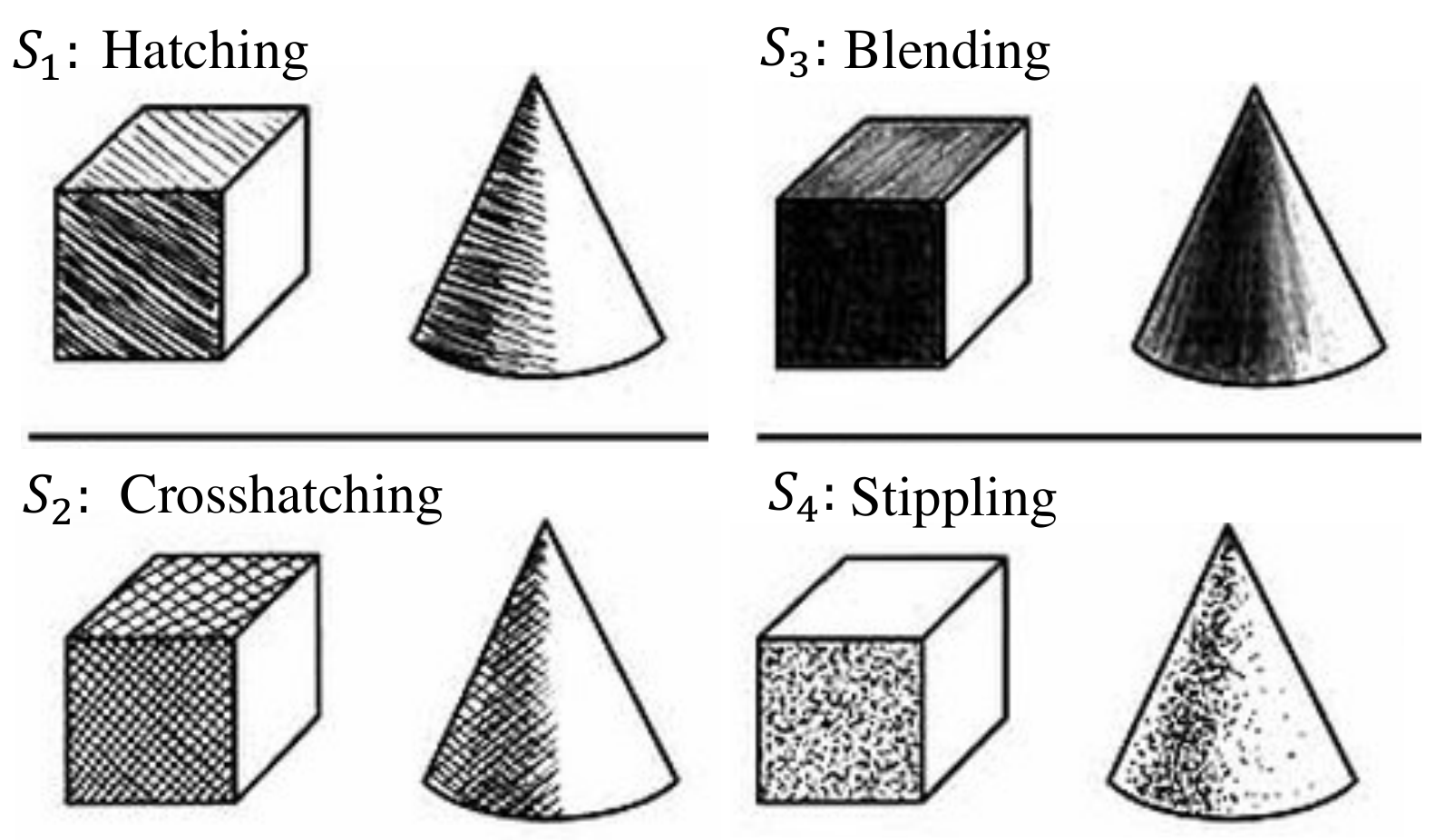} & \\

{(a) Pencil outlines} & {(b) Pencil shading} & \\

\end{tabular}
\vspace{0.05em}
\caption{Examples of real pencil drawings in the outline ($L_1\sim L_2$) and shading ($S_1\sim S_4$) styles that we train on.}
\label{fig:real_example}
\end{figure}

Pencil is a popular drawing medium often used for quick sketching or finely-worked depiction.
%
%In art, a pencil drawing is usually an artistic creation of human perception~\cite{lee2006real,lu-NPAR2012}.
%
%Other usage of pencil sketching includes shape abstraction~\cite{cole2008people}, sketch-based image retrieval~\cite{sangkloy2016sketchy,yu2016sketch} and forensic facial sketch matching~\cite{ouyang2016forgetmenot}.
%
%, such as conveying outlines, and conveying shading and texture through oriented hatching and cross-hatching \cite{guptill,hodges}.
%Such drawings contain rich pencil strokes of different thickness and density that reflect different amounts of shading and tone in the scene.
% 
%
Notably, two main components are the {\em outlines} that define region boundaries, and {\em shading} that reflects differences in the amount of light falling on a region as well as its intensity or tone and even texture.
Each of these may be applied in various different styles. 
For example, pencil outlines may be more or less ``sketchy'' (Figure~\ref{fig:real_example}(a)).
Shading may be also more or less sketchy and use different types of hatching strategies (Figure~\ref{fig:real_example}(b)).
Hence, we seek to accurately reproduce these drawing styles and allow users to select based on personal preferences.

We split the task into generating the outlines and shading separately, and express each as an image-to-image translation problem, learning the mapping from a collection of pencil drawings. 
Unfortunately, gathering paired training data for any artistic stylization task is challenging, due to the cost, as well as the spatial distortions in drawings \cite{cole2008people}.
%
%MH: check this sentence... obtain very little data?
%As such, existing studies operate on highly-restricted classes of styles and obtain very little data \cite{cole2008people}.
%
%As such, existing studies operate on highly-restricted classes of styles with few training examples .
%
%However, it is difficult to find a large set of photo-drawing pairs to learn the mapping.
%
%Even when an artist is asked to draw with a photo as reference, the photo is often viewed at a distance and the drawing is unlikely to be accurately aligned with the photo.
%
%An alternative is using the seminal work CycleGAN~\cite{CycleGAN-ICCV17} which supports learning with unpaired data. 
%
%However, modeling the distribution of pencil drawing domain is still hard for adversarial networks without pairing correspondence.
%
To avoid the burden of gathering ground-truth paired data, we instead propose to create data pairs. 
In particular, we filter each pencil drawing with procedures that extract outlines, tone, and edges. 
This produces two sets of paired data that can be used for our two subtasks, learning the outlines and shading drawing. 
These filters generate the same abstractions (outlines, tone, and edges) when applied to input photographs.  
Hence, at test-time, we filter an input photograph and then apply the trained model to produce pencil illustrations in a user-selected style.

\begin{figure}[t]
\centering
\begin{tabular}{c@{\hspace{0.005\linewidth}}c@{\hspace{0.005\linewidth}}c@{\hspace{0.005\linewidth}}c@{\hspace{0.005\linewidth}}c@{\hspace{0.005\linewidth}}c}

\includegraphics[width = .48\linewidth]{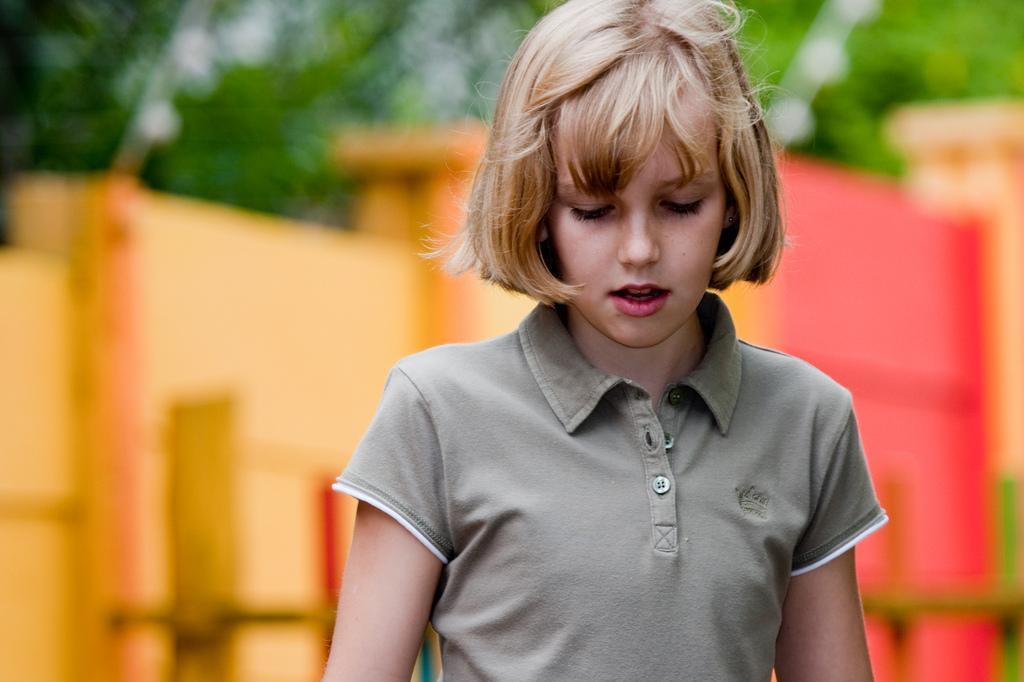} & 
\includegraphics[width = .48\linewidth]{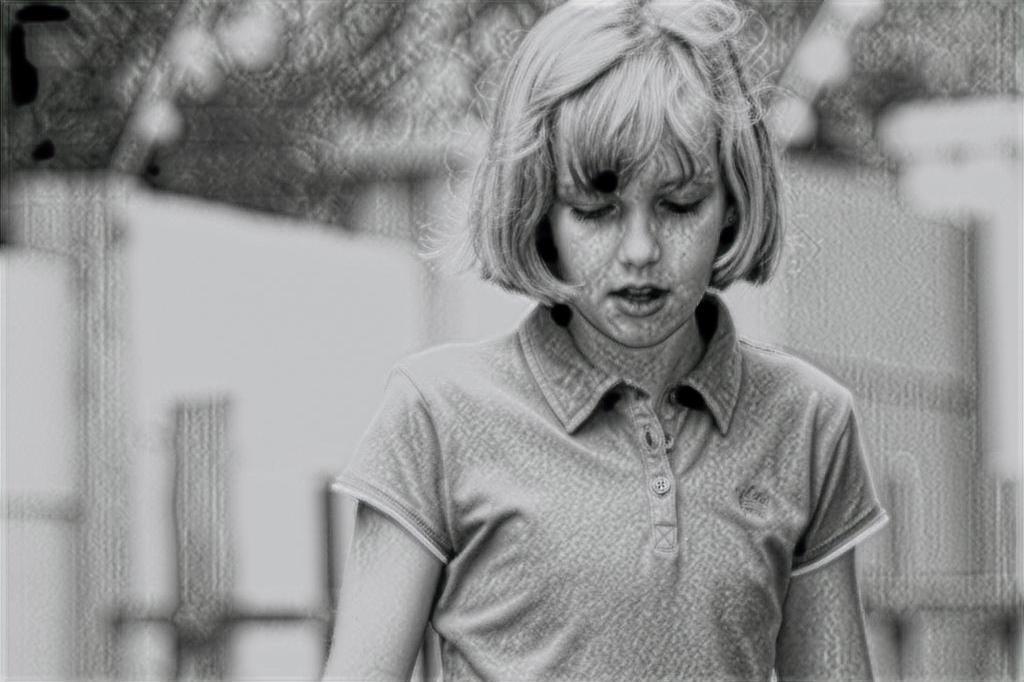} & \\

{(a) Input} &  {(b) CycleGAN~\cite{CycleGAN-ICCV17}} &\\

\includegraphics[width = .48\linewidth]{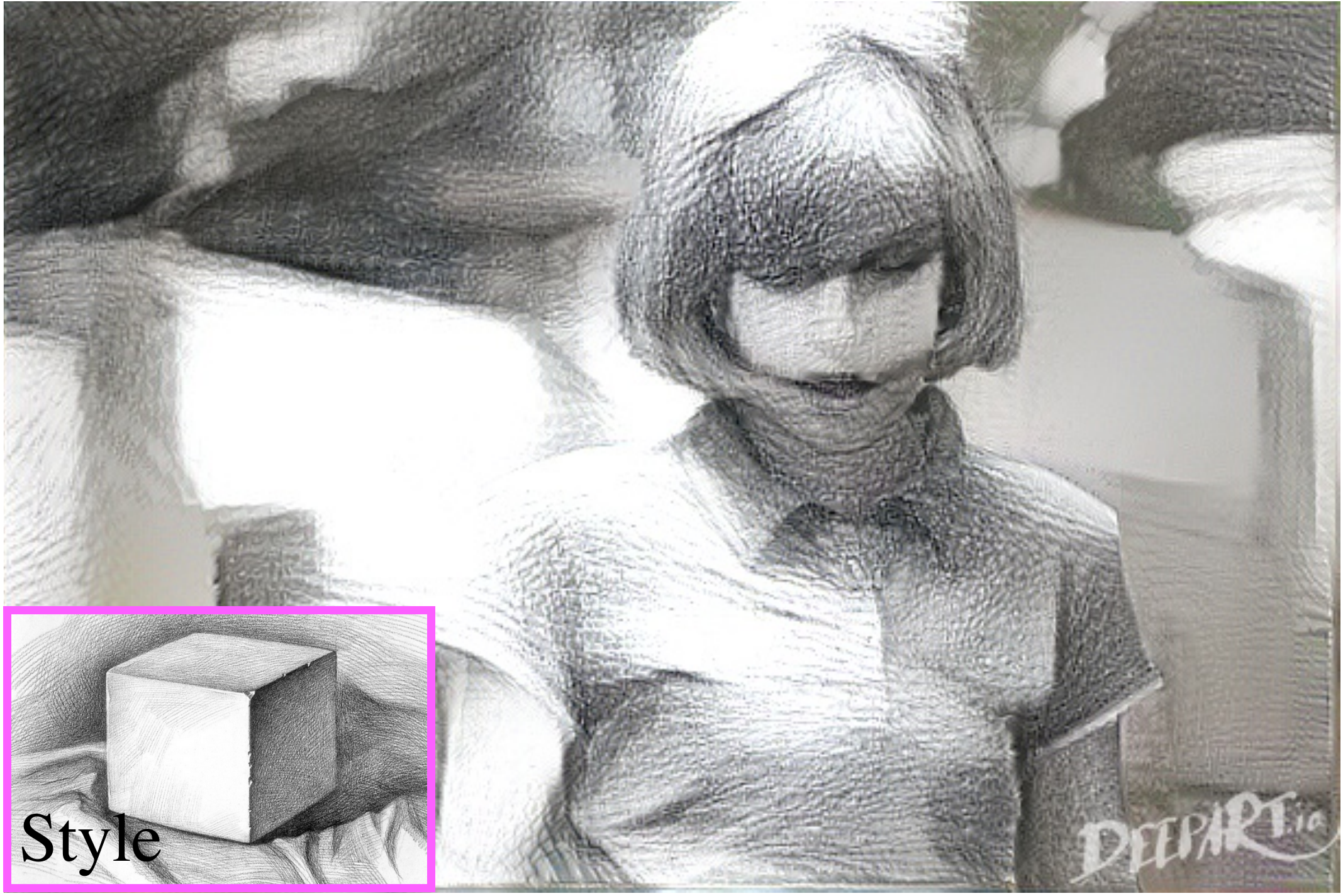} & 
\includegraphics[width = .48\linewidth]{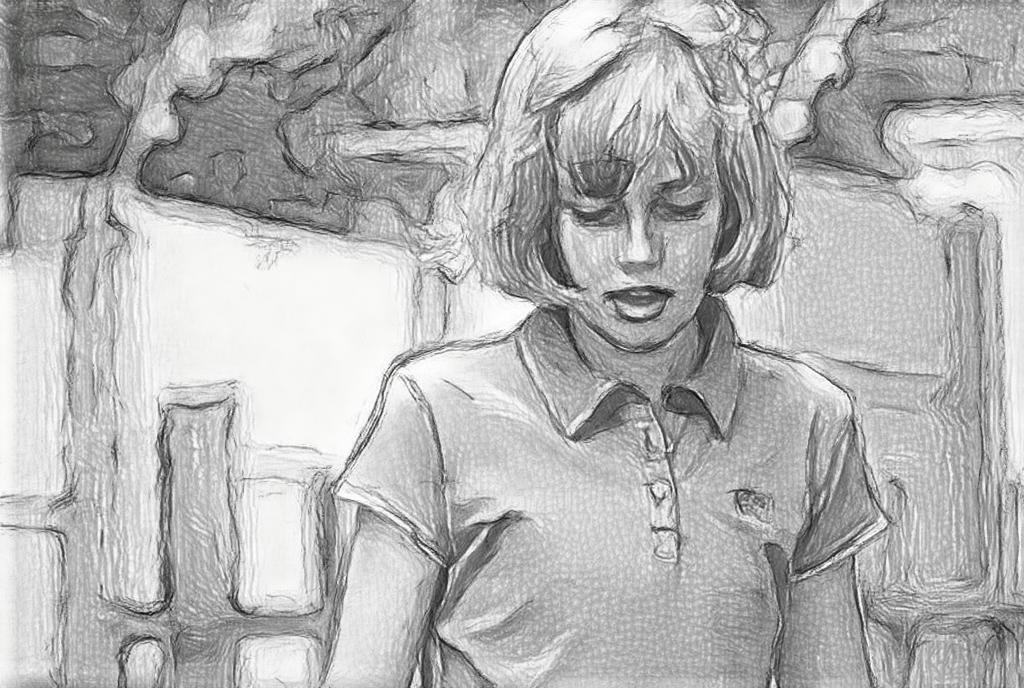} & 
\\

{(c) Gatys \etal~\cite{GatysTransfer-CVPR2016}} & {(d) Ours: $L_{1}+S_{2}$}&\\

\includegraphics[width = .48\linewidth]{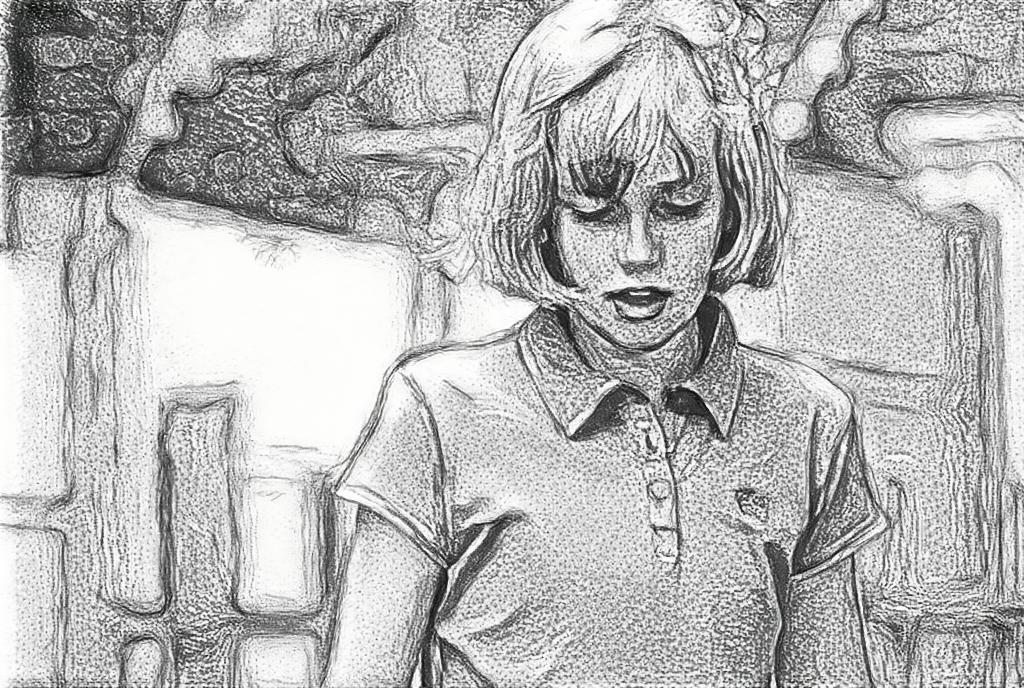} & 
\includegraphics[width = .48\linewidth]{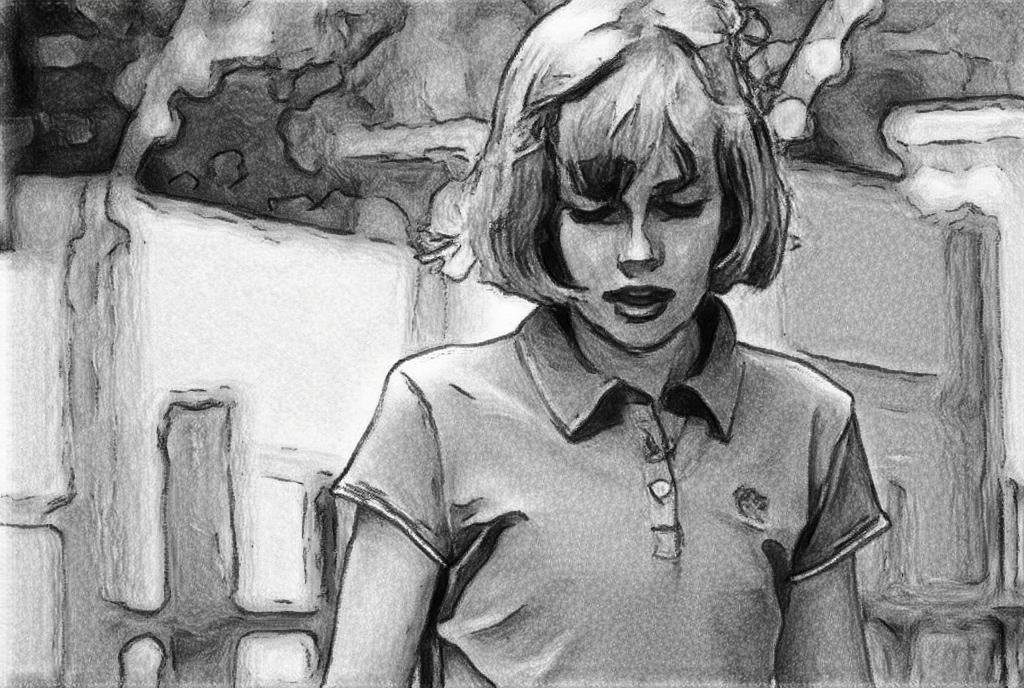} & 
\\

{(e) Ours: $L_{1}+S_{4}$} & {(f) Ours: $L_{2}+S_{3}$}&\\

\end{tabular}
\vspace{0.05em}
\caption{Synthesis results of our algorithm in  different combinations of outline and shading styles, compared with existing methods (\textbf{zoom in} for fine pencil strokes).
See Section~\ref{method_compared} for experimental details.
%\aaron{we should show/explain what training data was used for each baseline?}
%\yijun{I show how to obtain results (e.g., training data) of each baseline in Sec 4.1.}
}
\label{fig:teaser}
\end{figure}

We achieve control over different pencil styles (e.g., sketchiness and shading) by training the model with several distinct styles, where the style label is provided as an input selection unit. 
%
%We provide a selection unit to guide the generation towards the desired style within a %single network. 
%
Moreover, the filtering modules contain additional controllable parameters to generate different abstracted inputs, which are then mapped to pencil drawings with different characteristics.
%
%Both the NPR modules and translation model are flexible to provide users with control signals to switch between different types of pencil drawings. 
%
On the other hand, we find that image translation architectures can often produce undesirable hatching patterns. 
We describe these issues and show how careful design of network architectures can address these problems.

A long-term goal of this research direction is to provide more fine-grained control to neural-network stylization algorithms. 
Existing learning-based methods do not provide much control over style except by changing the training input. 
In contrast, classical procedural stylization algorithms can provide  many different styles (e.g., \cite{benard2018line,Rosin:2013}) but without the same quality and generality that can come from learning from examples.
Our method allows fine-grained stylistic control as we focus on separating outline and shading style, and learning several stylistic options for each. 
%
%MH: is it necessary to add this sentence?
%In the future, many of the ideas here may be useful in adding fine-grained style control to more general classes of artistic style.

The main contributions of this work are summarized as follows:
\begin{itemize}
\item We propose a two-branch framework that learns one model for generating sketchy outlines and one for tonal shading, from a pencil drawing dataset.
\item We show how to use abstraction procedures to generate paired training data for learning to draw with pencils. 
\item We demonstrate the ability to synthesize images in various different pencil drawing styles within a single framework.
\item We present an architecture that captures hatching texture well in the shading drawing, unlike existing baselines.
\end{itemize}

\section{Related Work}

%MH: no need for extra space because of section title
%\vspace{.5em}
\noindent\textbf{Procedural line drawing.}
There is a rich literature on procedural (non-learning) stylization in Non-Photorealistic Rendering (NPR) \cite{Rosin:2013, benard2018line}.
Early work focuses on interactive pen-and-ink drawing and hatching of 2D inputs \cite{Salisbury:1994:IPI,Salisbury:1997:OTI} and 3D models \cite{HatchLearning-2012, Winkenbach:1994,Winkenbach:1996}.
Pencil drawing is similar to pen-and-ink drawing, but it has more degrees-of-freedom since individual pencil strokes may have varying tone, width, and texture.
For 2D images, several procedural image stylization approaches have simulated pencil drawings \cite{lee2006real,mao2001automatic}.
These methods use hand-crafted algorithms and features for outlines and a pre-defined set of pencil texture examples for  shading.
%
%Often times, those predefined texture examples have to be uniform, so that their spatial arrangements will be less visible in outputs. \aaron{not sure I understand that}
While procedural approaches can be fast and interpretable, accurately capturing a wide range of illustration styles with purely procedural methods is still challenging.

%However, they only handle portrait sketches instead of general scenes.
%
%Most recently, Simo-Serra \etal~\cite{SimoSerra-TOG2016,SimoSerra-TOG2018} propose a generative model on learning how to simplify the sketches for animation, which is the inverse task of ours.
%

%HERE
\vspace{.5em}
\noindent\textbf{Image-to-image translation.}~
Due to the difficulty in authoring stylization algorithms,  numerous approaches have been proposed to learn them from examples.
%
%MH: why sometimes you use analogies and sometimes analogy? (i.e., Aaron's and MSRA's)?
The Image Analogy approach uses texture synthesis applied to a single training pair \cite{Hertzmann-2001-IA}, requiring strict alignment between the input photograph and the output drawing or painting.
For line drawing, the paired input images are created manually by applying blurring and sharpening operators separately to each input drawing.  
This method performs well only for restricted classes of drawings, e.g., when most hatching strokes have a consistent orientation.
Although neural image-translation methods have been recently developed~\cite{Pix2Pix-CVPR2017,BicycleGAN-NIPS2017,pix2pixHD},  none of these has been demonstrated for stylization, due to the difficulty of gathering aligned, paired training data. 
Chen \etal~\cite{chen2017fast} do learn stylization, but trained on the results generated by an existing procedural pencil rendering method~\cite{lu-NPAR2012}. % on a collection of photos to create a photo-pencil paired dataset and then train a feed-forward network for fast processing. 
%
%MH: watch your tone
%However, the results can be no better than those of \cite{lu-NPAR2012}. 
%MH: check this sentence
%However, the styles and shading of rendered drawings are limited. 
However, the rendered drawings exhibit limited sketching and shading styles. 
More recently, several methods have been developed for learning mappings from unpaired data~\cite{CycleGAN-ICCV17,DiscoGAN,UNIT,MUNIT-ECCV2018,DRIT-ECCV2018} with two conditional adversarial losses and cycle-consistency regularization.
%
%MH: again, watch your tone
%But as shown in Figure~\ref{fig:teaser}(b), these methods do a poor job at capturing hatching texture. Deep Image Analogy \cite{MSRA-2017-visual} does not require paired data, but only applies when the exemplar and target photo have extremely similar content.
However, as shown in Figure~\ref{fig:teaser}(b), these methods do not perform well at capturing hatching texture. 
Although the Deep Image Analogy \cite{MSRA-2017-visual} method does not require paired data, it requires data where the exemplar and target photo have very similar content.
Several learning-based algorithms have been developed solely for facial portraiture~\cite{wang2009face,song2014real,Berger-2013,Fiser:2016}. 

%modeling the distribution of pencil drawing domain is still hard for adversarial networks without pairing correspondence.

\begin{figure*}[t]
\centering
\begin{tabular}{c@{\hspace{0.005\linewidth}}c}

\includegraphics[width = .96\linewidth]{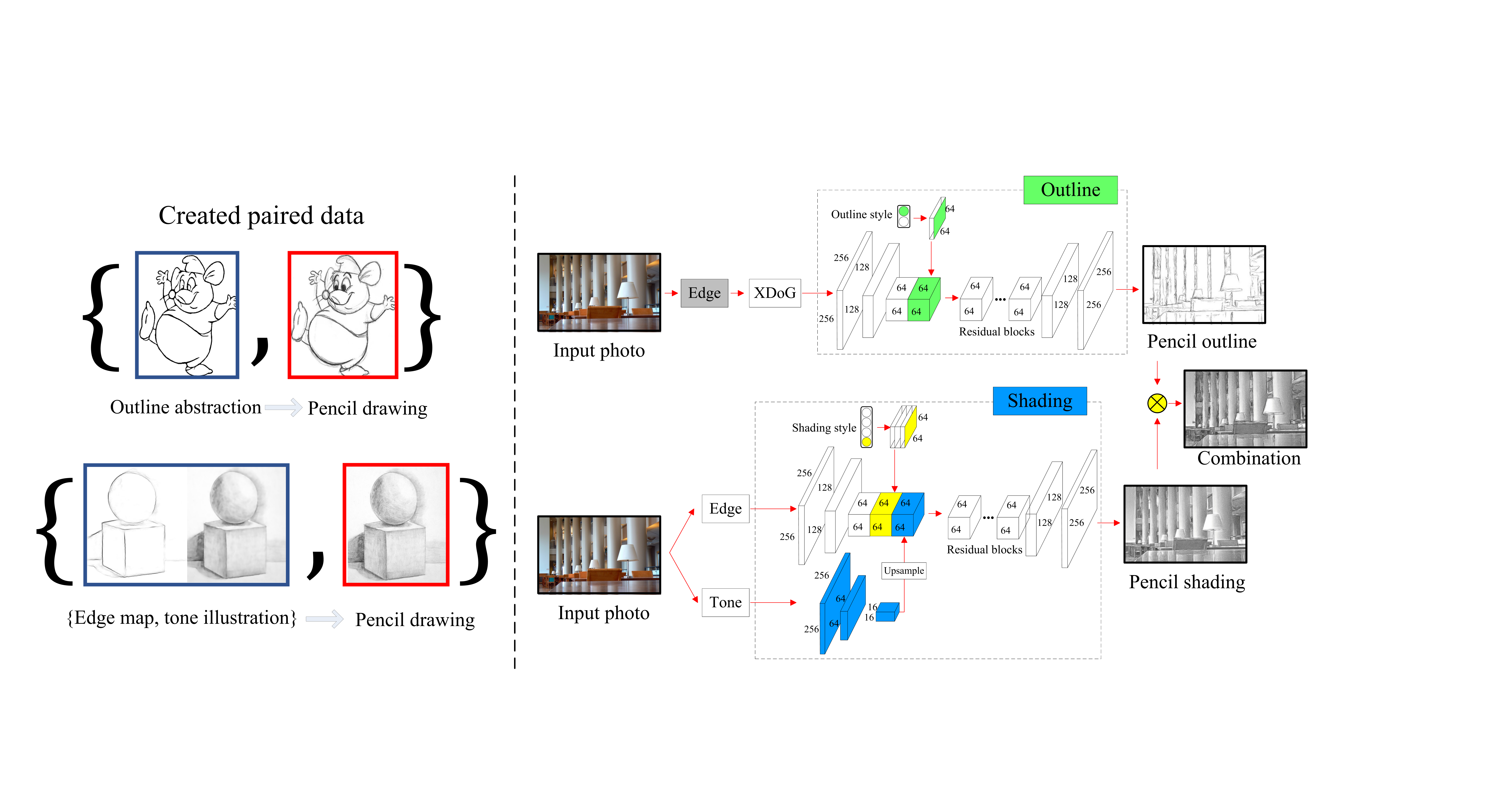} & \\

\end{tabular}
\vspace{0.05em}
\caption{\textbf{Pipeline of the proposed algorithm}. 
Left: the created paired training data generated by using an abstraction procedure on pencil drawings for training. 
Right: the testing phase (including network details).
Two branches will output an outline and shading drawing result respectively, which can be combined together through pixel-wise multiplication as the third option of pencil drawing result.
The edge module in gray in the outline branch (top) is a boundary detector~\cite{DollarICCV13edges}, which is optional at test-time. For highly-textured photos, it is suggested to use this module to detect boundaries only. See Section~\ref{technique} for technical details.}
\label{fig:framework}
\end{figure*}

\vspace{.5em}
\noindent\textbf{Neural style transfer.}~A third approach is to transfer deep texture statistics of a style exemplar, which does not employ paired training data.
Since Gatys \etal~\cite{GatysTransfer-CVPR2016} proposed  
an algorithm for artistic stylization based on matching the
correlations (Gram matrix) between deep features, numerous methods have been developed for improvements in different aspects~\cite{Survey-NST}, e.g., efficiency~\cite{Perceptual-ECCV2016,Texturenet-ICML2016}, generality~\cite{chen2017fast,Huang-2017-arbitrary,WCT-NIPS-2017,chen2017stylebank,Me-2017-diversified}, quality~\cite{MrfTransfer-CVPR2016,MSRA-2017-visual,jing2018stroke}, diversity~\cite{Me-2017-diversified}, high-resolution~\cite{HD2018style}, and photorealism~\cite{DPST-CVPR2017,PhotoWCT-ECCV2018}. 
%
%MH: no need to cite this in a separate sentence. Just add it to one of the above discussion. 
%A comprehensive survey can be found in~\cite{Survey-NST}.
%
However, these methods do not perform well for pencil drawing.
The rendered results (Figure~\ref{fig:teaser}(c)) in the pencil style only capture the overall gray tones, but without capturing distinctive hatching or outline styles well.

\vspace{.5em}
\noindent\textbf{Style control.}~
Procedural methods often provide fine-grained style control, e.g., \cite{Grabli:2010}, but at the cost of considerable effort and 
difficulty in mastering certain styles. 
%Since the general image stylization task does not have ground truth (GT) outputs, to equip the model with controls is always a highly demanded feature in order to let users switch between diverse outputs.
%
%Many procedural stylization work~\cite{odonovan-2012, Gatys2016-control,zhang2017real,Me-2017-diversified} provide flexible user controls in the NPR and neural style transfer literature, which are usually designed in the interactive fashion.
%
Image-to-image translation~\cite{MUNIT-ECCV2018,DRIT-ECCV2018} and neural style transfer methods provide only high-level control, e.g., by selecting training inputs in a different style, interpolating between unrelated styles~\cite{Dumoulin:2017,Gatys2016-control},
or selecting among high-level transfer parameters~\cite{Gatys2016-control}.  
In this work, we focus on developing a method with fine-grained style control that allows subtle adjustments to pencil drawing.

%We improve the translation model by embedding NPR modules and training with multiple styles in one single network to handle different styles in the pencil drawing.

\section{Stylization Approach}
\label{technique}
%We now describe the technical details of our proposed model. 
Our approach is based on the observation that pencil drawings can be separated into two components: outlines, and shading. 
The outlines delineate object boundaries and other boundaries in the scene, and shading or tone uses tonal techniques such as hatching to depict reflected lighting, texture, and materials. 
Hence, our method includes a separate outline branch and a shading branch. 
These two models are trained separately but can be combined at test-time to generate different combinations of illustration styles.
Figure~\ref{fig:framework} shows the main modules of the proposed method.

For each network branch, paired training data is unavailable, and thus we need to create the input-output pairs from line drawings directly. 
We generate training data by using an abstraction procedure on pencil drawings, where the abstraction estimates outlines or edges and tones.
These filters are designed to produce similar abstractions from line drawings as from photographs. 
Hence, at test-time, the same abstraction filters can be applied to an input photograph, to produce an input in the same domain as the training inputs.
%\aaron{I wrote this paragraph, but it's so abstract, not sure it's helpful...}

\subsection{Outline branch}
The goal of the outline branch is to produce pencil-like outlines from photos.
Since there is no paired training data for this task, we use an outline extraction algorithm, both to process the training data and test images at run-time.

%turn clean outlines in photos into sketchy outlines drew by pencils.
%
%Since there exist no photo-pencil outline pairs available for learning such a mapping, we propose to simplify both drawings and photos by abstracting their clean outlines.
%
%The abstracted outlines of drawings will be paired with original drawings to constitute a paired dataset, which facilitates the model learning on how to restore original drawings by generating realistic pencil strokes.
%
%During testing, our system also applies an abstraction filter to the input photo. % use directly feed the abstracted photo outlines into the learned model to obtain pencil outline results.
%
%We choose the XDoG~\cite{XDoG2011,XDoG2012} filter to both filtering steps. %perform the simplification on outlines because it is able to abstract both sketch outlines in pencil drawings and  clean edges in photos.
%\aaron{the term "simplify" here is vague.}

\vspace{.5em}
\noindent\textbf{Outline extraction.}
We use the Extended Difference-of-Gaussians (XDoG) filter \cite{XDoG2012}, which performs well whether the input photo is a pencil drawing or a photograph.
%XDoG is a image filter that can be used to stylize or abstract images, depending on its parameter settings. %image stylization algorithm that can produce drawing-like images, with parameters to control its edge detection sensitivity. 
%
%Strictly speaking, it is not an edge detector but has the decent property of rendering aesthetically pleasing edge lines without post-processing. 
%
%For example, the XDoG could tolerate the line thickness, rendering a thick line as a single line still with proper parameters. \aaron{huh?}
%
The XDoG method takes an input image $I$, and convolves it with two separate Gaussian filters, with standard deviations $\sigma$ and $k\cdot \sigma$. 
A sigmoidal function is then applied to the difference of these two images:
\begin{equation}\label{DoG}
D(I;\sigma,k,\tau) = G_{\sigma}(I) - \tau \cdot G_{k\cdot \sigma}(I),
\end{equation}
\vspace{-.5em}
\begin{equation}\label{XDoG}
E_{X}(D,\epsilon,\varphi) \rm{=} \left\{
\begin{array}{ll}
1, ~\textrm{if}~~D \geq \epsilon\\
1 + \tanh(\varphi \cdot (D - \epsilon)), \textrm{otherwise}
\end{array}
\right.
\end{equation}
The behavior of the filter is determined by five parameters: $\{\sigma,k,\tau,\epsilon,\varphi\}$ for flexible control over detected edges. 
%
%\aaron{give all the parameters here?} \eli{I think we fix $\{k,\tau,\epsilon,\varphi\}$ and then vary $\sigma$ to get particular effects, right? This should be clarified.}
%
At test-time, users can adjust any parameters to control the line thickness and sketchiness as shown in Figure~\ref{fig:edge_result} and~\ref{fig:edge_result_fish}.
%
%Details of those parameters and applications of XDoG can be found in~\cite{XDoG2012}.

\begin{figure}[t]
\centering
\begin{tabular}{c@{\hspace{0.005\linewidth}}c@{\hspace{0.005\linewidth}}c@{\hspace{0.005\linewidth}}c@{\hspace{0.005\linewidth}}c}

\includegraphics[width = .24\linewidth]{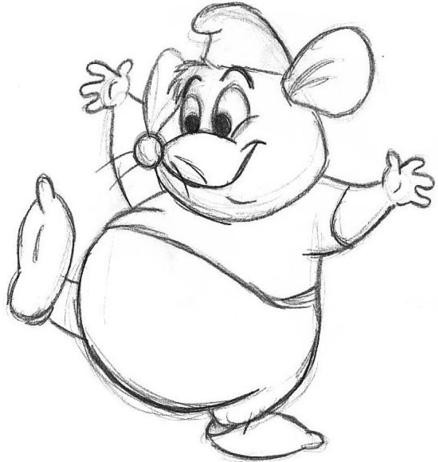} &

\includegraphics[width = .24\linewidth]{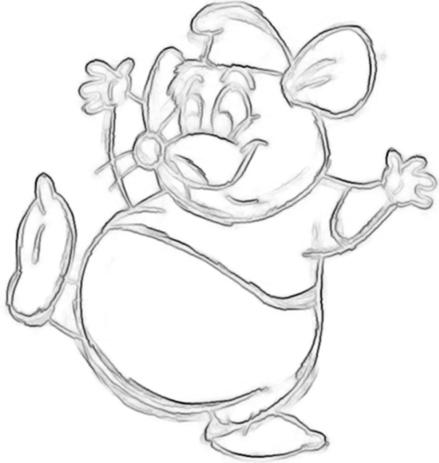} & 
\includegraphics[width = .24\linewidth]{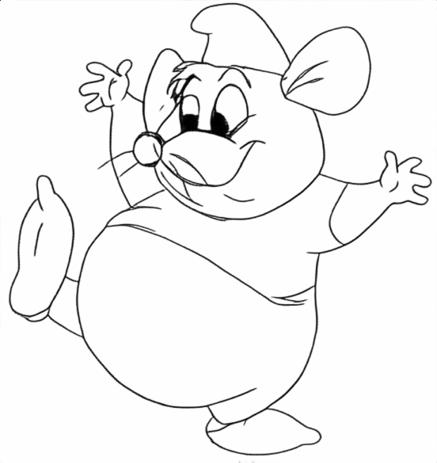} & 
\includegraphics[width = .24\linewidth]{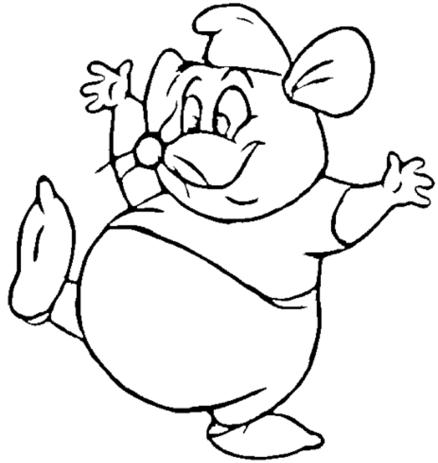} & \\

\includegraphics[width = .24\linewidth]{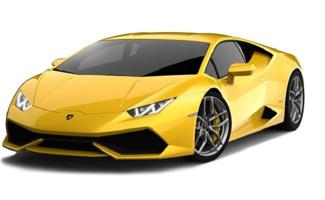} &

\includegraphics[width = .24\linewidth]{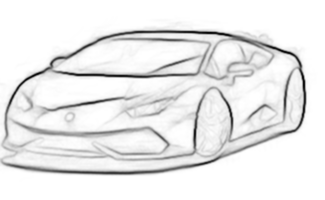} &
\includegraphics[width = .24\linewidth]{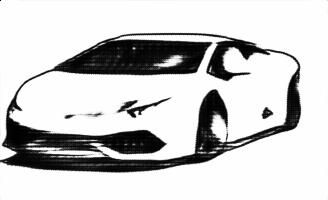} &
\includegraphics[width = .24\linewidth]{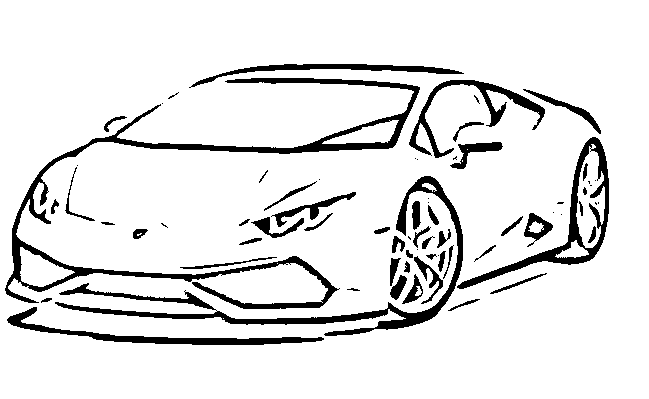} & \\

{(a) Input} & {(b)~\cite{DollarICCV13edges}} & {(c)~\cite{SimoSerra-TOG2016}} & {(d) XDoG~\cite{XDoG2012}} &\\

\end{tabular}
\vspace{0.05em}
\caption{Comparisons of different outline abstraction results on pencil drawings (top) and photos (bottom).}
\label{fig:line_simplify_compare}
\end{figure}

To demonstrate the effectiveness of XDoG, we compare it with two alternative approaches for outline abstraction. 
The first one is a boundary detector based on structured random forests~\cite{DollarICCV13edges}.
%
%MH: try not to say "fail"
%The result in Figure~\ref{fig:line_simplify_compare}(b) shows that although it detects outlines in photos well, it generally fails to handle the line thickness in pencil drawings, resulting in the two sides of lines detected.
%
The results in Figure~\ref{fig:line_simplify_compare}(b) show that although it detects outlines in photos well, it generally does not handle thick lines in pencil drawings well and generates two strokes. 
The second one is a method designed specifically for sketch simplification~\cite{SimoSerra-TOG2016}.
As shown on the top of Figure~\ref{fig:line_simplify_compare}(c), it obtains  simplification results on main contours but does not handle smooth non-outline regions well (e.g., eyes).
%
%MH: told you to use perform well not work well...
More importantly, this sketch simplification method does not perform well on abstracting outlines of photo inputs (see the bottom row of Figure~\ref{fig:line_simplify_compare}(c)).
In contrast, the XDoG filter handles line thickness and smooth non-outline regions well (Figure~\ref{fig:line_simplify_compare}(d)). 
%
%At test-time, if the input photograph contains heavy textures, where texture edges are also likely to be detected by XDoG, it is suggested to first use the edge detector~\cite{DollarICCV13edges} to highlight the main outline and then apply XDoG to control the outline style.
%
For some highly-textured photos at test-time, the XDoG may produce far too many edges (the second row of Figure \ref{fig:shading_simplify}(b)). In these cases, we first use the boundary detector \cite{DollarICCV13edges} to extract their contours, which are then filtered by the XDoG. 

\vspace{.5em}
\noindent\textbf{Paired training data.}~In order to generate paired training data, we first gather a set of pencil \textit{outline} drawings with very little shading, %\aaron{are you saying that (a) nearly all of them have no shading? or (b) all of them have very little shading?}
from online websites (e.g., Pinterest).
We annotate each drawing with one of two outline style labels: ``rough'' or ``clean'' (Figure~\ref{fig:real_example}(a)).
The data is collected by searching the outline style as the main query on web. We collected 30 images for each style.
We use a 2-bit one-hot vector as a selection unit to represent these two styles, which serves as another network input to guide the generation towards the selected style.
Then for each drawing, we manually select a set of XDoG parameters that produce good outlines.
For example, for a sketchier input, we use a bigger $\sigma$ to produce one single thick line to cover all sketchy lines along the same boundary.
% 
%\aaron{explain style vector. explain training data source and generation.}
%
We crop patches of size 256$\times$256 on the created paired data and conduct various augmentations (e.g., rotation, shift), resulting in about 1200 training pairs.

%\aaron{how many training pairs are used? can you be more specific about how they're gathered?}

\vspace{.5em}
\noindent\textbf{Translation model.}~As shown on the top row of Figure~\ref{fig:framework}(right), the translation model is designed as an auto-encoder with a few residual-based convolutional blocks in-between.
The selection unit is first mapped from a 2-bit vector to a 2-channel map, which is then encoded as feature maps (through convolutions) and concatenated with the features of the outline input.
Before the translation module, an XDoG filter is used to extract outlines from photos. %
This module is \textit{not} included during training, and the rest of the model is trained on the outline/drawing pairs described above. For highly-textured images, the boundary (edge) detector may optionally be used prior to XDoG as well.
As described in Section~\ref{use_control}, adjusting parameters to this XDoG filter can be used to vary outline drawing styles, such as line thickness and sketchiness.
%
%Unlike Pix2Pix~\cite{Pix2Pix-CVPR2017} which translates one style only, our model enables multi-style translation with controls and supports meaningful interpolation between styles. 

%Moreover, by controlling parameters in (\ref{DoG}) and (\ref{XDoG}), we can realize different levels of abstraction, which will be later transformed to pencil outlines in different levels of sketchiness.

\subsection{Shading branch}

The goal of our shading branch is to generate textures in non-outline regions according to the tonal values of the input.
%
%MH: colloquial
%As there is still no paired training data to use, so we again apply an abstraction procedure to both produce the training data, and to preprocess inputs at test-time. The goal of the abstraction is to extract edges and tones, and the model learns to apply pencil shading according to these maps.
As no paired data is available for learning the shading branch network, we apply an abstraction procedure to generate the training data and preprocess inputs at test-time. 
The abstraction aims at extracting edges and tones, and removing detailed textures. Then the model learns to apply pencil shading according to these maps.

\begin{figure}[t]
\centering
\begin{tabular}{c@{\hspace{0.005\linewidth}}c@{\hspace{0.005\linewidth}}c@{\hspace{0.005\linewidth}}c@{\hspace{0.005\linewidth}}c}

\includegraphics[width = .24\linewidth]{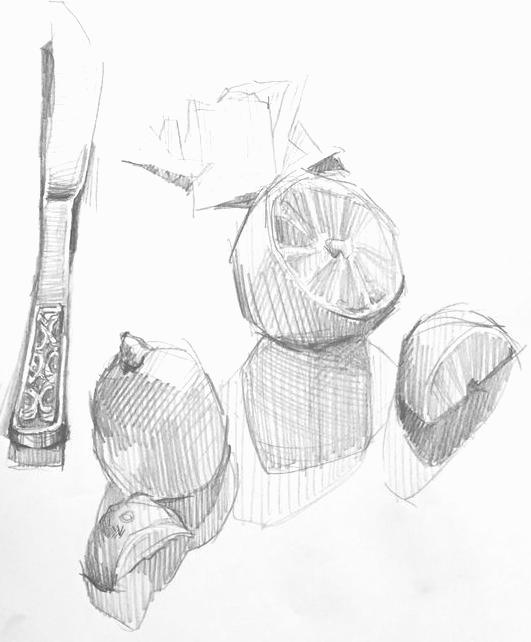} &
\includegraphics[width = .24\linewidth]{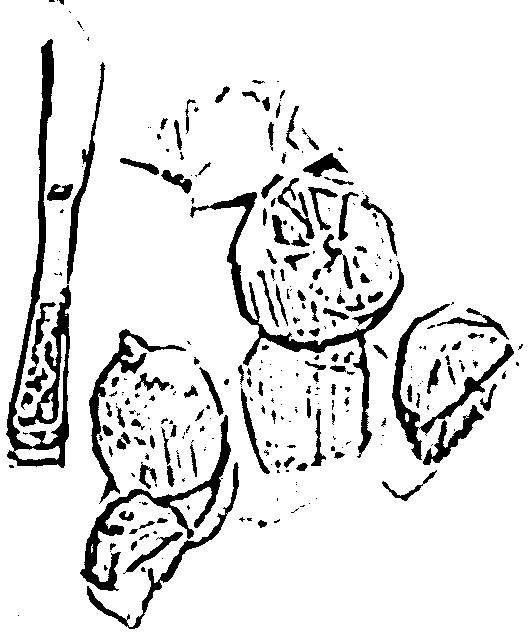} &
\includegraphics[width = .24\linewidth]{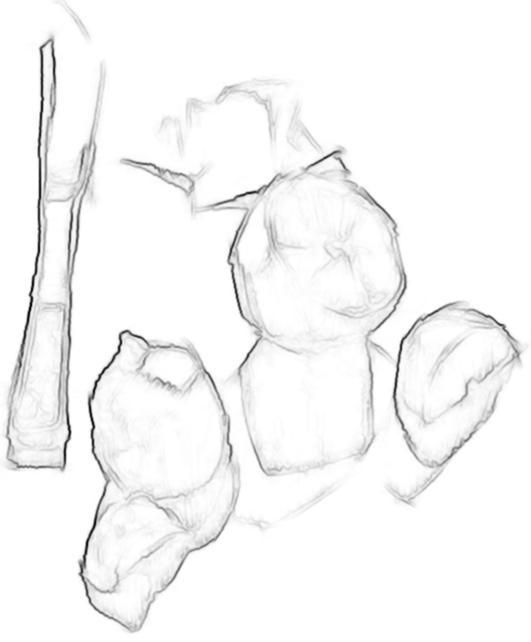} & 
\includegraphics[width = .24\linewidth]{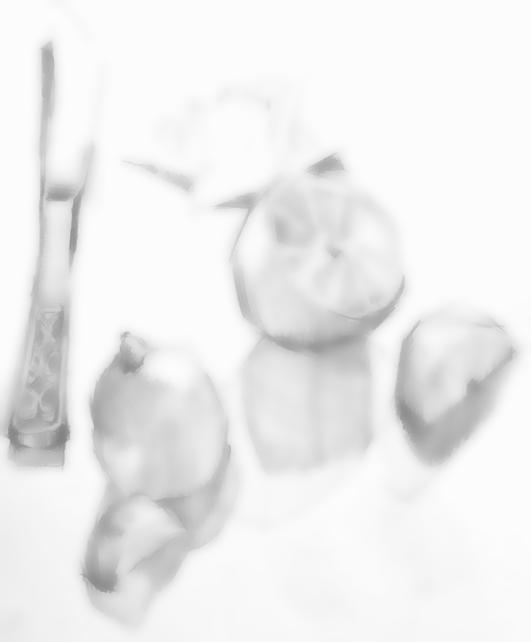} & \\

\includegraphics[width = .24\linewidth]{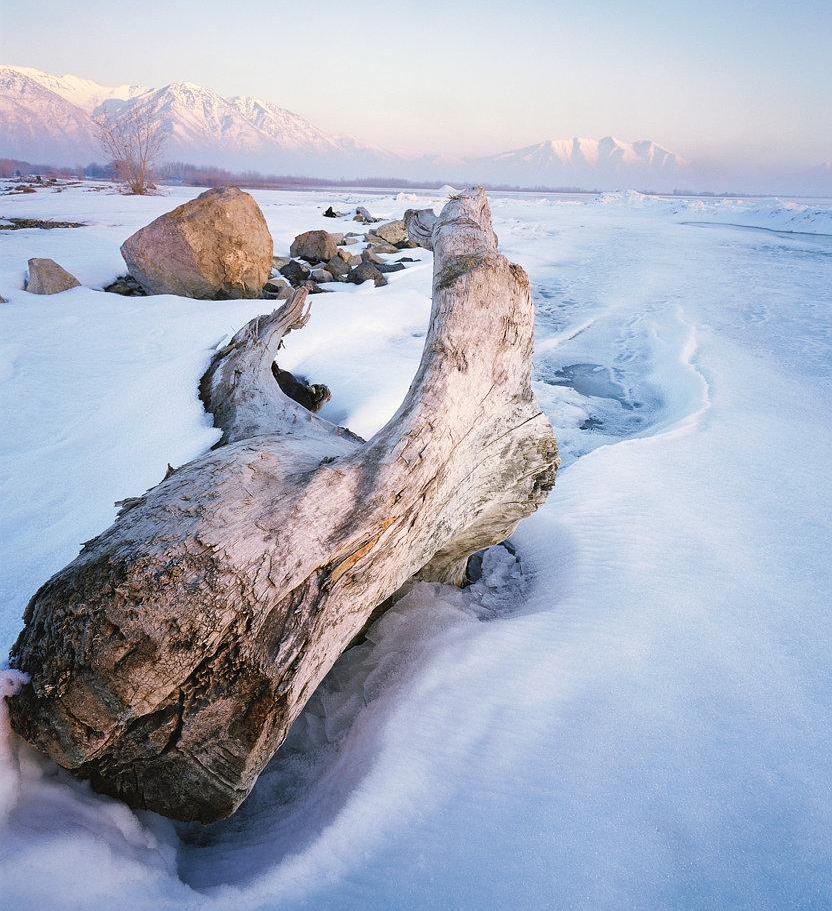} & 
\includegraphics[width = .24\linewidth]{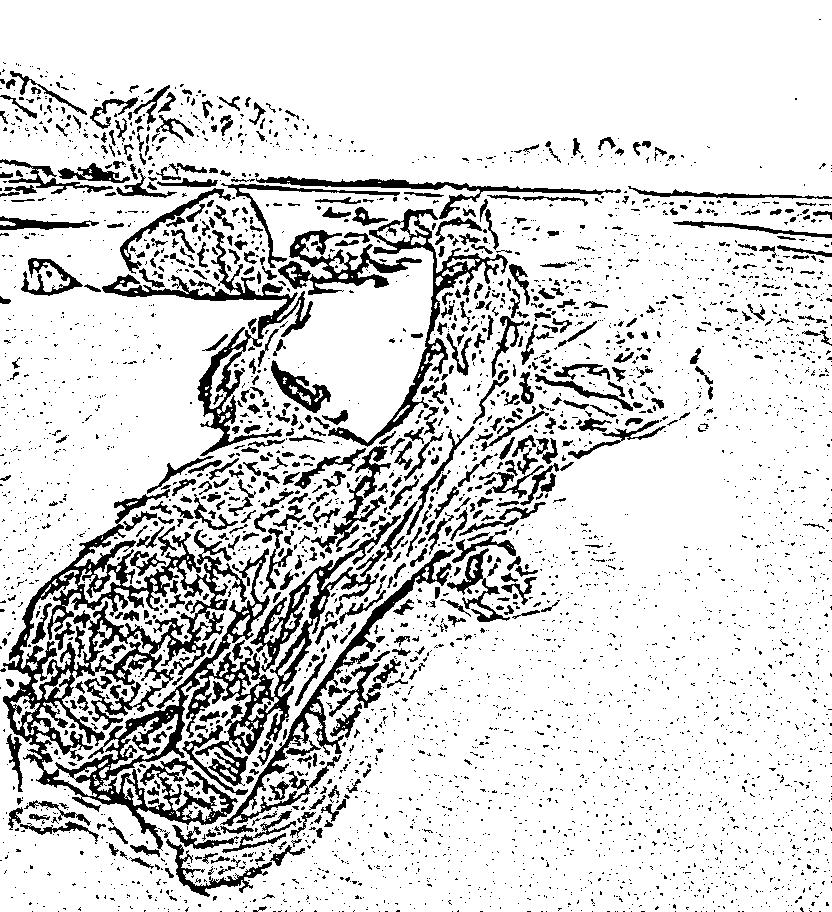} & 
\includegraphics[width = .24\linewidth]{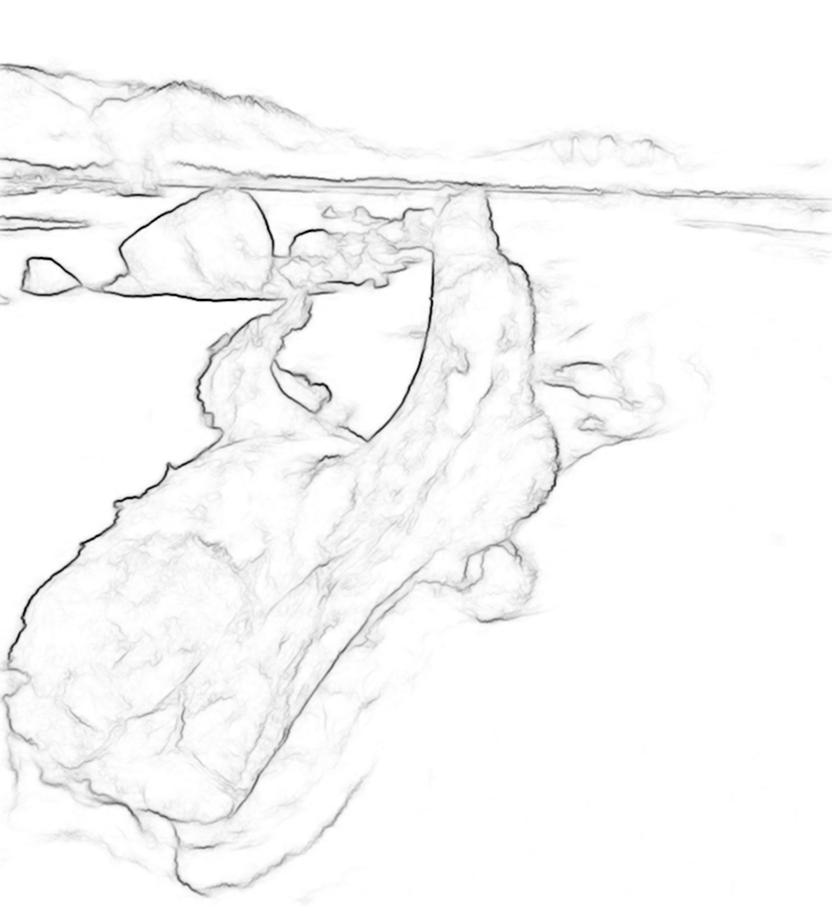} & 
\includegraphics[width = .24\linewidth]{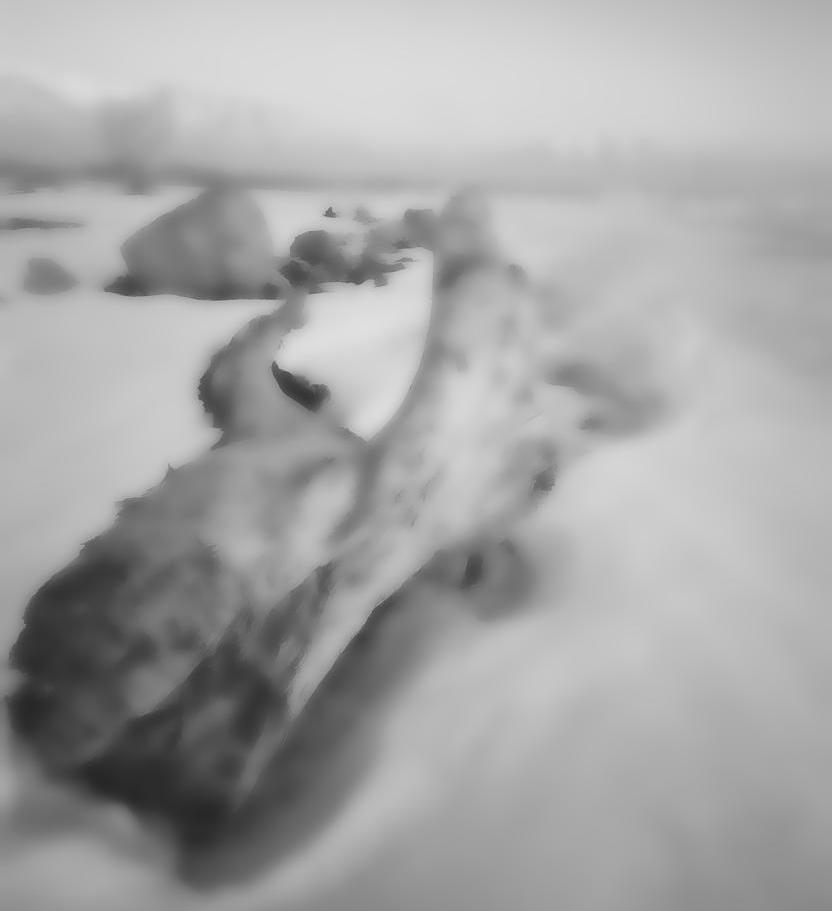} & \\

{(a) Input } & {(b) XDoG~\cite{XDoG2012}} & {(c) Edge~\cite{DollarICCV13edges}} & {(d) Tone}\\

\end{tabular}
\vspace{0.05em}
\caption{Examples of extracted edge and tone results for highly-textured inputs.
}
\label{fig:shading_simplify}
\end{figure}

\vspace{.5em}
\noindent\textbf{Edge and tone extraction.}~
For the edge map, we use the boundary detector by Doll\'ar  and Zitnick \cite{DollarICCV13edges}, which identities important edges, even in highly-textured images. We do not use XDoG after boundary detection, because clean outlines are not necessary for shading generation.
%The shading drawing mainly focuses on non-outline regions.
%
%So we first extract the edge map of input as a starting outline image.
%
%As shading drawings (and some photos) often contain heavy textures (e.g., hatching lines) in non-outline regions, the XDoG filter is not a good choice here because it will also highlight those textures edges, as shown in Figure~\ref{fig:shading_simplify}(b).
%
%Instead, we use the boundary detector by Doll\'ar \etal~\cite{DollarICCV13edges} which is more effective in delineating object contours than the XDOG filter.
%
An example  comparison between XDoG and the boundary detector for a highly-textured input is shown in Figure~\ref{fig:shading_simplify}(b) and (c).

%Starting with the edge map, the shading is drawn according to the tone information of input to reflect different amount of light falling on a region.
%
To extract the tone map, we apply the Guided Filter (GF)~\cite{he2013guided} on the luminance channel of shading drawings or photos to remove details and generate a smoothing output as the tone extraction.
Examples of extracted tone results are shown in Figure~\ref{fig:shading_simplify}(d).

\vspace{.5em}
\noindent\textbf{Paired training data.}~We collect a set of pencil \textit{shading} drawings from online websites,
and annotate each drawing with one of our four style labels, i.e., \textit{hatching}, \textit{crosshatching}, \textit{blending}, and \textit{stippling} (Figure~\ref{fig:real_example}(b)).
We searched the data with each shading style as the main web query and collected 20 shading drawings for each style.
As in the outline branch, we use a 4-bit one-hot vector as a selection unit. 
For each drawing, we extract its edge map and tone map to construct the paired data.
We manually select the best parameters (e.g., the neighborhood size in GF~\cite{he2013guided}) for each shading drawing to produce good abstraction results.
By cropping patches of size 256$\times$256 and doing data augmentations (e.g., rotation) on the paired data, we create about 3000 training pairs.

%\aaron{details about training set size, methodology for gathering images}

\begin{figure}[t]
\centering
\begin{tabular}{c@{\hspace{0.005\linewidth}}c@{\hspace{0.005\linewidth}}c@{\hspace{0.005\linewidth}}c@{\hspace{0.005\linewidth}}c}

\includegraphics[width = .48\linewidth]{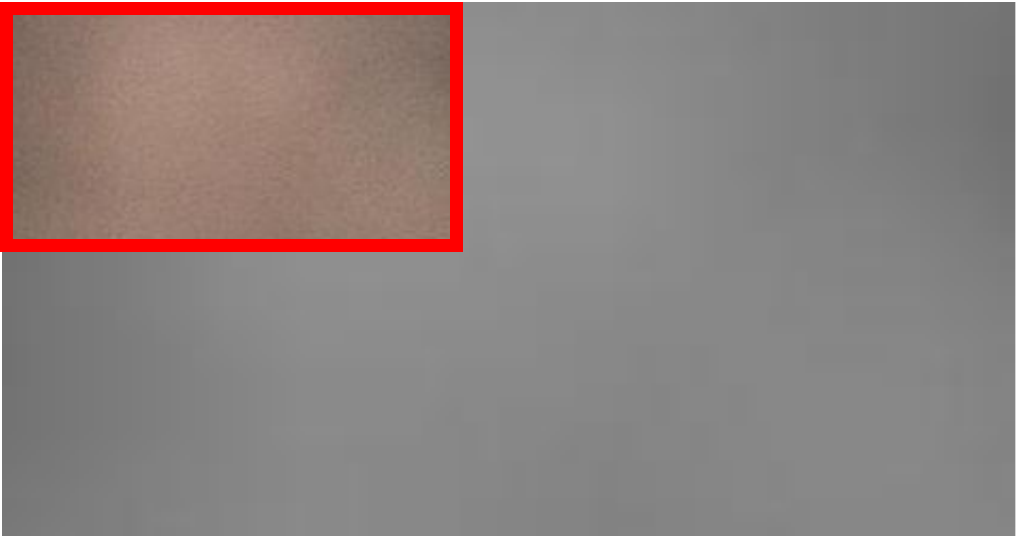} &
\includegraphics[width = .48\linewidth]{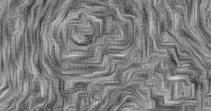} & \\

{(a) Tone (inset: photo patch)} & { (b) Edge tangent field} &\\

\includegraphics[width = .48\linewidth]{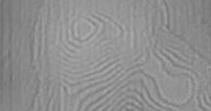} &
\includegraphics[width = .48\linewidth]{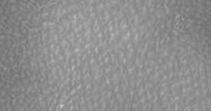} & \\

{(c) Single stream} & {(d) Two-stream} &\\
{tone$\rightarrow$shading} & { \{edge, tone\}$\rightarrow$ shading} &\\

\end{tabular}
\vspace{0.05em}
\caption{Comparisons of shading results on a photo patch (c)-(d) obtained using different network architectures. The input is a smooth photo patch (red inset).
 (a) Extracted tone map. (b) Edge tangent field of (a).
 (c) Hatching result from single-stream architecture. Artificial patterns appear that are unlike normal hatching.
 (d) Hatching result from two-stream architecture. Hatching-like textures are produced and artificial patterns are suppressed.
%\aaron{need more explanation about what each row is and what this is demonstrating, what to look for. seems like you have to zoom in 10x to see anything...}
}
\label{fig:shading_network_comparison}
\end{figure}

\vspace{.5em}
\noindent\textbf{Translation model.}~We find that the direct translation from the extracted tone to the shading drawing generates significant artifacts when applied to photos, especially in smooth regions.
In such regions, artists draw pencil textures with varying orientations; these strokes typically approximate the tone but not the specific gradients in the image. However, naive training produces results that attempt to follow the input gradients too closely.
%the specific pencil textures drawn may be only loosely-related to the gradients
%
%This can be attributed to the different image gradient characteristics of drawings and photos.
%
%The training data of pencil drawings usually have strong gradients (draw or not draw) such that the learned model is biased to actively respond to gradients in the tonal input.
% 
%However, small image gradients often occur in smooth regions in real photos (e.g., sky, wall, facial skin) but almost never in drawings.
%
Figure~\ref{fig:shading_network_comparison}(a) shows a smooth photo patch and its extracted tone. 
We visualize its gradient field in terms of edge tangent field which is perpendicular to the image
gradients using the linear integral convolutions~\cite{LIC-cabral1993}.
When simply relying on the tonal input, the shading result in Figure~\ref{fig:shading_network_comparison}(c) shows that the generated hatching lines looks quite unnatural by following these small gradients.
%
%\eli{Not sure this is the right explanation. Strong gradients happen in real drawings so they are fine (and the problem is not with gradient direction). The problem is with small gradients that occur in smooth region in real photos and almost never occur in drawings.}

To address the above-mentioned issue, we design a two-stream translation model to generate the shading (Figure~\ref{fig:framework}(right-bottom)).
The main stream is from the edge map of the input, where there is no indication of small image gradients.
We employ the tonal abstraction for weak guidance of tone in a secondary input stream that is fused into the main stream at a deeper layer.
Figure~\ref{fig:shading_network_comparison}(d) shows that the shading output of our two-stream network architecture significantly reduces the artifacts and exhibits more natural and realistic strokes.
In addition, the 4-bit selection unit is fed to the network in the same way as in the outline branch to guide the generation towards different shading styles.

%MH: Learning to draw or sketch...
%\subsection{Learning the translation}
\subsection{Learning to draw}

%MH: translated to? or from?
%With the created paired data, we train the model to learn how the simplified drawings are translated to original ones.
With the paired data, we train the model to learn to translate from abstracted inputs to drawings.
Our training is based on the existing translation frameworks, e.g., Pix2Pix~\cite{Pix2Pix-CVPR2017}. 
However, we use different loss functions for pencil drawing as the existing ones do not perform well for our task. 
%\aaron{be more specific... maybe when we have the results to look at.}
%
%We make several improvements over losses for a successful learning on pencil drawing generation. 
%We use different loss functions 
%
We now describe our loss functions used for both branches of our model.

\begin{figure}[t]
\centering
\begin{tabular}{c@{\hspace{0.005\linewidth}}c@{\hspace{0.005\linewidth}}c@{\hspace{0.005\linewidth}}c@{\hspace{0.005\linewidth}}c}

\includegraphics[width = .24\linewidth]{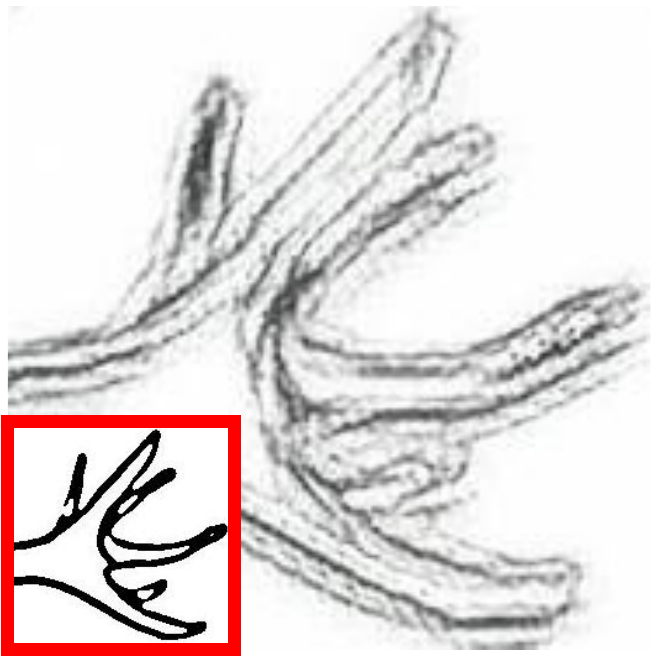} &
\includegraphics[width = .24\linewidth]{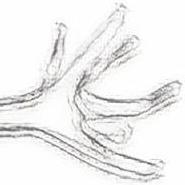} & 
\includegraphics[width = .24\linewidth]{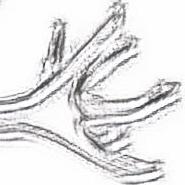} &
\includegraphics[width = .24\linewidth]{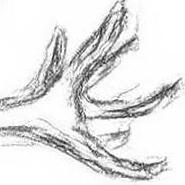} &\\

{(a)} & {(b)} & {(c)} & {(d)}&\\
{~$L_{rec}$+$L_{adv}$ } & {~$L_{per}$+$L_{adv}$} & {~$L_{rec}$+$\sum L_{adv}^{i}$} & {Ours}\\

\end{tabular}
\vspace{0.05em}
\caption{Comparisons of pencil outline results obtained by models trained with different loss functions. Inset image in red rectangle: the XDoG input. $L_{rec}$: reconstruction loss. $L_{adv}$: adversarial loss using single discriminator on patches of 256$\times$256. $L_{per}$: perceptual loss. $\sum L_{adv}^{i}$: adversarial loss using three discriminators on patches of 256$\times$256, 128$\times$128 and 64$\times$64. Our final loss is the combination of $L_{per}$ and $\sum L_{adv}^{i}$.
}
\label{fig:loss_ablation}
\end{figure}

\begin{figure*}[t]
\centering
\begin{tabular}{c@{\hspace{0.005\linewidth}}c@{\hspace{0.005\linewidth}}c@{\hspace{0.005\linewidth}}c@{\hspace{0.005\linewidth}}c@{\hspace{0.005\linewidth}}c}

\includegraphics[width = .19\linewidth]{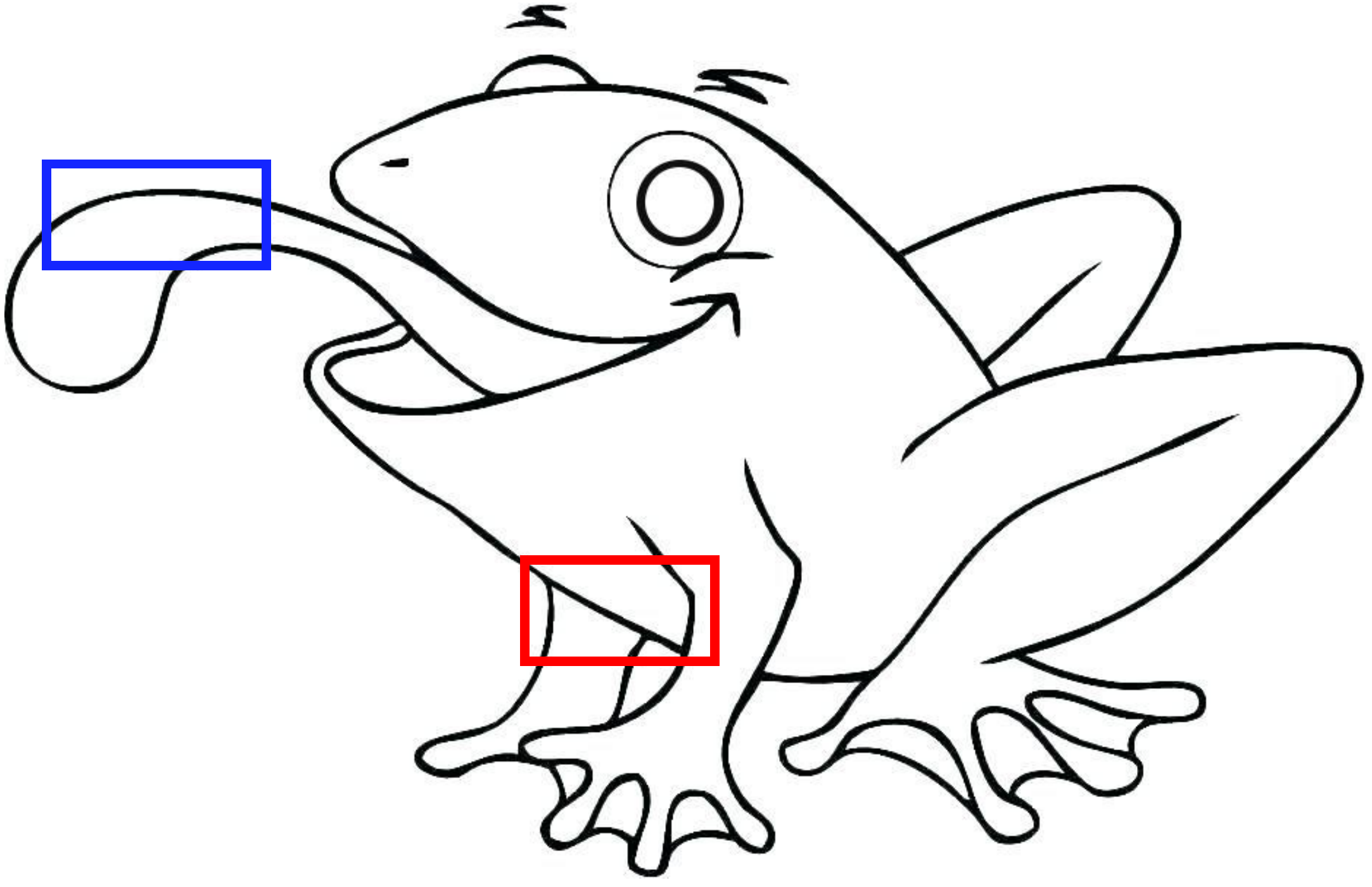} &
\includegraphics[width = .19\linewidth]{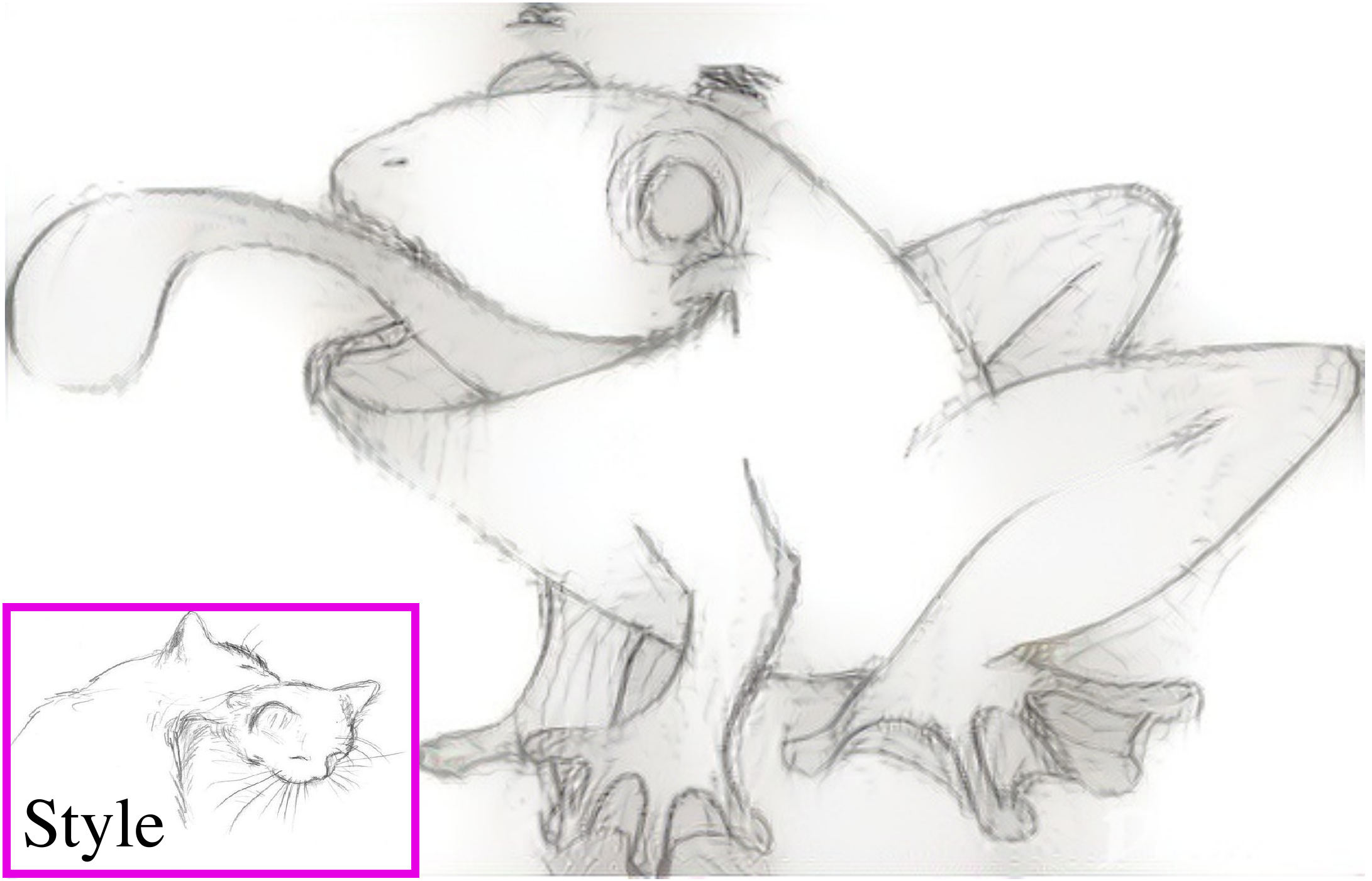} &
\includegraphics[width = .19\linewidth]{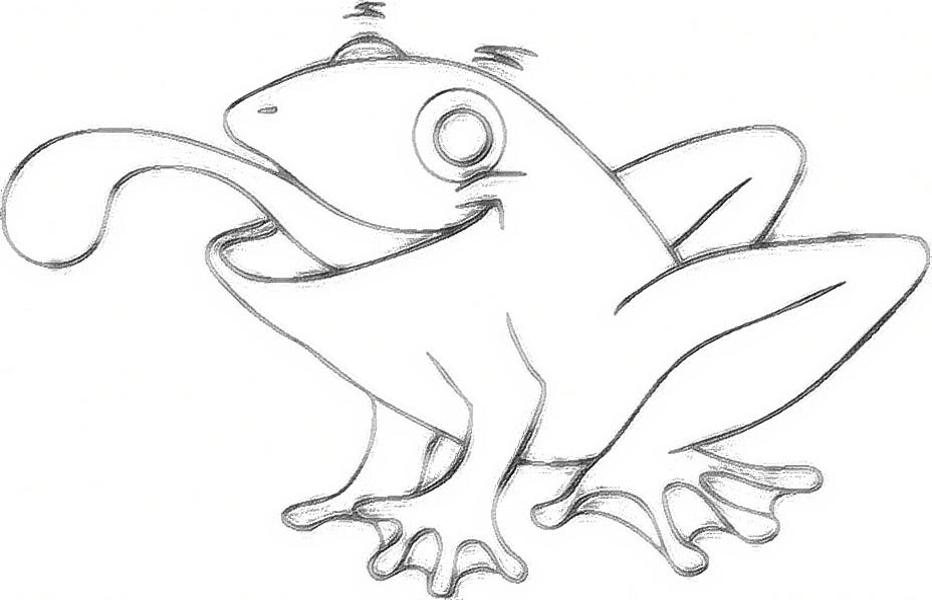} & 
\includegraphics[width = .19\linewidth]{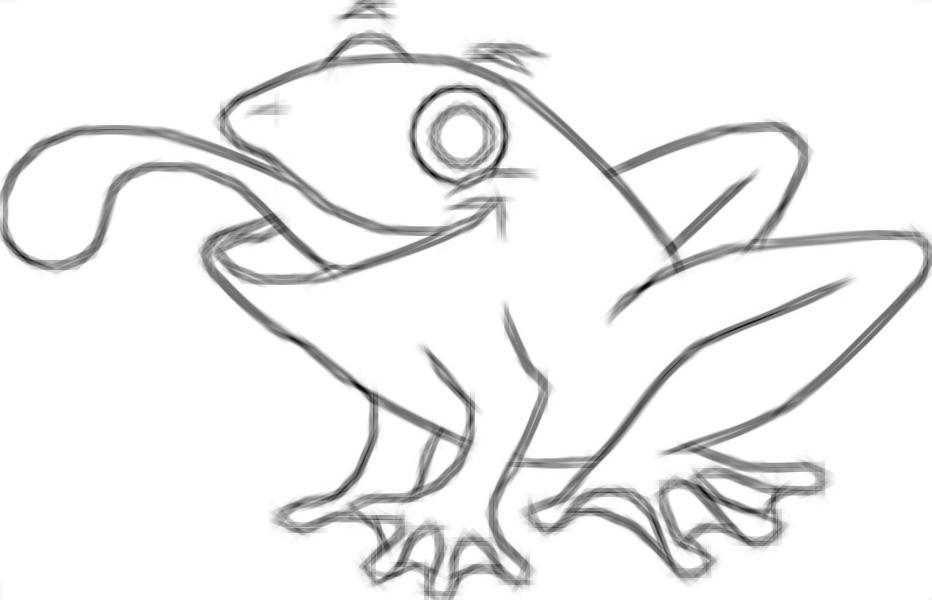} & 
\includegraphics[width = .19\linewidth]{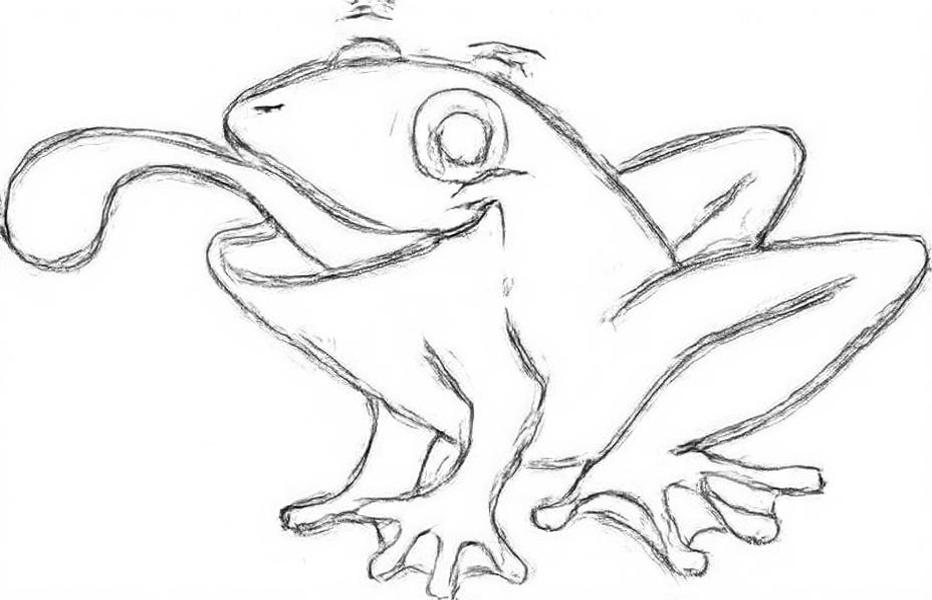} & \\

\includegraphics[width = .19\linewidth]{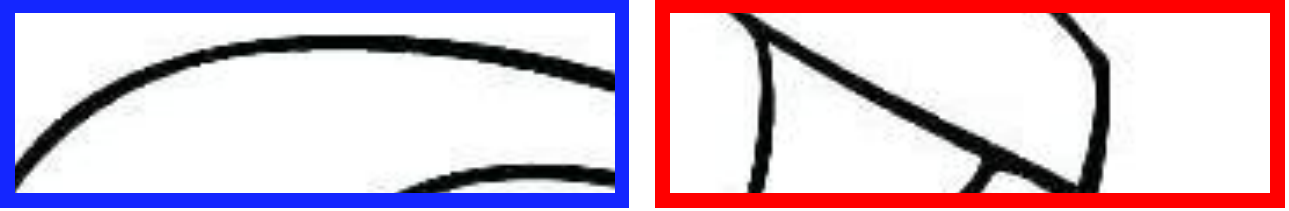} &
\includegraphics[width = .19\linewidth]{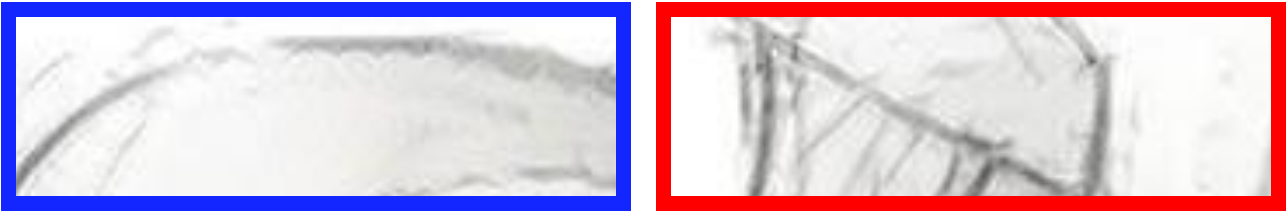} &
\includegraphics[width = .19\linewidth]{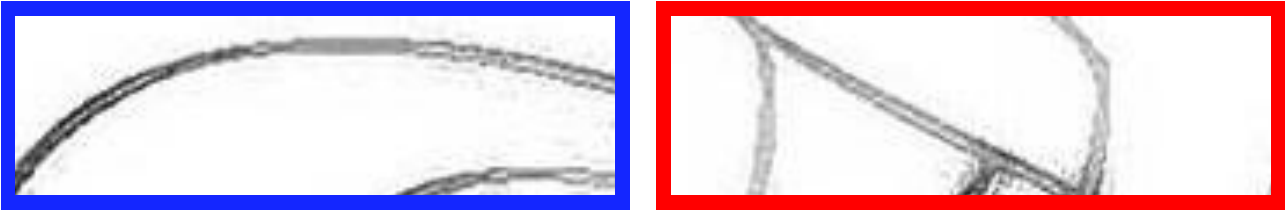} & 
\includegraphics[width = .19\linewidth]{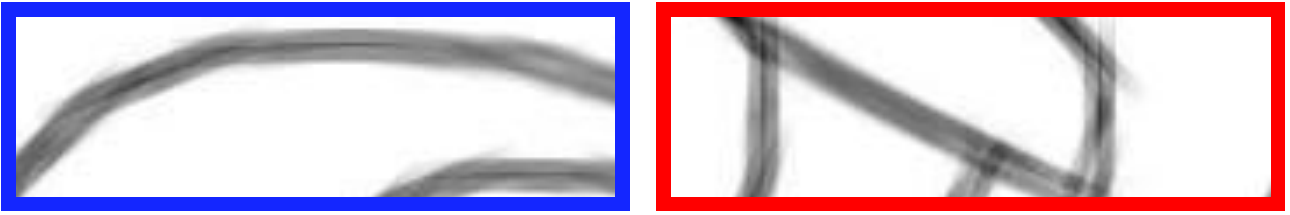} & 
\includegraphics[width = .19\linewidth]{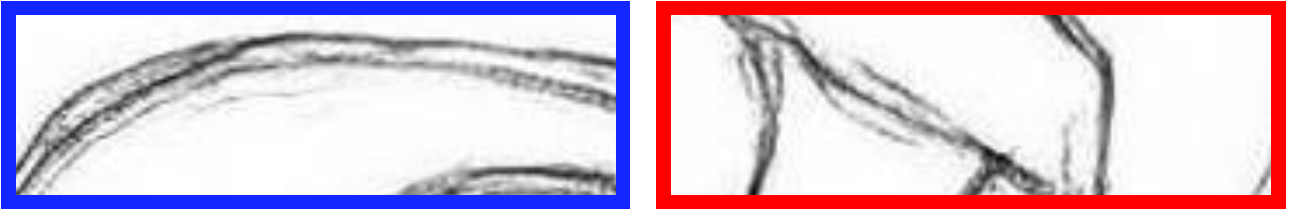} & \\

\includegraphics[width = .19\linewidth]{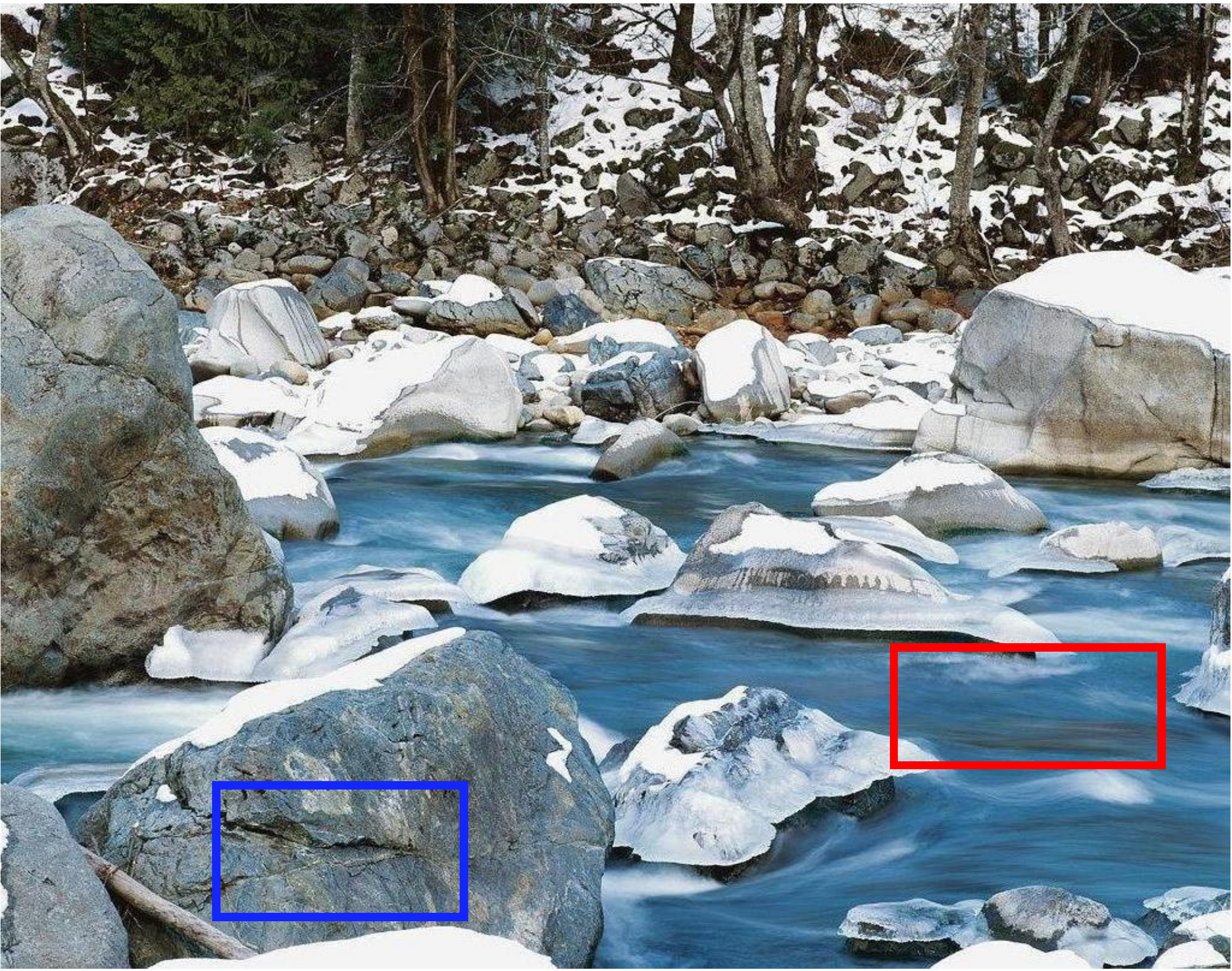} &
\includegraphics[width = .19\linewidth]{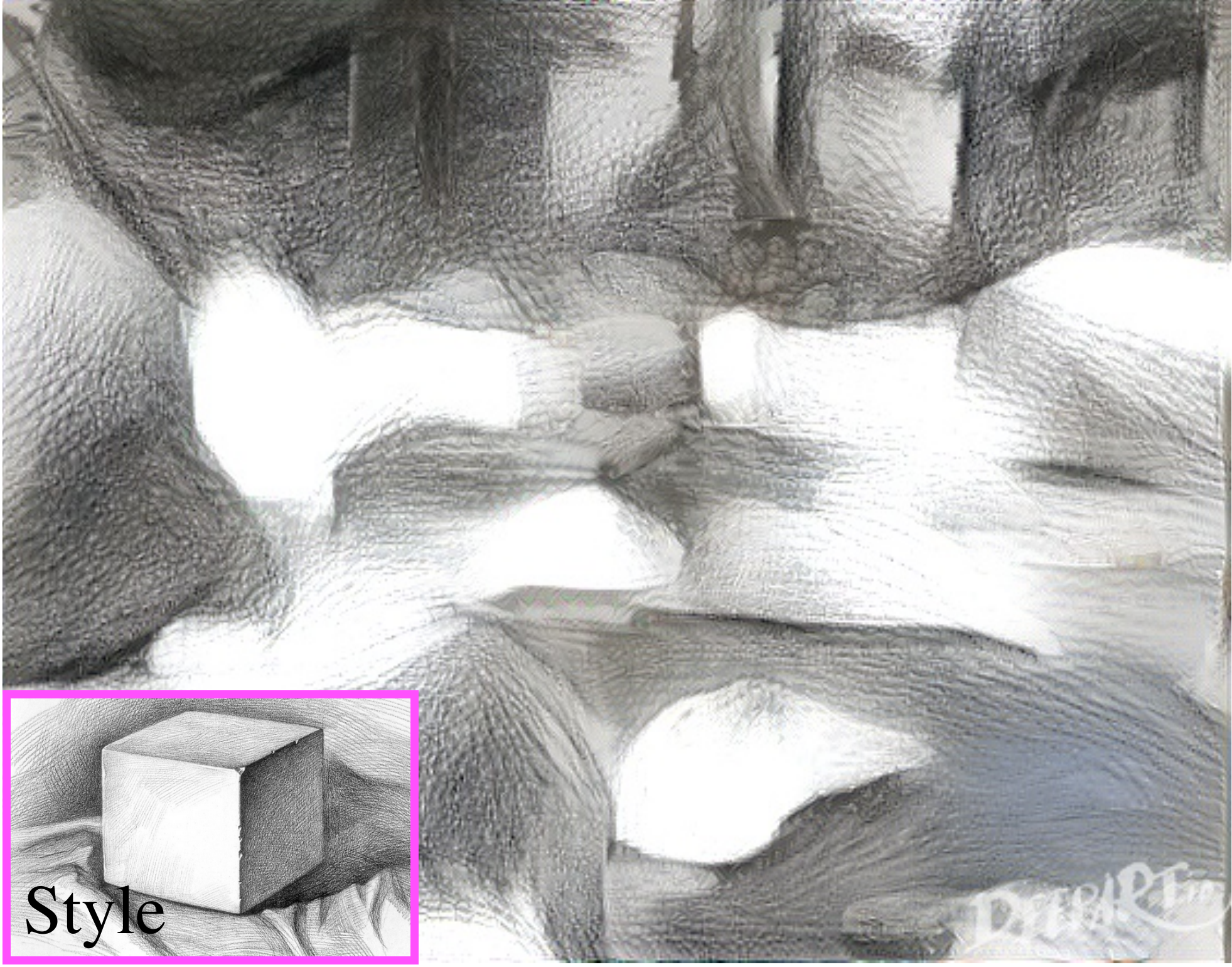} &
\includegraphics[width = .19\linewidth]{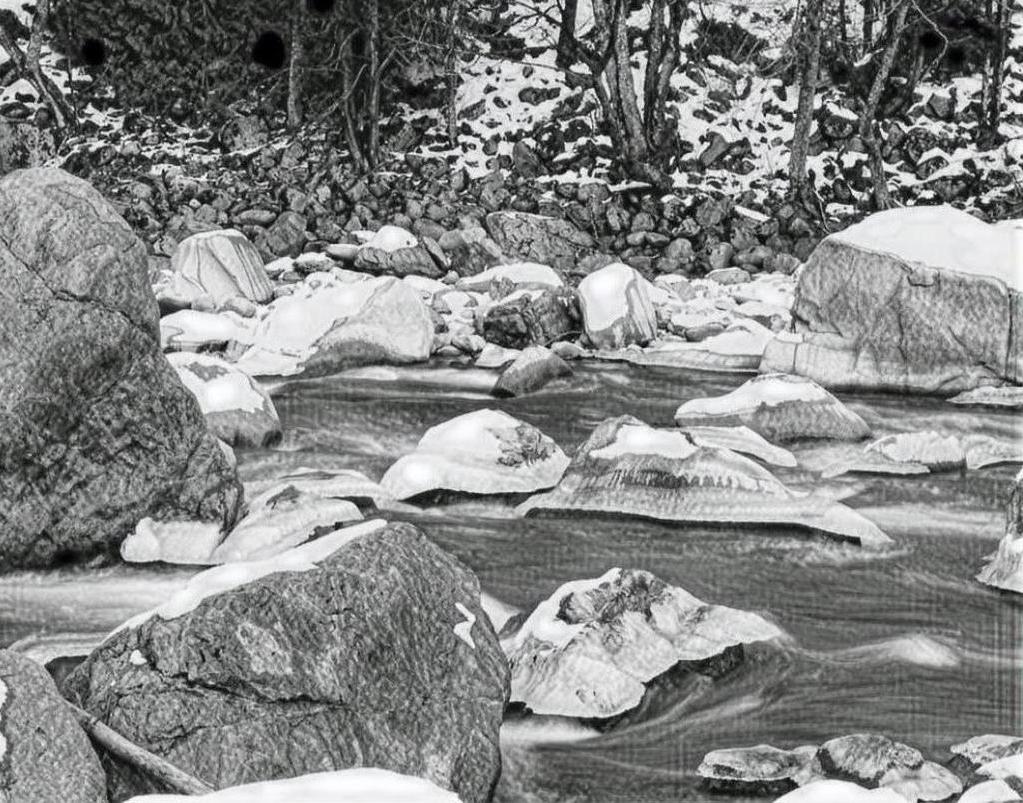} & 
\includegraphics[width = .19\linewidth]{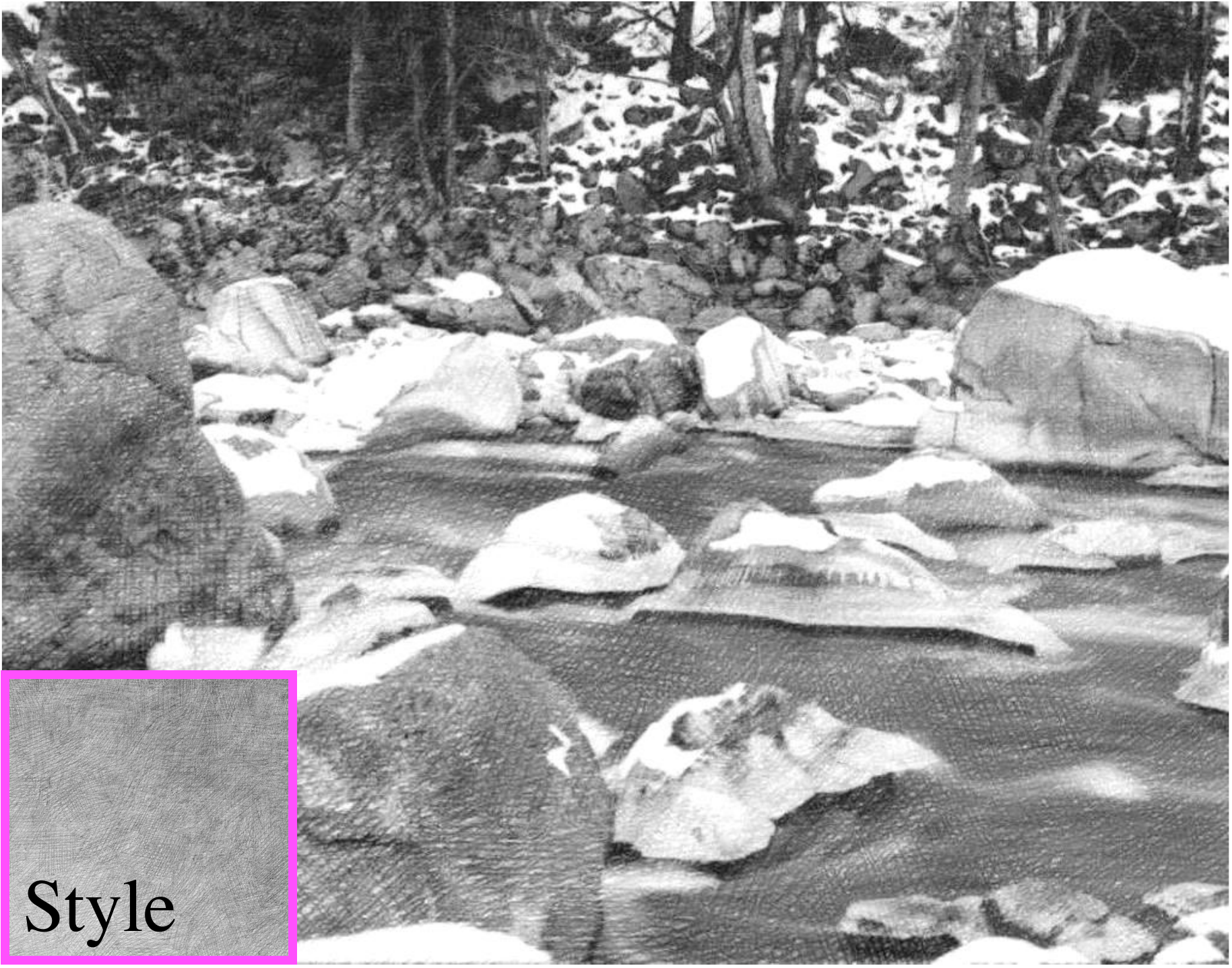} & 
\includegraphics[width = .19\linewidth]{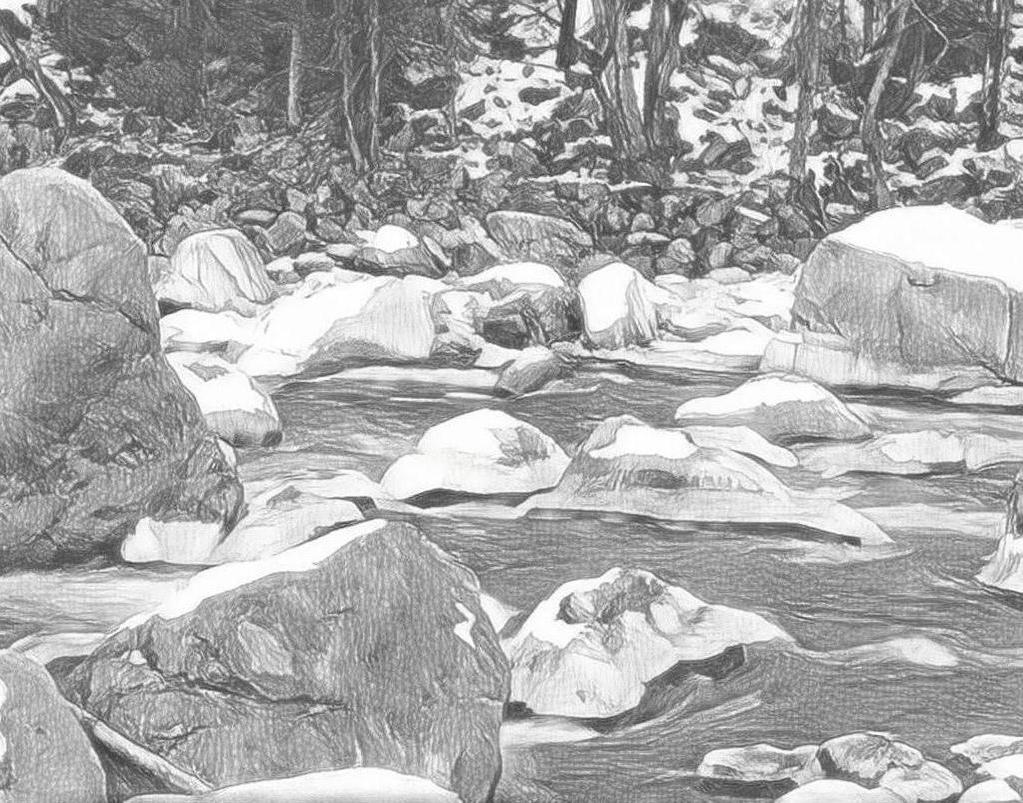} & \\

\includegraphics[width = .19\linewidth]{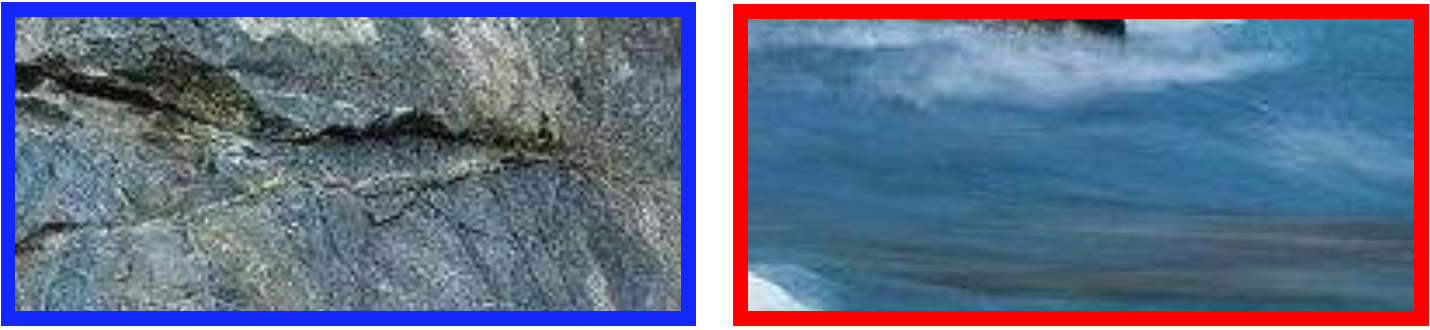} &
\includegraphics[width = .19\linewidth]{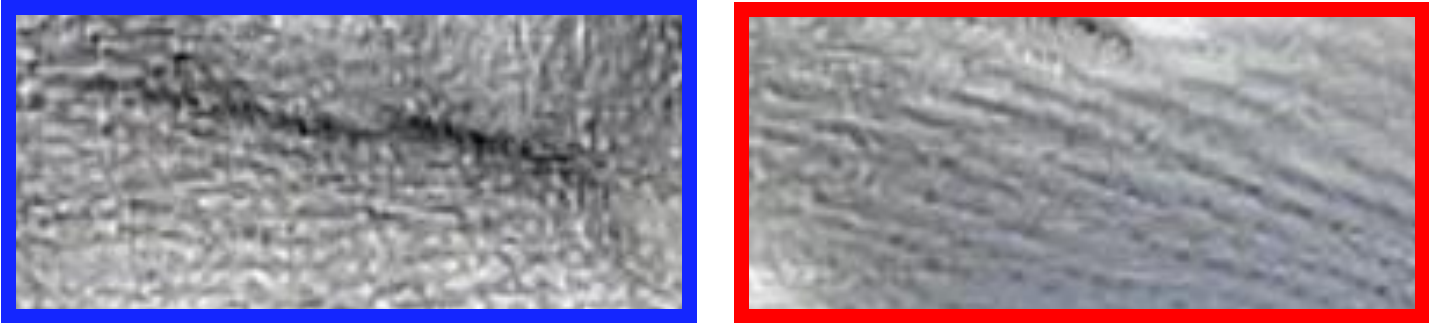} &
\includegraphics[width = .19\linewidth]{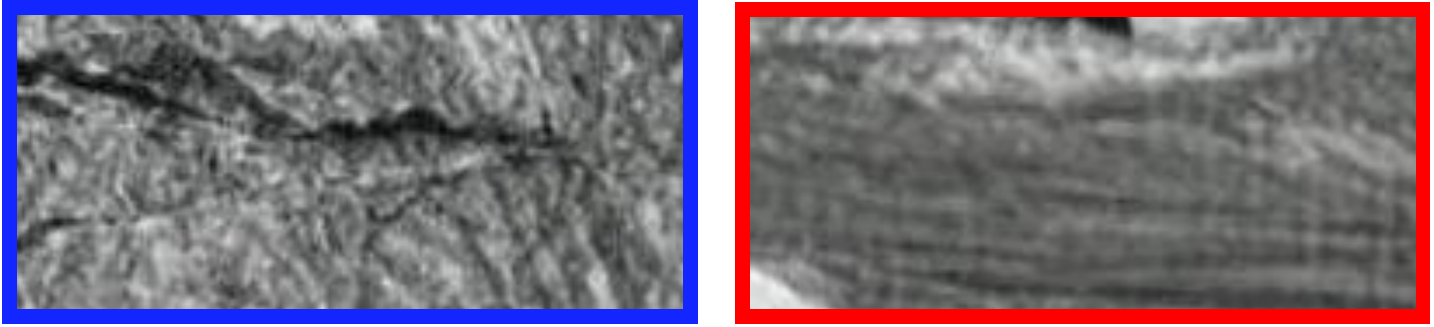} & 
\includegraphics[width = .19\linewidth]{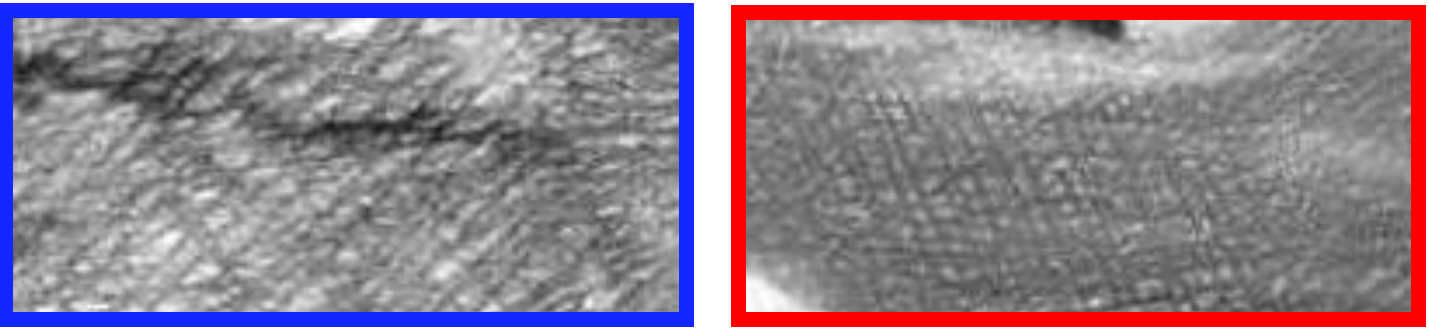} & 
\includegraphics[width = .19\linewidth]{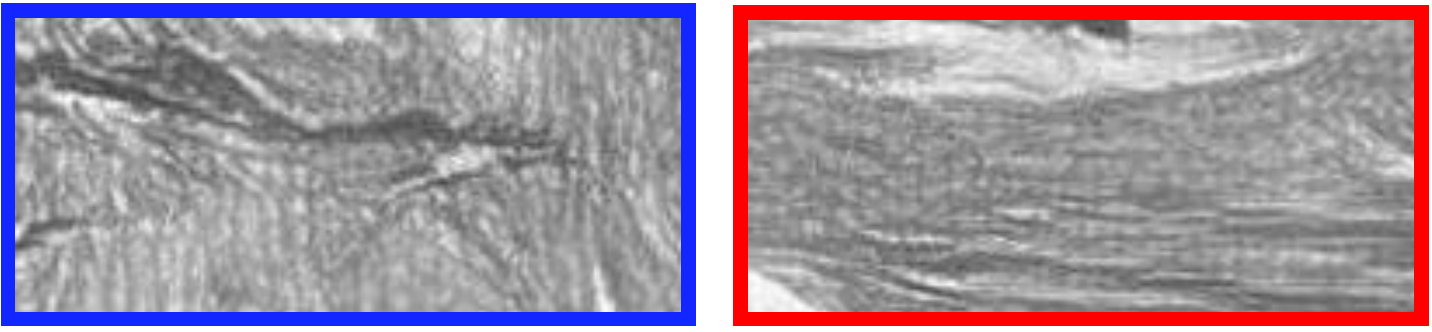} & \\

{(a) Input} & {(b) Gatys \etal~\cite{GatysTransfer-CVPR2016}} & {(c) CycleGAN~\cite{CycleGAN-ICCV17}} & {(d) Lu \etal~\cite{lu-NPAR2012}} & {(e) Ours}&\\

\end{tabular}
\vspace{0.05em}
\caption{Visual comparisons of different methods for rendering pencil drawing effects (\textbf{zoom in for details}).
The exemplars used for~\cite{GatysTransfer-CVPR2016,lu-NPAR2012} are shown in the pink rectangle.
Top: to directly show pencil effects for the outline, we select a simple line input which is only filtered by the XDoG (no need to detect boundaries first). Bottom: pencil shading effects on a real photo example. 
}
\label{fig:comparison}
\end{figure*}

\vspace{.5em}
\noindent\textbf{Perceptual loss.}~We use the perceptual loss~\cite{Perceptual-ECCV2016} to minimize the difference between the network output and the ground truth (GT) pencil drawing based on their deep features:
%
%It is defined as:
%
\begin{equation}
\label{Perceptual_loss}
L_{per} = \sum \limits_{i=1}^{4} \mid \mid \Phi_{i}(G(x)) - \Phi_{i}(y) \mid \mid^{2}_{2}~,
\end{equation}
where $x$, $y$ are the input and the GT pencil drawing, $G$ is the translation model, and $\Phi_{i}$ is the VGG-19~\cite{VGG-2014} network up to the ReLU\_i\_1 layer.
The feature-based perceptual loss has been shown to generate sharper results than the pixel-based reconstruction loss $L_{rec}$. 
%
%The comparison between $L_{rec}$ and $L_{per}$ is shown in Figure~\ref{fig:loss_ablation}(a) and (b).
%\eli{Rewrite this and say that we found that a perceptual loss works better for comparing drawings than a color loss, and leads to sharper results.}

\vspace{.5em}
\noindent\textbf{Adversarial loss.}~In addition to the translation model $G$, we use the discriminator network $D$ to discriminate between real samples from the pencil drawings and generated results.
The goal of $G$ is to generate images that cannot be distinguished by $D$. 
This can be achieved by using an adversarial loss~\cite{goodfellow-2014-GAN}:
\begin{dmath}
\label{GAN_loss}
L_{adv} = \min \limits_{G}\max \limits_{D} ~\mathcal{E}_{y \sim P_{Y}}[\log D(y)] +\mathcal{E}_{x\sim P_{X}}[\log (1- D(G(x)))],
\end{dmath}
where $P_{Y}$ and $P_{X}$ represent the distributions of pencil drawing samples $y$ and their abstracted samples $x$.

To better capture both the global style and local strokes of pencil drawings, we propose to discriminate between real data and fake generations at {\em multiple scales} during the training. 
Specifically, given the generated output of size 256$\times$256, we use three discriminators to discriminate patches on three scales (256$\times$256, 128$\times$128, 64$\times$64). Each discriminator is designed as the PatchGAN used in~\cite{Pix2Pix-CVPR2017}.
The overall loss function is defined by:
\begin{equation}
\label{total_loss}
 L = L_{per} + \beta \sum \limits_{i=1}^{3} L_{adv}^{i},
\end{equation}
where $\beta$ is the weight to balance different losses, and $L_{adv}^{i}$ is the adversarial loss of the $i$th discriminator.
We set $\beta=100$ in all experiments.
When learning with multiple styles, in each iteration, all examples in the batch are limited to be of the same style.
Meanwhile, we set the corresponding bit that represents the selected style as 1 and leave other bits as 0 in the selection unit.

%MH: can be written more clearly using fewer words
%Figure~\ref{fig:loss_ablation} shows the comparisons between outline results of models trained with different losses.

Figure~\ref{fig:loss_ablation} shows the outline results from models trained with different losses.
It is observed that the perceptual loss encourages better sharpness for a single line drawn along the silhouette compared with the reconstruction loss.
Meanwhile, using multiple discriminators helps synthesize better pencil strokes with more sketchy details than employing just one discriminator.
%
%MH: pay attention to details... looks?
%Figure~\ref{fig:loss_ablation}(d) shows that our outline results obtained by combining both losses in (\ref{total_loss}) looks more like a real drawing.
%
Figure~\ref{fig:loss_ablation}(d) shows that outline results obtained by the proposed method using the loss function in (\ref{total_loss}) look more like a real drawing.

\section{Experimental Results}

In this section, we present extensive experimental results
to demonstrate the effectiveness of our algorithm. 
We compare with methods from both NPR and deep neural network-based stylization.
We experiment with synthesizing pencil drawings in various outline and shading styles in a user-controllable manner.
%
%The source code will be made available to the public. 
More results and comparisons are shown in the supplementary material\footnote{\url{http://bit.ly/cvpr19-im2pencil-supp}}.

%\subsection{Datasets}

%We collect a pencil drawing dataset from online websites. This dataset consists of pencil drawings only and is divided into two parts, i.e., outline drawings and shading drawings.
% 
%Images in the outline drawing subset mainly contain sketchy outlines without shadings, and we further classify them into two categories in terms of the sketchiness, i.e., rough and clean sketches.
% 
%The shading drawing subset is made of images in heavy shadings without obvious sketchy outlines, and we divide them into four categories representing four shading styles, including: hatching, crosshatching, blending, and stippling. 
% 
%Each category in the two subsets has 30 images respectively.
%
%During the training, we randomly crop patches of size 256$\times$256 on the collected pencil data so that we have thousands of training patches.

\subsection{Comparisons}
\label{method_compared}

\begin{table}[t]
  \caption{User preference towards different methods (\%).
  Each row represents one user study, comparing three stylization algorithms. The top row is applied to the input image directly, and the bottom row uses the tonal adjustment of \cite{lu-NPAR2012} as a preprocess.
  }
  \vspace{0.5em}
  \label{table:user_study}
  \centering
  \begin{tabular}{cccc}
    \toprule
    Methods & CycleGAN~\cite{CycleGAN-ICCV17} & Lu \etal~\cite{lu-NPAR2012} & Ours\\
    \midrule
    Original tone  & 10.3 & 11.4 & \textbf{78.3} \\
    Adjusted tone  & 7.1 & 32.6 & \textbf{60.3} \\
    \bottomrule
  \end{tabular}
\end{table}

We compare with three  algorithms~\cite{GatysTransfer-CVPR2016,CycleGAN-ICCV17,lu-NPAR2012} that represent neural style transfer, unpaired image-to-image translation, and NPR respectively.
As the method of Gatys \etal~\cite{GatysTransfer-CVPR2016} is example-based, we select a representative pencil outline example (the pink inset in Figure~\ref{fig:comparison}(b)) from our dataset to obtain their style transfer results.
For CycleGAN~\cite{CycleGAN-ICCV17}, in order to train a model for pencil drawing, we collect a photo dataset that consists of 100 images from online websites.
Together with our pencil dataset, they construct an unpaired dataset that is used to train CycleGAN for translation between the two domains.
Note that since CycleGAN only supports transferring a certain kind of style, one needs to train different CycleGAN model for each outline and shading styles.
The NPR method of Lu \etal~\cite{lu-NPAR2012} has a two-phase design as well, treating the outline and shading drawing separately.
The shading drawing phase also requires a real pencil shading example.
Since Lu \etal~\cite{lu-NPAR2012} do not release the shading example used for their results, we select a representative pencil shading example (the pink inset in Figure~\ref{fig:comparison}(d)) from our dataset to generate their results.
%
%Pencil examples used for the CycleGAN and Lu \etal results are shown as insets in the pink rectangle in .

\begin{figure}[t]
\centering
\begin{tabular}{c@{\hspace{0.005\linewidth}}c@{\hspace{0.005\linewidth}}c@{\hspace{0.005\linewidth}}c@{\hspace{0.005\linewidth}}c@{\hspace{0.005\linewidth}}c}

\includegraphics[width = .48\linewidth]{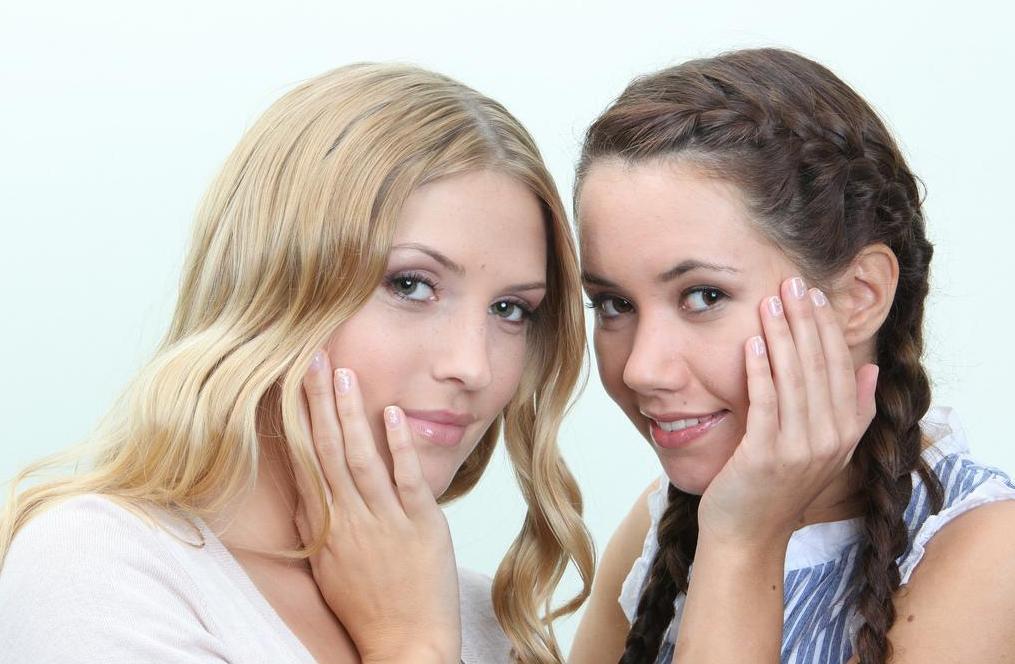} &  
\includegraphics[width = .48\linewidth]{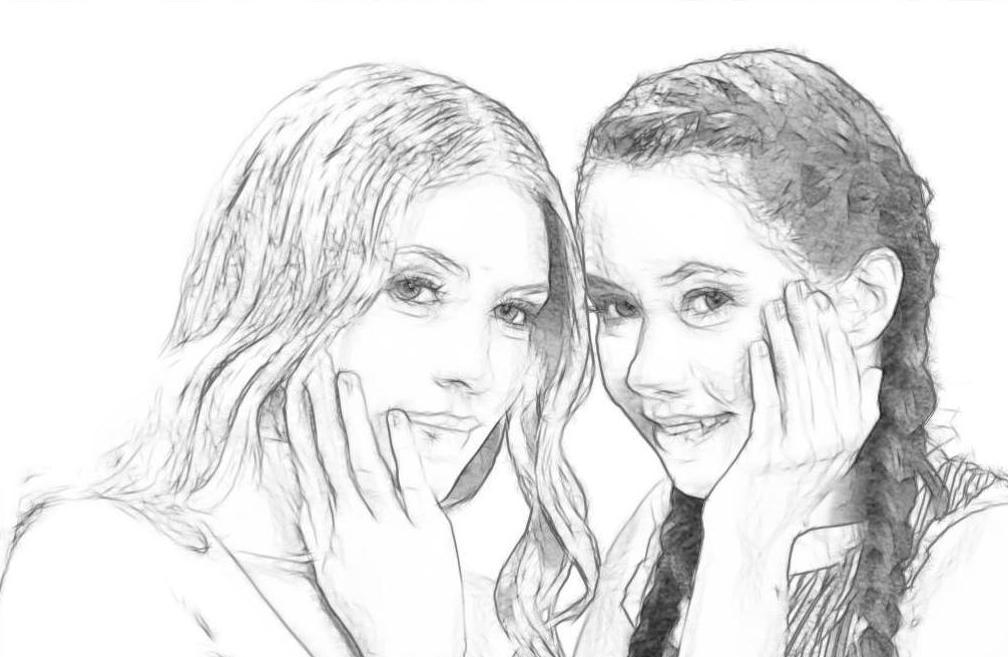} & \\

{(a) Input } &{(b)~\cite{lu-NPAR2012} w/ tone adjust} &  \\

\includegraphics[width = .48\linewidth]{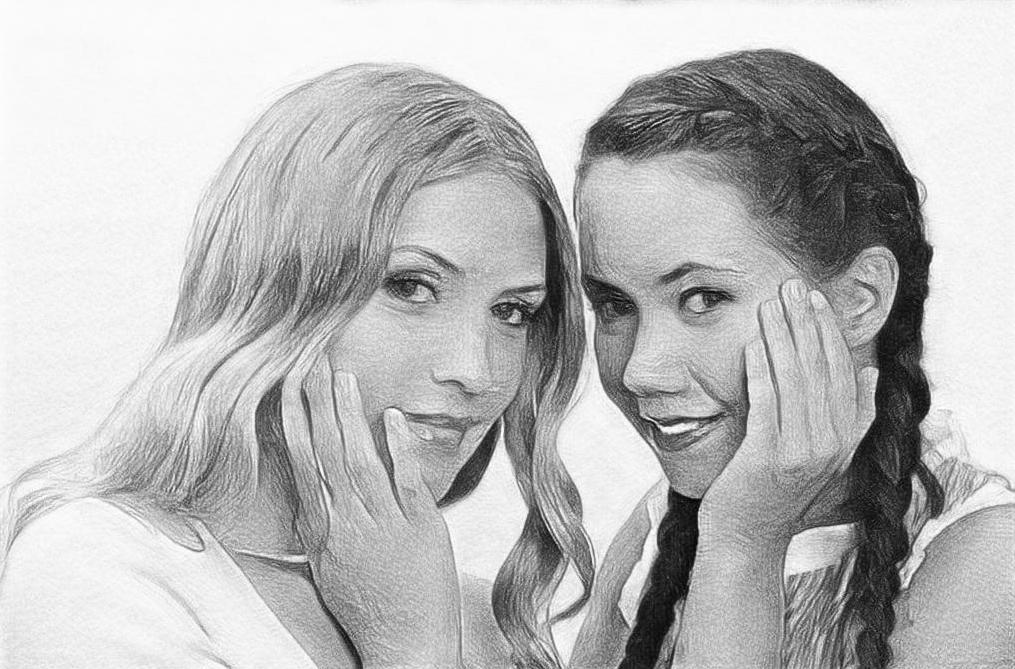} & 
\includegraphics[width = .48\linewidth]{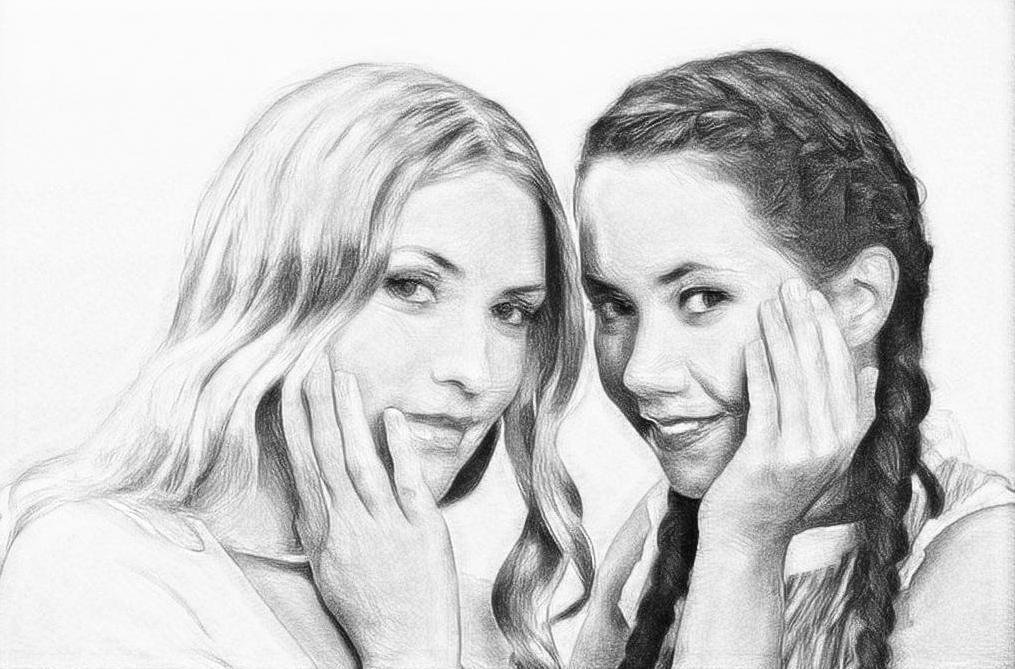} & \\

{(c) Ours w/o tone adjust}& {(d) Ours w/ tone adjust} & \\

\end{tabular}
\vspace{0.05em}
\caption{Comparisons of shading results between w/ and w/o adjusting the tone of input.}
\label{fig:tone}
\end{figure}

Figure~\ref{fig:comparison} shows the visual comparison of three methods on outline drawing (top) and shading drawing (bottom).
The results by Gatys \etal~\cite{GatysTransfer-CVPR2016} in (b) only exhibit some global gray-like feel of pencil drawing.
The lines are unnatural and shading is not correctly positioned to reflect the contrast in the input.
CycleGAN~\cite{CycleGAN-ICCV17} generates a few random textures around outlines and inside regions, leading to results that look like the gray image of the input without clear pencil strokes, as shown in Figure~\ref{fig:comparison}(c).
Without a paired correspondence, simply relying on the cycle constraint and adversarial training is still limited in capturing the real distribution of the pencil drawing domain, with realistic strokes. 
The shading results of Lu \etal~\cite{lu-NPAR2012} in the second row of Figure~\ref{fig:comparison}(d) show shading lines and crossings that have no correlation with the underlying lines, structures and ``natural'' orientations of some content (e.g., the water flow). The shading lines come from a real drawing but the overall result does not look like one.
%some obvious crossings at the line adjunctions but they appear too frequently at every turning point. 
%
In addition, their gradient-based features also result in detecting the two sides of thick lines in the first row of Figure~\ref{fig:comparison}(d), which is uncommon in drawing strokes.
%
%The bottom row of Figure~\ref{fig:comparison}(d) show the shading result of~\cite{lu-NPAR2012}.
%
%Note that the spatial arrangement of the pencil texture is fully preserved in the result (a close-up is shown in red rectangle).
%
%Moreover, the line direction in the pencil texture is unlikely to fit well for every region in the input. 
%
%For example, the water in second example is flowing horizontally but shading lines are still drawn vertically.
%
In contrast, our results in Figure~\ref{fig:comparison}(e) present more realistic pencil strokes in reasonable drawing directions and contain better shading that corresponds to the contrast in the input.

\begin{figure}[t]
\centering
\begin{tabular}{c@{\hspace{0.005\linewidth}}c@{\hspace{0.005\linewidth}}c@{\hspace{0.005\linewidth}}c@{\hspace{0.005\linewidth}}c@{\hspace{0.005\linewidth}}c}

\includegraphics[width = .32\linewidth]{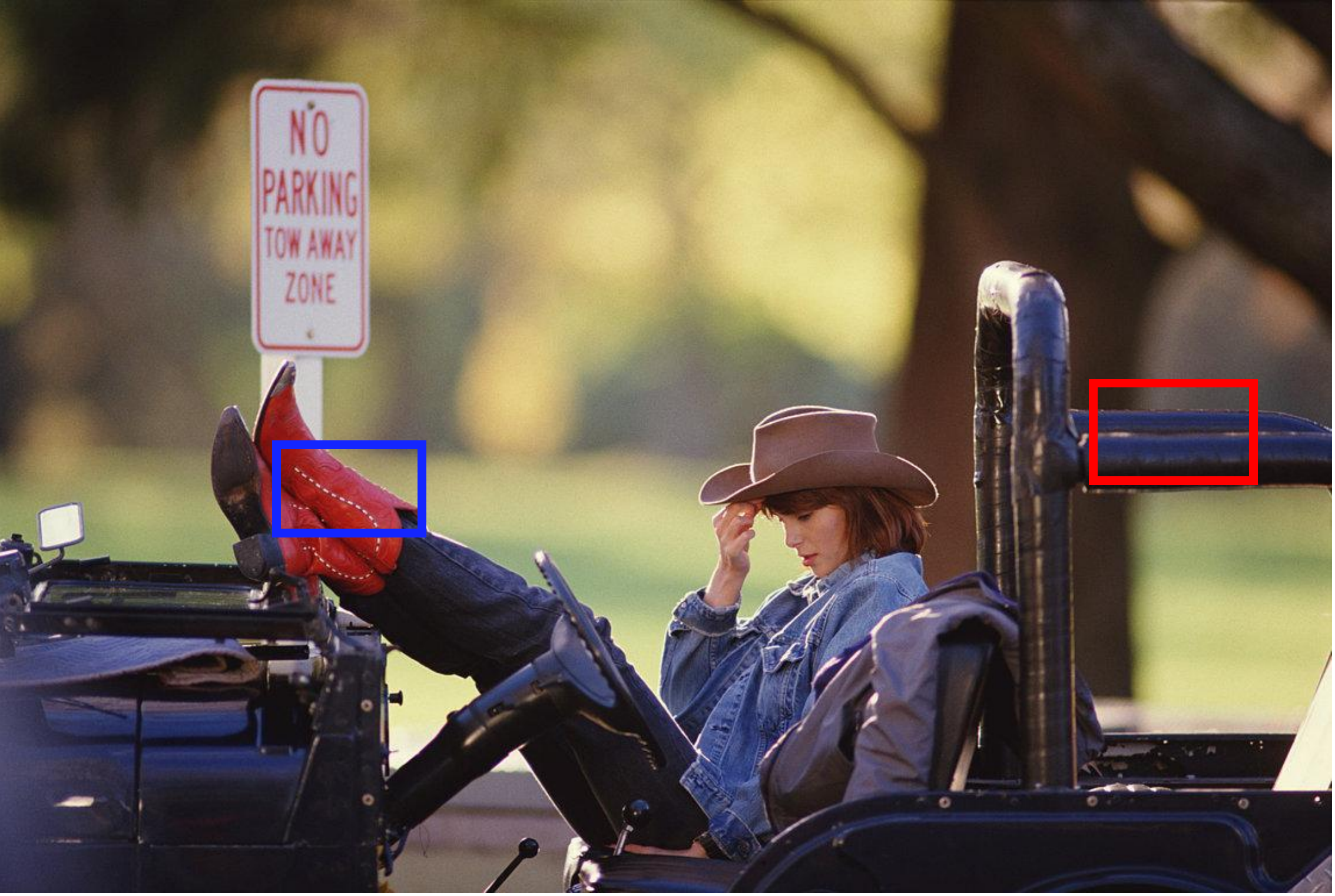} & 
\includegraphics[width = .32\linewidth]{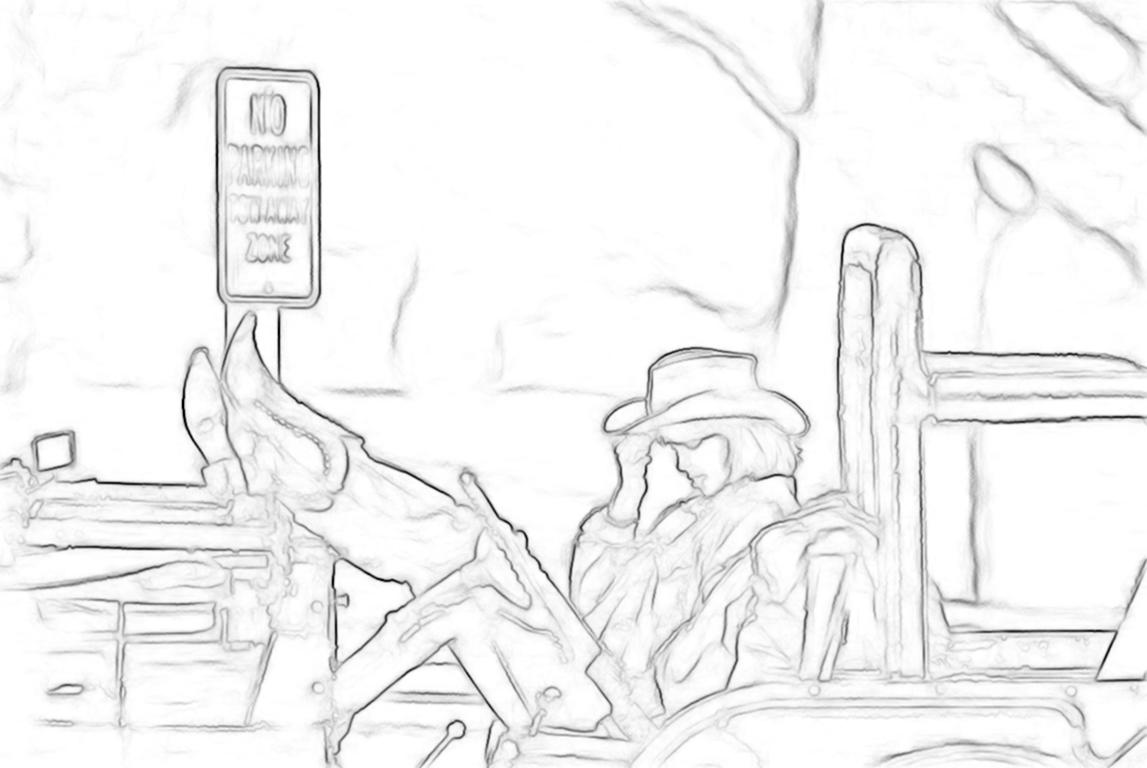} & 
\includegraphics[width = .32\linewidth]{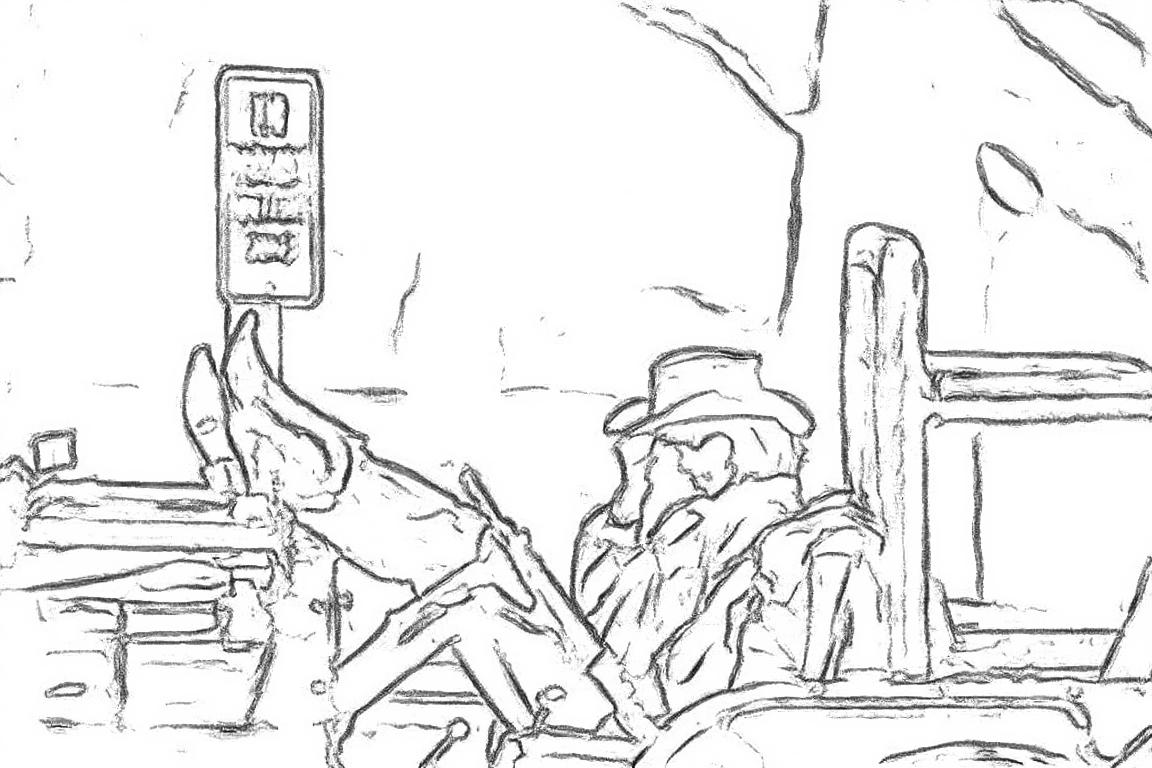} & \\

\includegraphics[width = .32\linewidth]{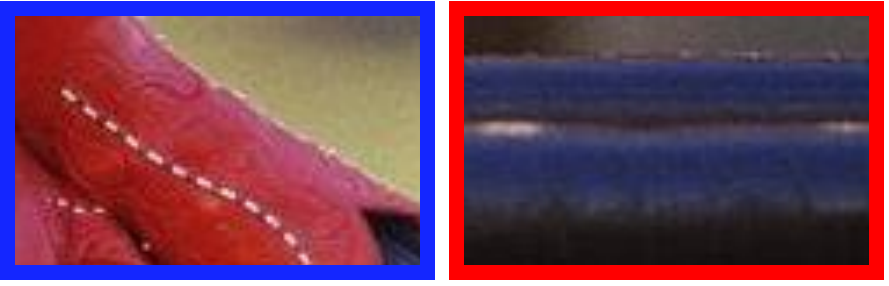} & 
\includegraphics[width = .32\linewidth]{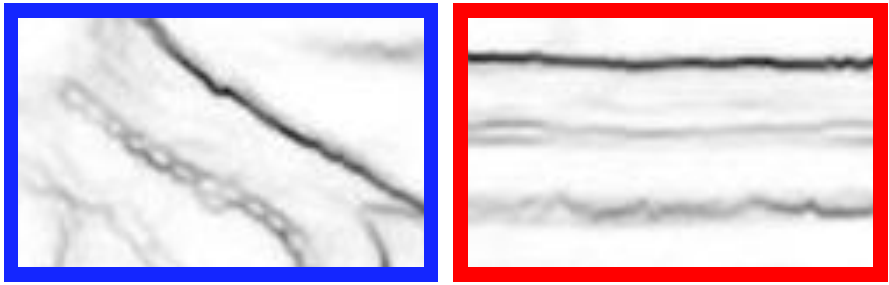} & 
\includegraphics[width = .32\linewidth]{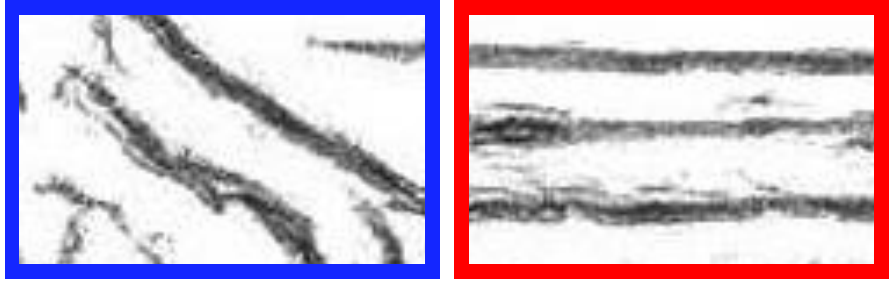} & \\

{(a) Input} & {(b) Boundary~\cite{DollarICCV13edges}} & { (c) Clean: base} &\\

\includegraphics[width = .32\linewidth]{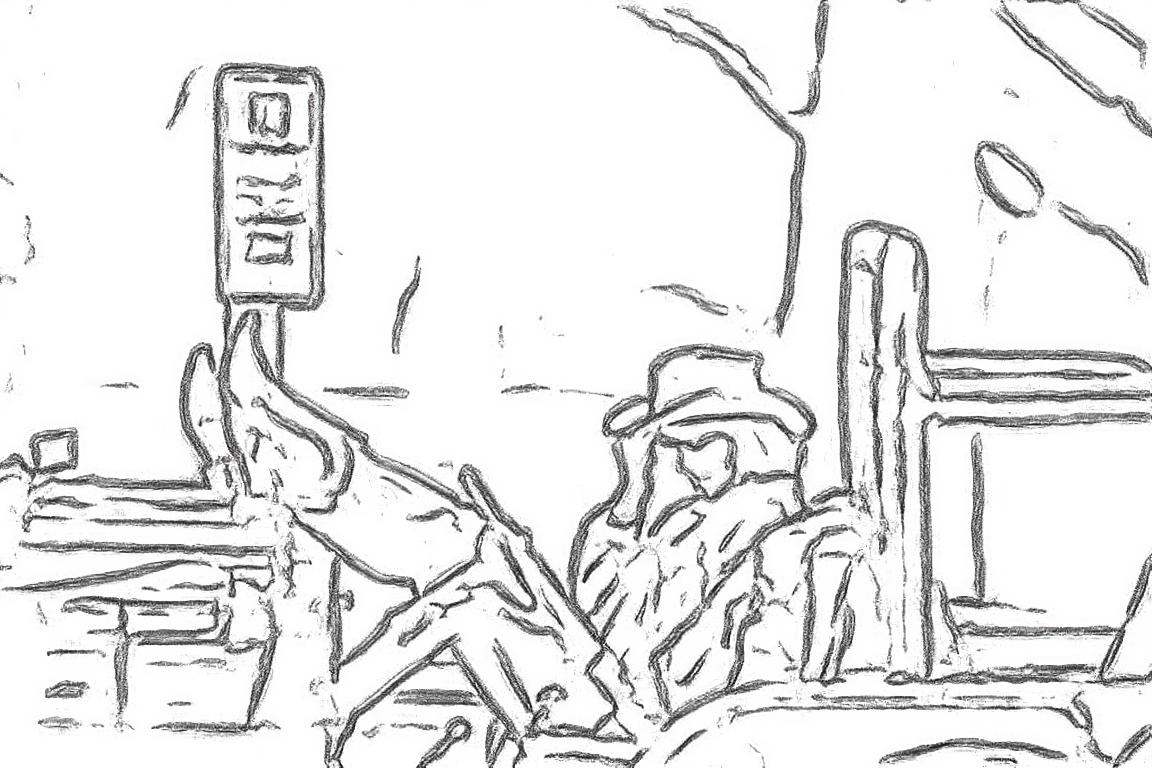} & 
\includegraphics[width = .32\linewidth]{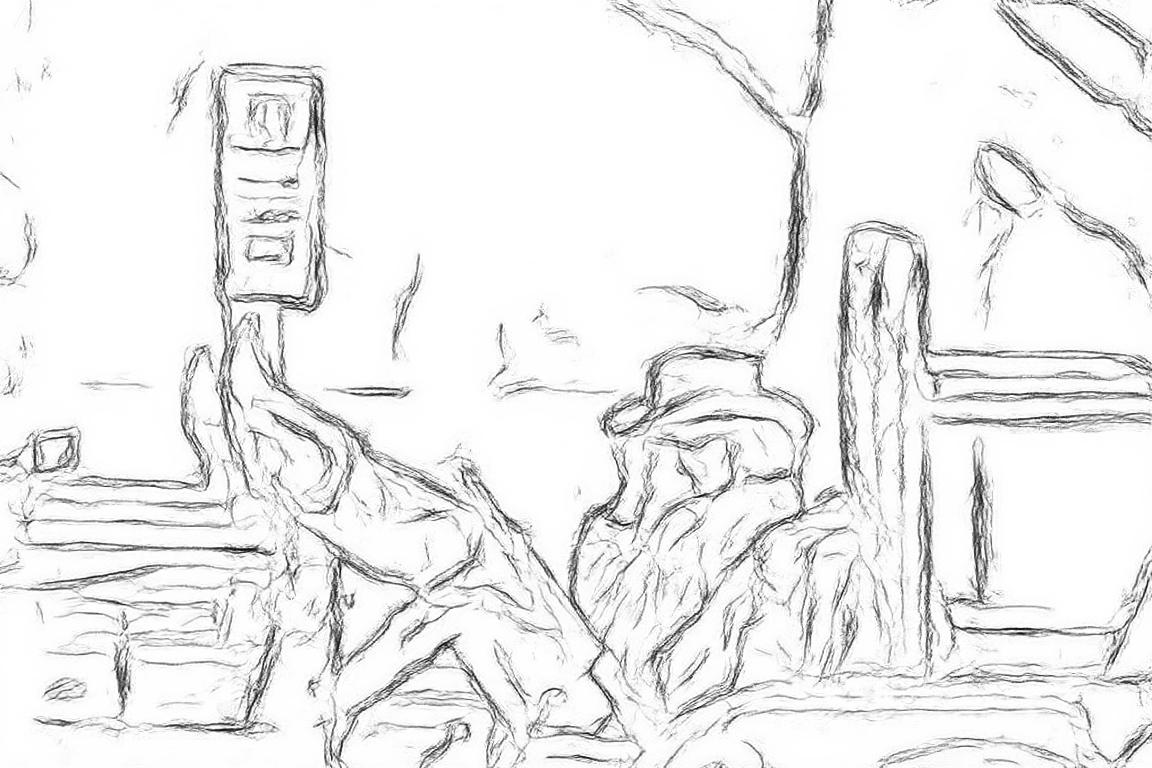} & 
\includegraphics[width = .32\linewidth]{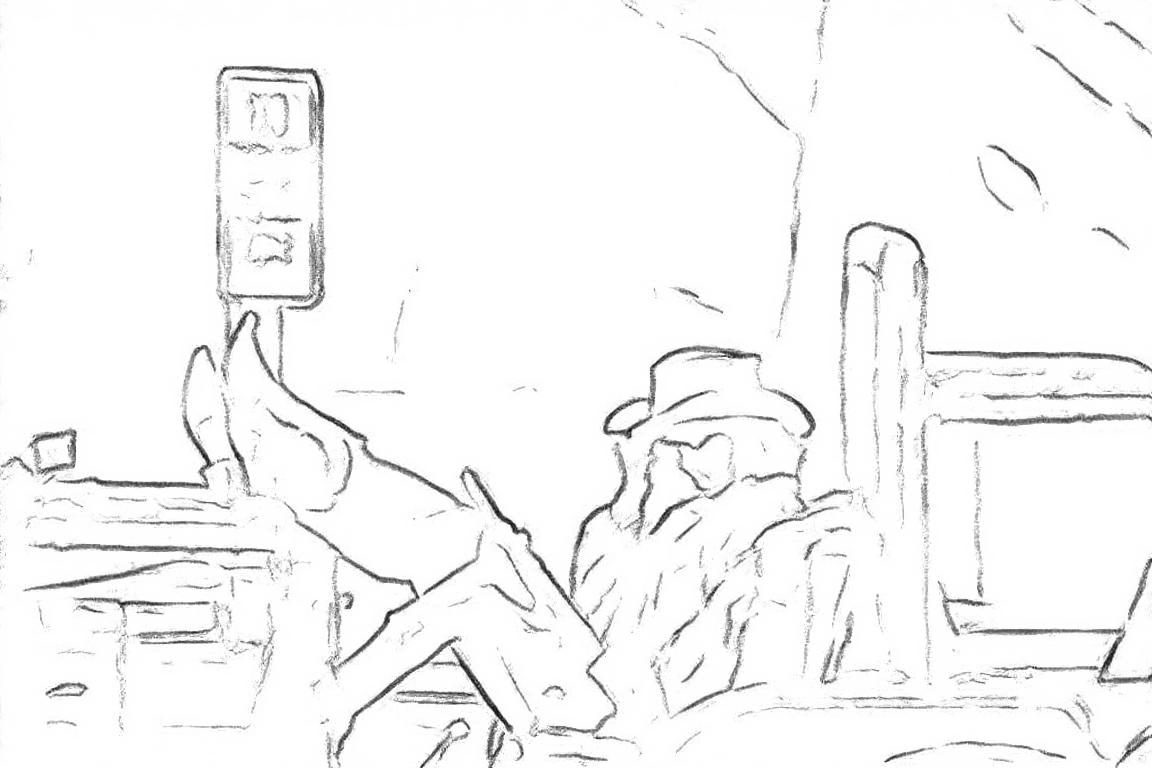} & \\

\includegraphics[width = .32\linewidth]{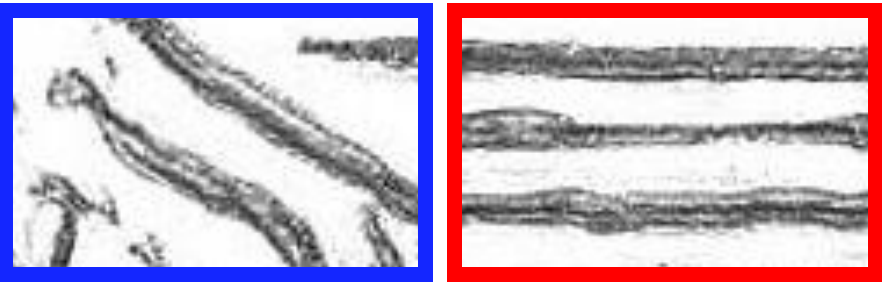} & 
\includegraphics[width = .32\linewidth]{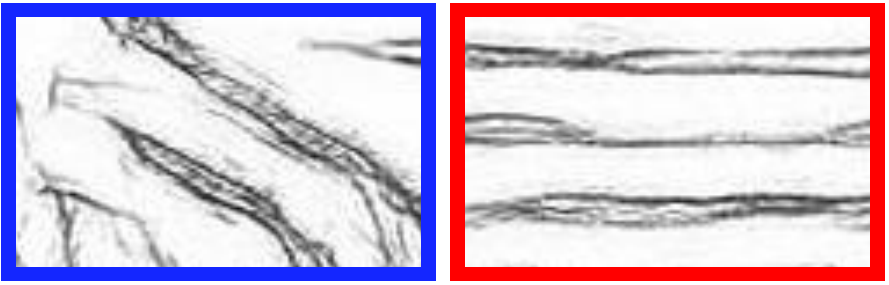} & 
\includegraphics[width = .32\linewidth]{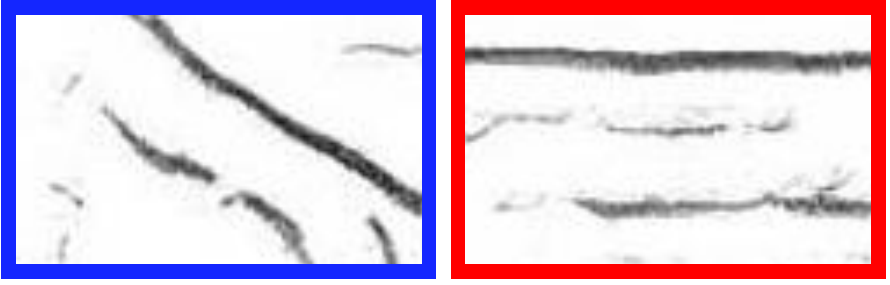} & \\

{(d) Clean: $\sigma$=3} & {(e) Rough: $\sigma$=3} & {(f) Clean: $\tau$=0.97} &\\

\end{tabular}
\vspace{0.05em}
\caption{Outline results for a highly-textured photo.
(b) is the boundary map of input (a), which is then filtered by the XDoG with different parameters.
We set $\sigma=2.0$,$\tau=0.99$, $k=1.6$, $\epsilon=0.1$, $\varphi=200$ in XDoG and show the base pencil outlines in (c). 
(d)-(f) show the results by adjusting one parameter while keeping others fixed.
}
\label{fig:edge_result}
\end{figure}

\begin{figure}[t]
\centering
\begin{tabular}{c@{\hspace{0.005\linewidth}}c@{\hspace{0.005\linewidth}}c@{\hspace{0.005\linewidth}}c@{\hspace{0.005\linewidth}}c@{\hspace{0.005\linewidth}}c}

\includegraphics[width = .32\linewidth]{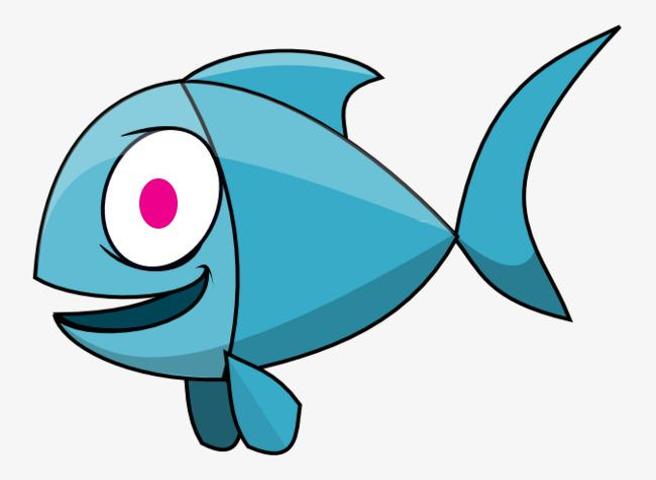} & 
\includegraphics[width = .32\linewidth]{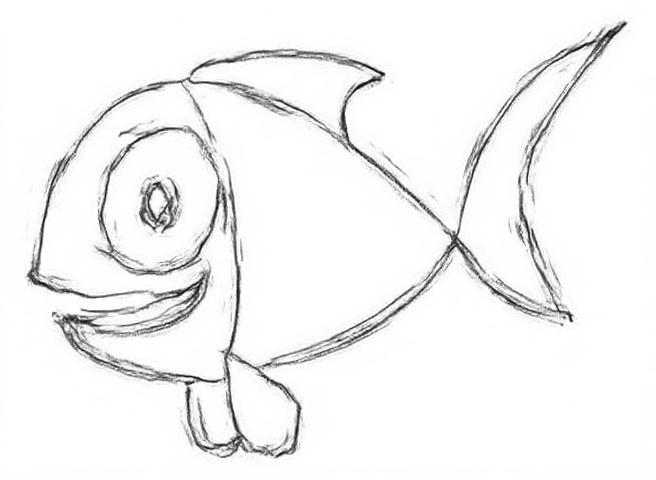} &
\includegraphics[width = .32\linewidth]{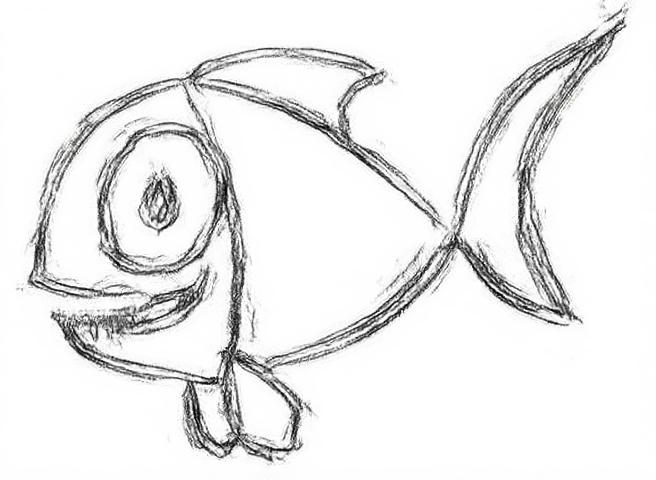} & \\

{(a) Input} & {(b) Rough: base} & { (c) Rough: $\sigma$=4.5} &\\

\includegraphics[width = .32\linewidth]{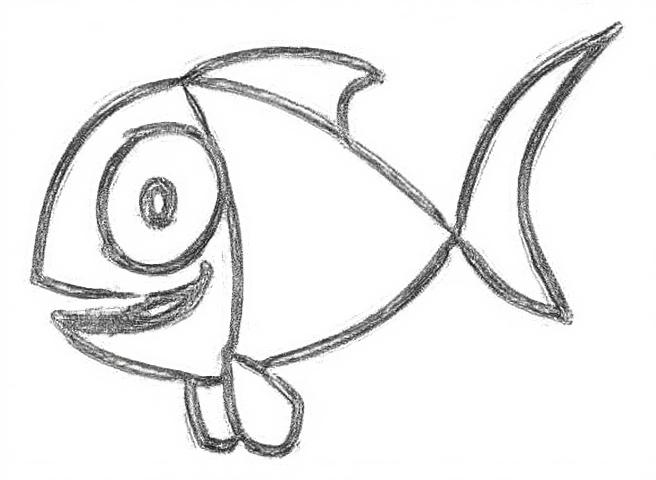} & 
\includegraphics[width = .32\linewidth]{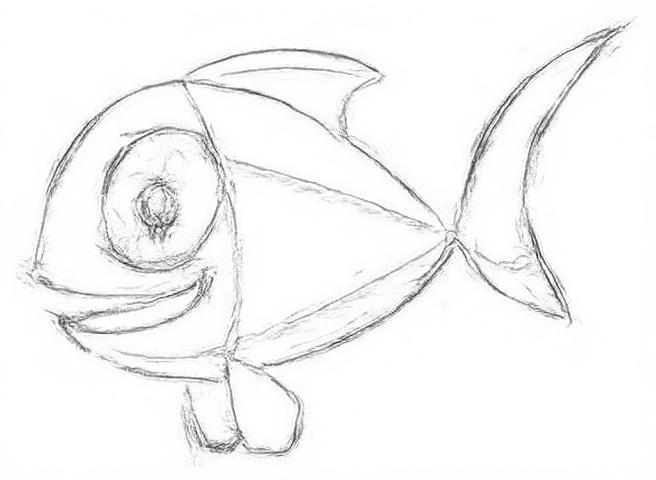} & 
\includegraphics[width = .32\linewidth]{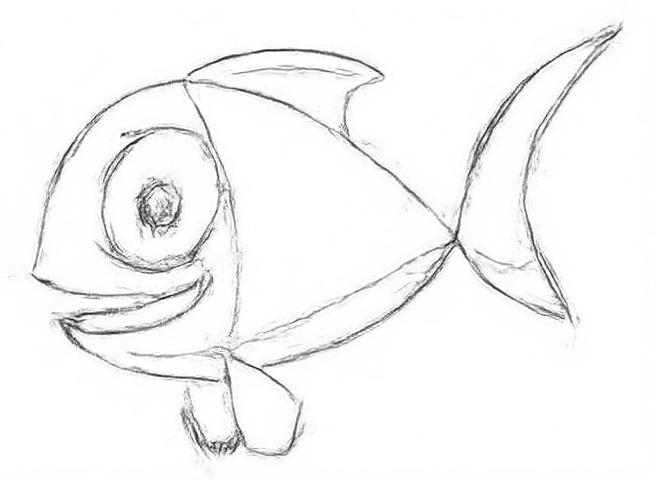} & \\

{(d) Clean: $\sigma$=4.5} & 
{(e) Rough: $\tau$=0.98} & {(f) Rough: $\epsilon$=1.6 } &\\

\end{tabular}
\vspace{0.05em}
\caption{Outline results for a simple cartoon image. The input in (a) is directly filtered by the XDoG.
We set $\sigma=2.5$,$\tau=0.96$, $k=1.6$, $\epsilon=0.1$, $\varphi=200$ and show the base pencil outlines in (b).
(c)-(f) show diverse outline results by controlling parameters.
}
\label{fig:edge_result_fish}
\end{figure}

\begin{figure*}[t]
\centering
\begin{tabular}{c@{\hspace{0.005\linewidth}}c@{\hspace{0.005\linewidth}}c@{\hspace{0.005\linewidth}}c@{\hspace{0.005\linewidth}}c@{\hspace{0.005\linewidth}}c}

\includegraphics[width = .19\linewidth]{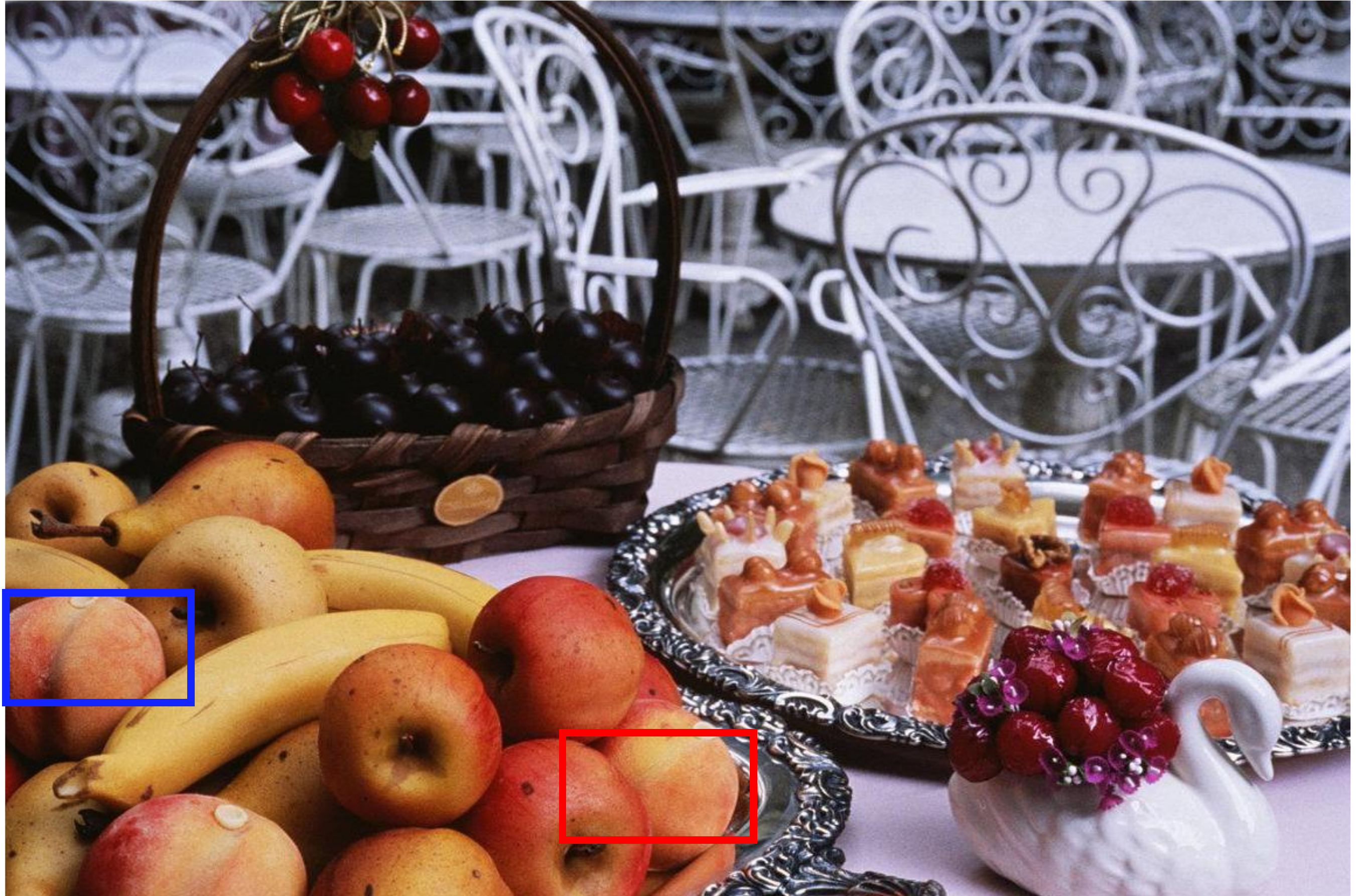} & 
\includegraphics[width = .19\linewidth]{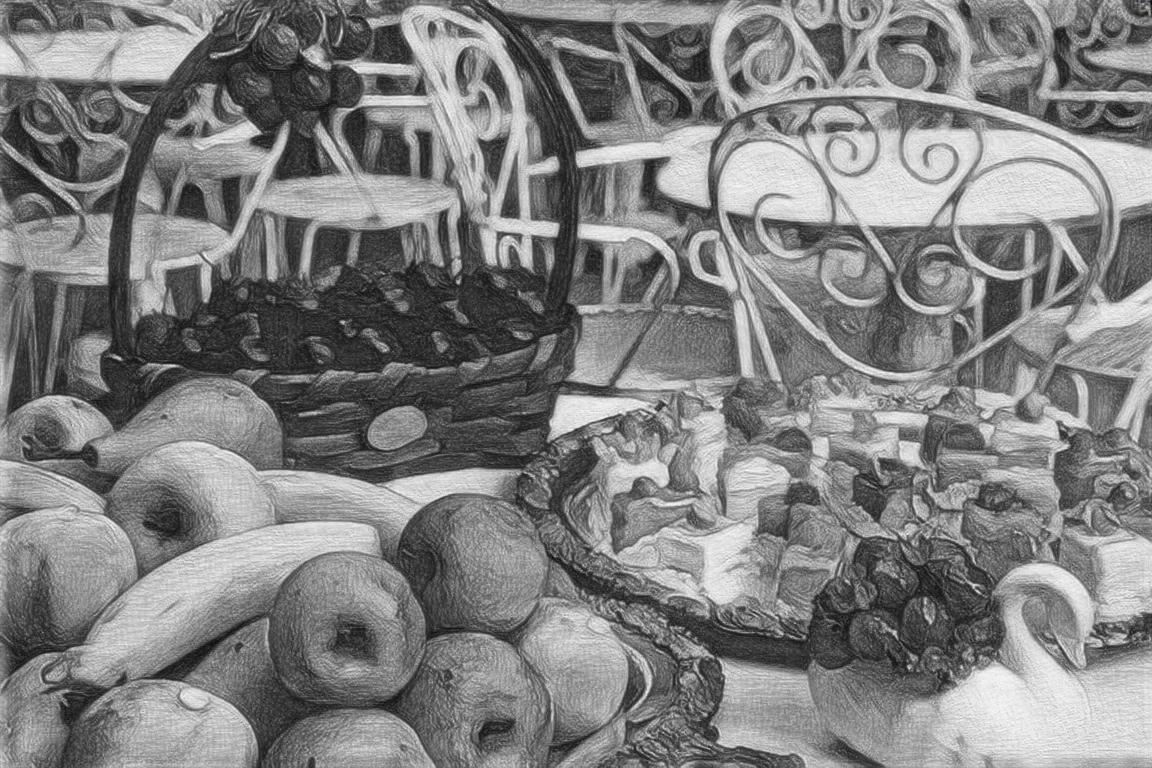} & 
\includegraphics[width = .19\linewidth]{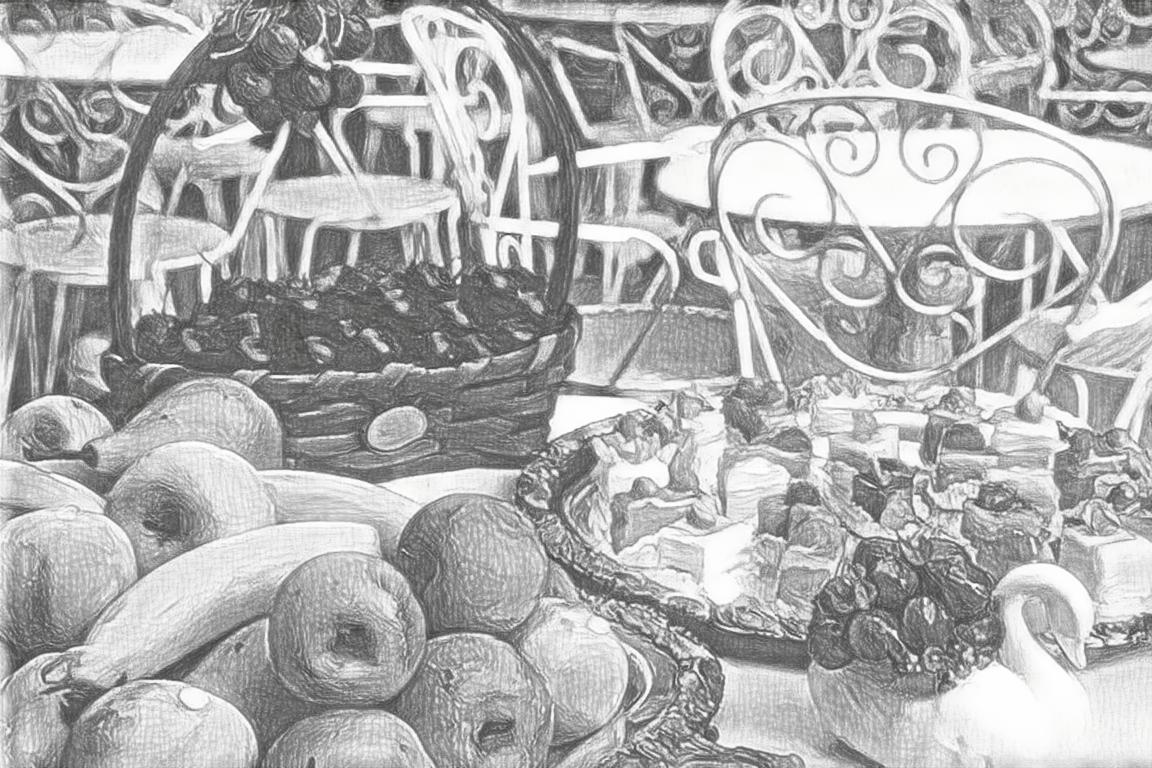} & 
\includegraphics[width = .19\linewidth]{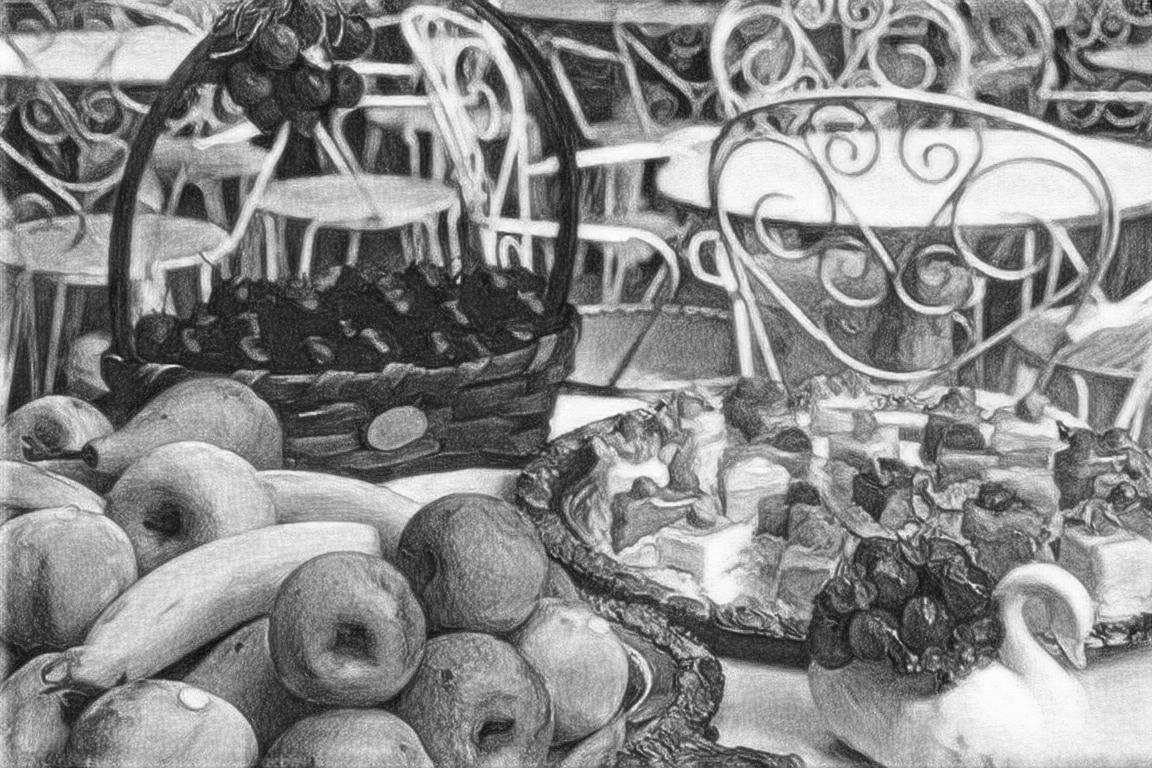} & 
\includegraphics[width = .19\linewidth]{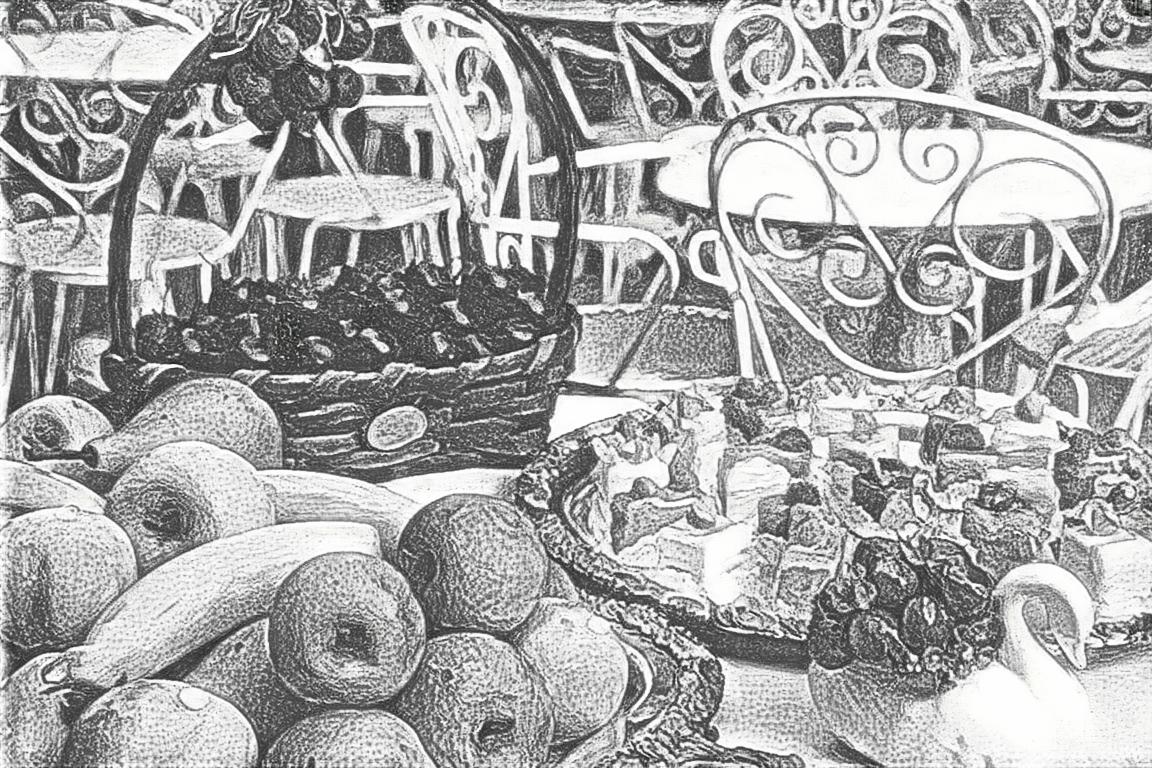} & \\

\includegraphics[width = .19\linewidth]{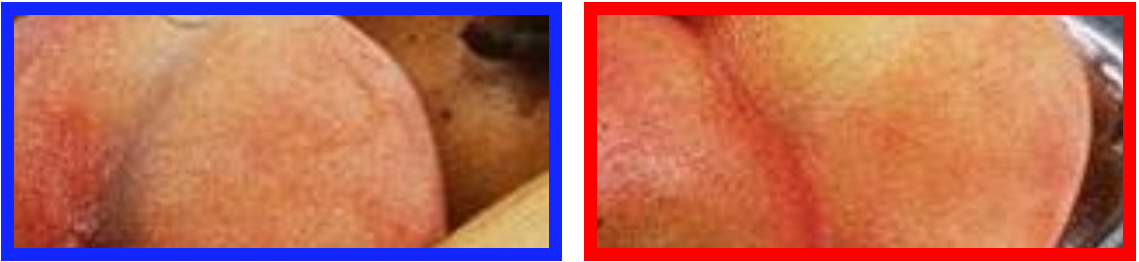} & 
\includegraphics[width = .19\linewidth]{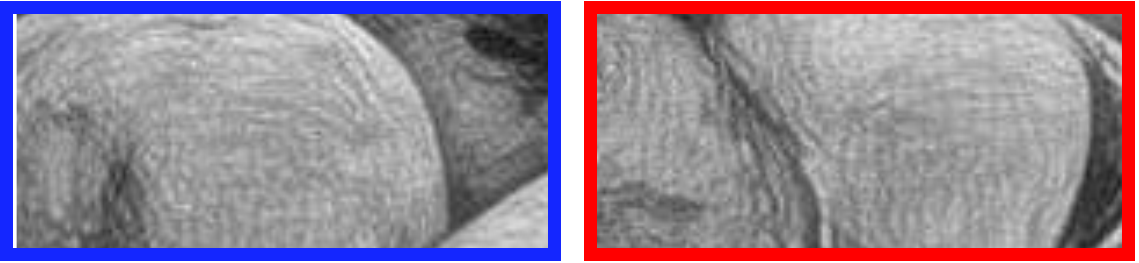} & 
\includegraphics[width = .19\linewidth]{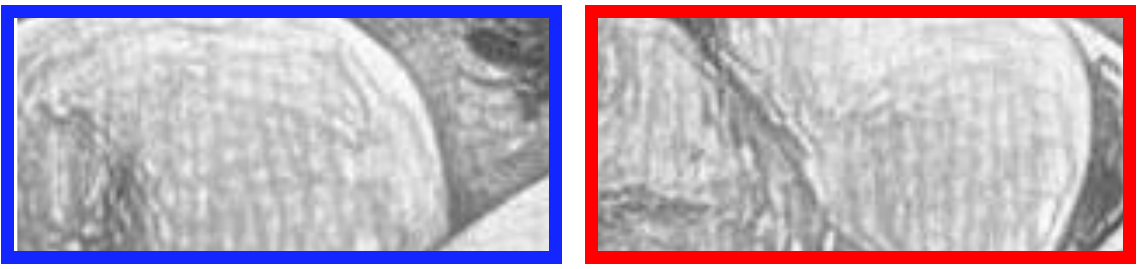} & 
\includegraphics[width = .19\linewidth]{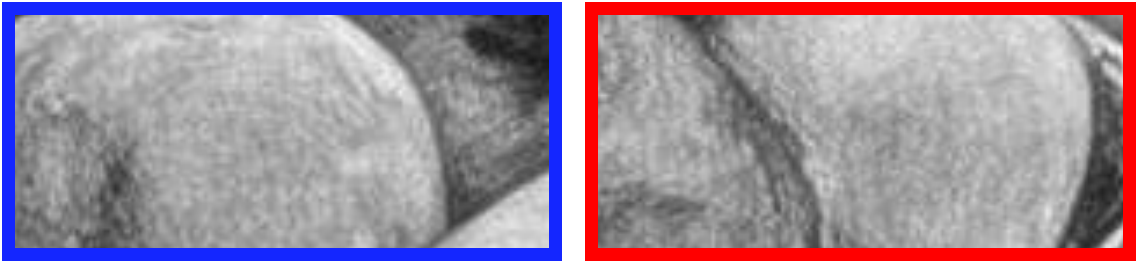} &
\includegraphics[width = .19\linewidth]{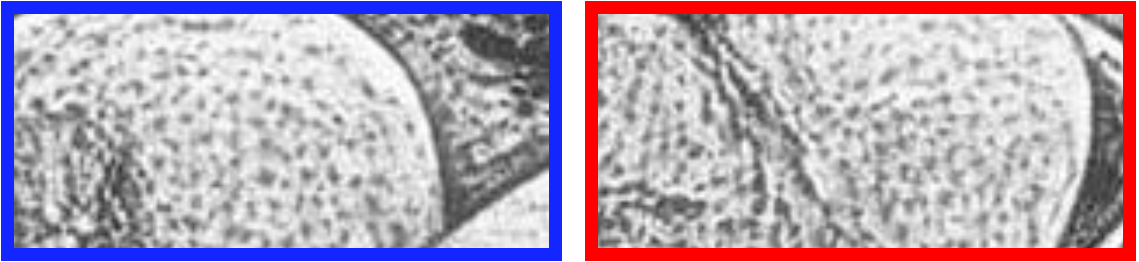} &\\

\includegraphics[width = .19\linewidth]{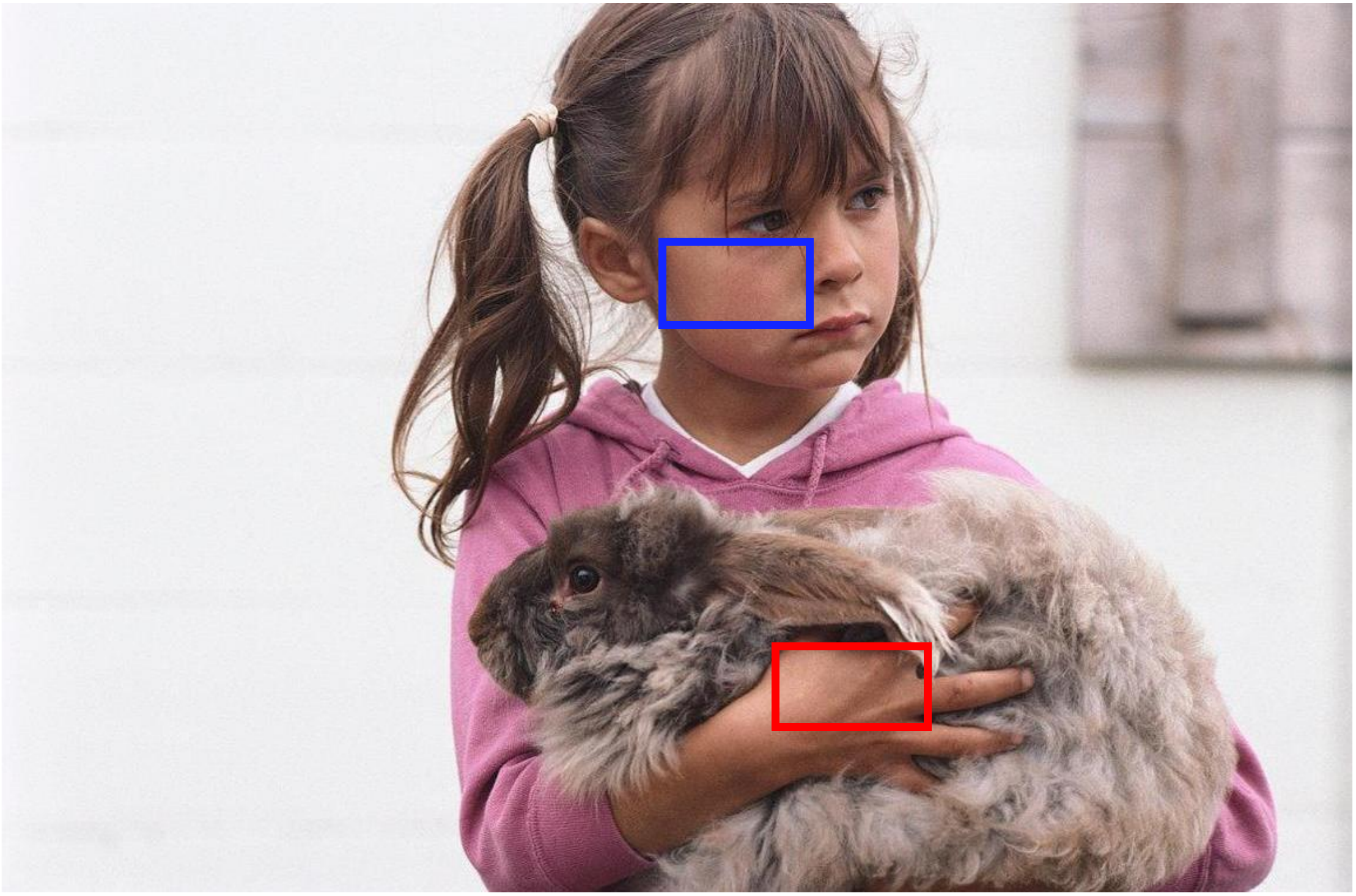} & 
\includegraphics[width = .19\linewidth]{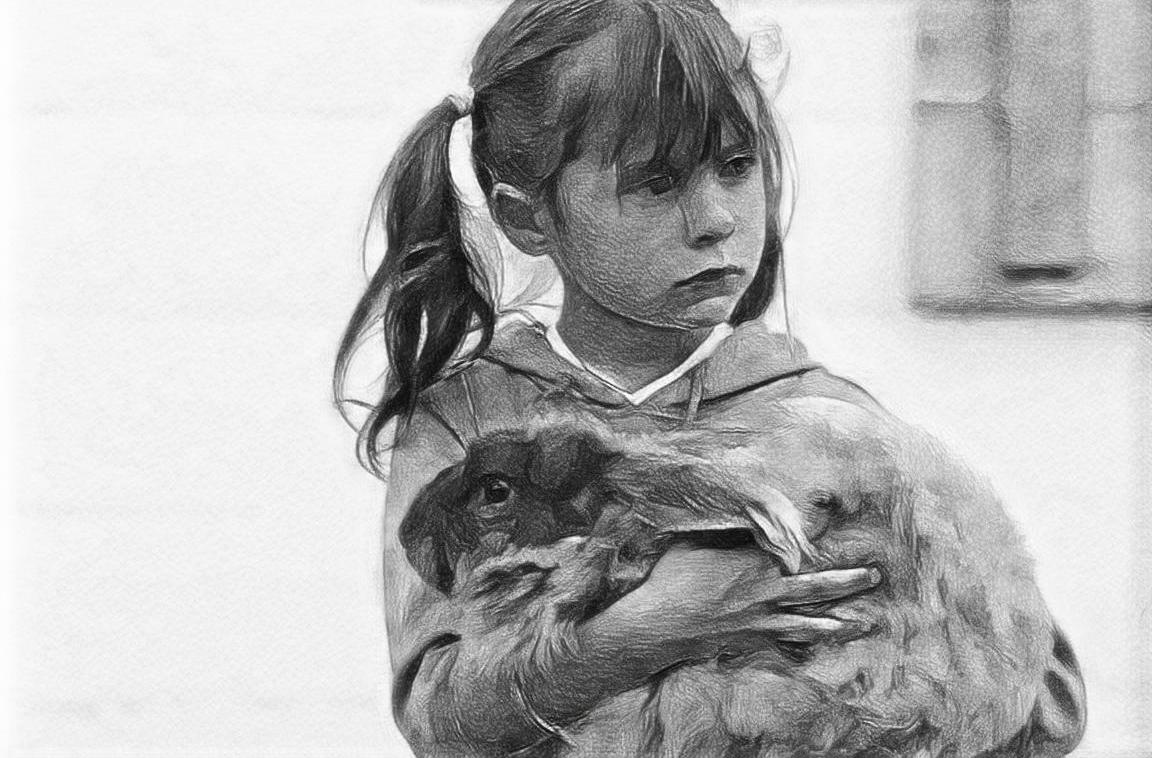} &
\includegraphics[width = .19\linewidth]{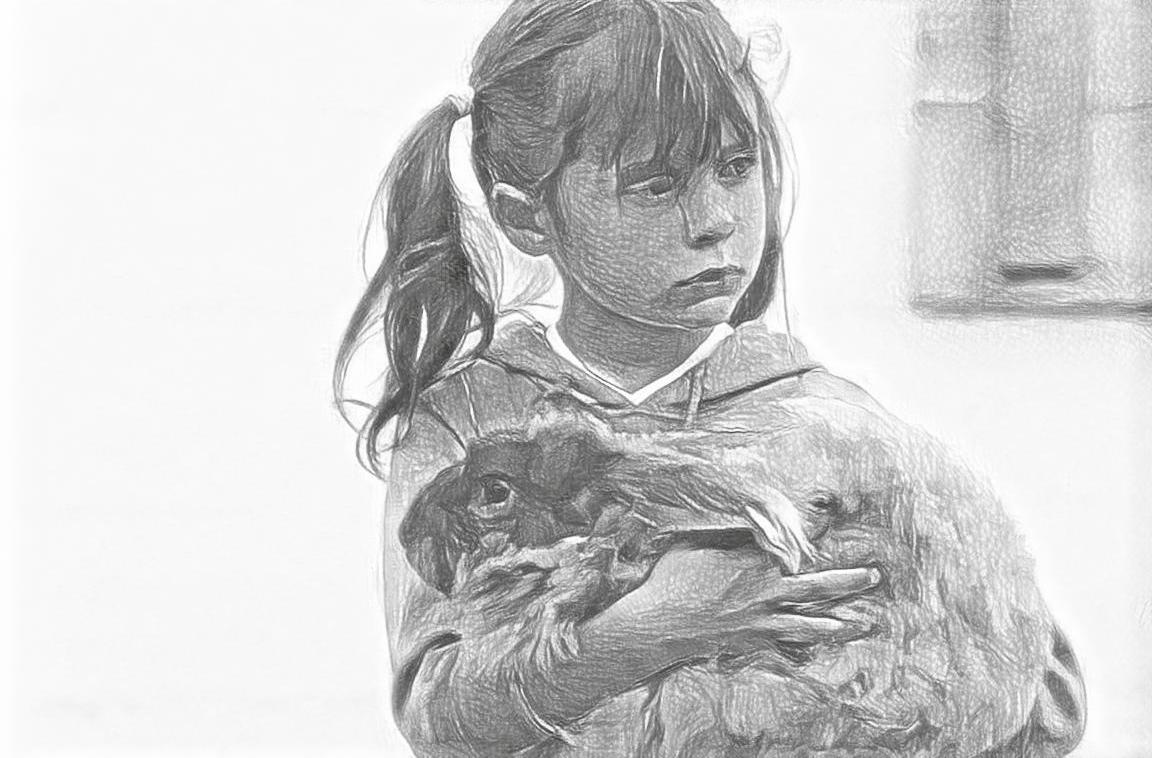} &
\includegraphics[width = .19\linewidth]{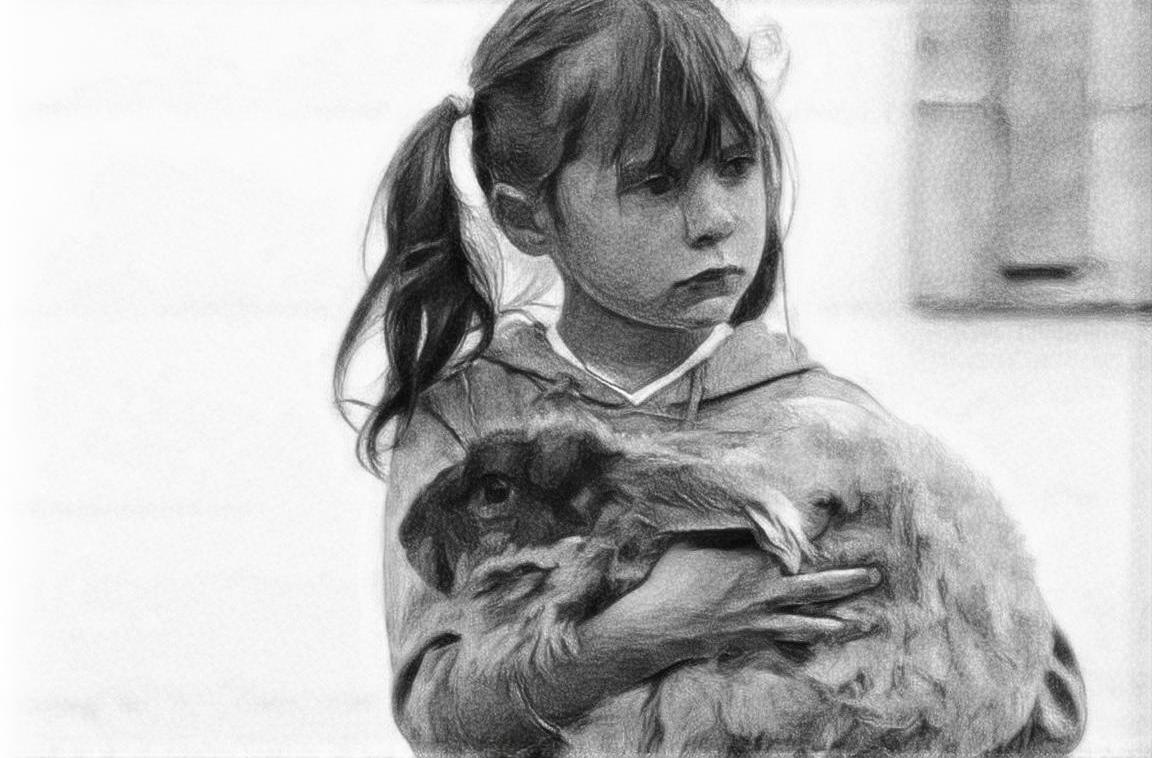} &
\includegraphics[width = .19\linewidth]{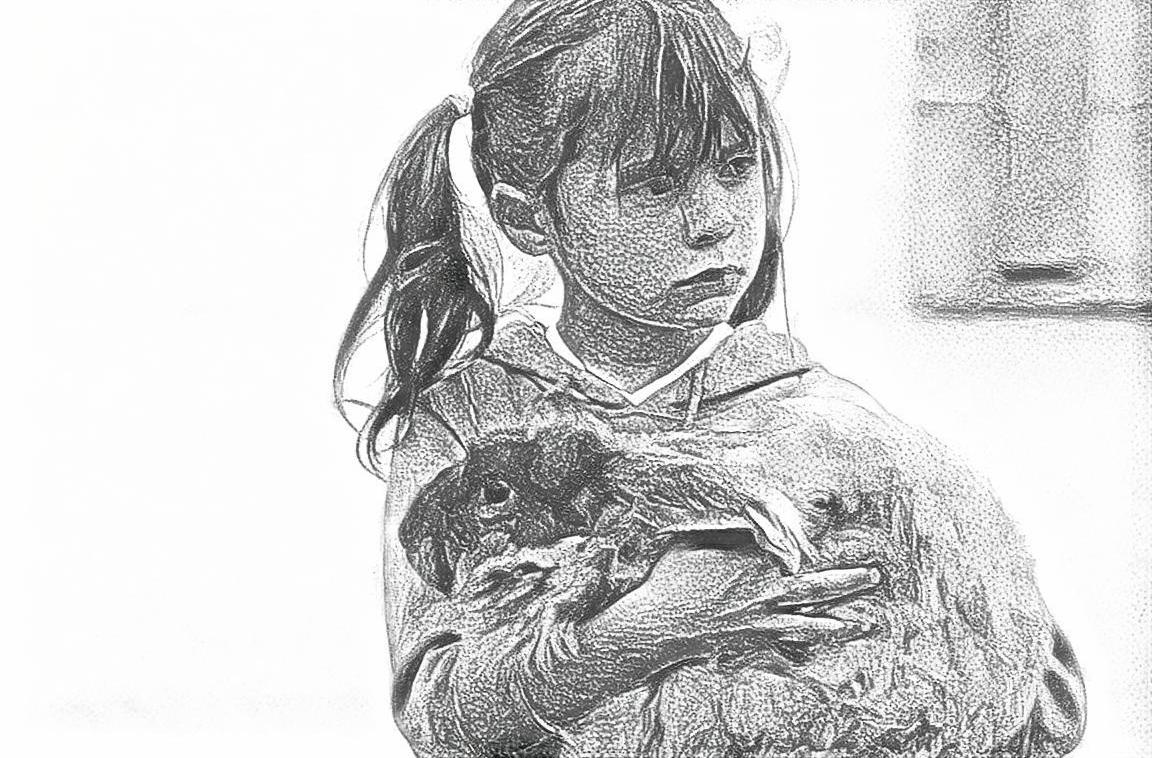} & \\

\includegraphics[width = .19\linewidth]{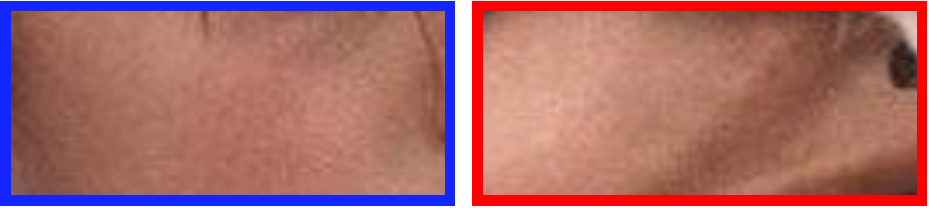} & 
\includegraphics[width = .19\linewidth]{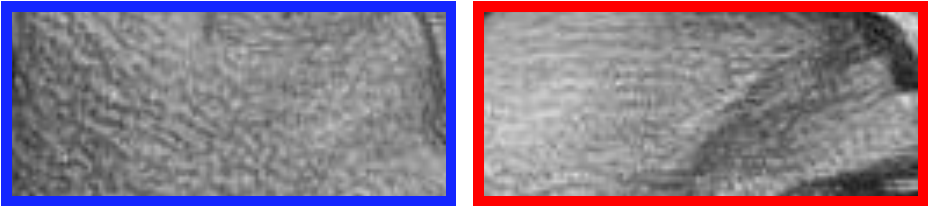} &
\includegraphics[width = .19\linewidth]{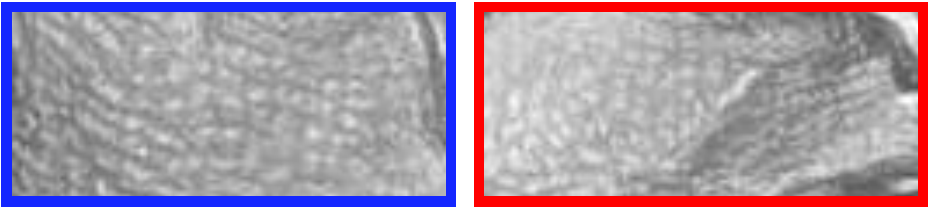} &
\includegraphics[width = .19\linewidth]{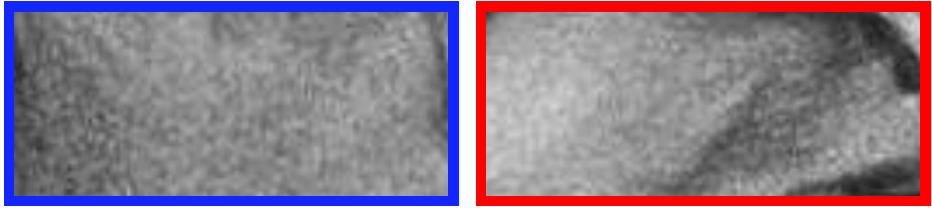} &
\includegraphics[width = .19\linewidth]{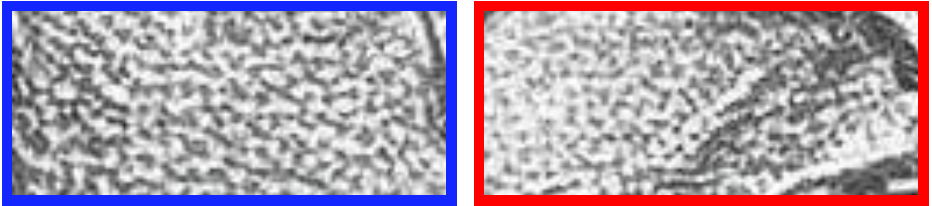} & \\

{Input} & {Hatching} & {Crosshatching} & {Blending} & {Stippling} &\\

\end{tabular}
\vspace{0.05em}
\caption{Four types of shading results of the proposed algorithm by switching bits in the selection unit (\textbf{zoom in} for details).}
\label{fig:shading_result}
\end{figure*}

\vspace{.5em}
\noindent\textbf{User study.}~We resort to user studies for the quantitative evaluation of~\cite{CycleGAN-ICCV17,lu-NPAR2012} and our method as pencil drawing synthesis is originally a highly subjective task.
The method of Gatys \etal~\cite{GatysTransfer-CVPR2016} is not included in the user study because it was clearly inferior to the others in our early experiments (Figure~\ref{fig:comparison}(b)). %transfers the specific style of a pencil example which does not quite respect the tone (or contrast) in the target image, as shown in Figure~\ref{fig:comparison}(b).
We use 30 natural images provided in~\cite{lu-NPAR2012} and randomly select 15 images for each subject.
We display the results by all three methods side-by-side in random order and ask each subject to vote for one result that looks the most like a pencil drawing. 
We finally collect the feedback from 50 subjects of totally 750 votes and show the percentage of votes each method received in the top row of Table~\ref{table:user_study}. 
The study shows that our method receives the most votes for better pencil effects, nearly seven times as much as those of other two methods.

Lu \etal~\cite{lu-NPAR2012} observed that pencil drawings often exhibit global tonal changes from the input, and described a histogram-based tone adjustment step to model this observation. 
In order to fairly compare with this step, we perform a second user study where the input is preprocessed by this step. 
The user study results with tone adjustment are shown in the bottom row of Table~\ref{table:user_study}. Again, our method obtains substantially more votes than the previous methods. 
We show our results with and without tone adjustment in Figure~\ref{fig:tone}(c) and (d) as well as the corresponding result of Lu \etal\cite{lu-NPAR2012} in (b).
The tone adjustment step provides an additional user control for our method as well.

\subsection{User control}
\label{use_control}

Our translation model provides fine-grained control over different pencil drawing styles.
Figure~\ref{fig:edge_result} and~\ref{fig:edge_result_fish} show that users could either switch between clean and rough outline style through the selection unit, or adjust parameters in XDoG to obtain different outline results.
The photo example in Figure~\ref{fig:edge_result}(a) is highly-textured in the clothes and background, so we first use the boundary detector~\cite{DollarICCV13edges} to detect its boundary map, which is then filtered by the XDoG.
As the most sensitive and important parameter in XDoG, the $\sigma$ defines the line thickness and sketchiness. 
Generally, when a clean style is selected, increasing the value of $\sigma$ leads to thicker lines (Figure~\ref{fig:edge_result}(c)-(d)).
When a rough style is selected, increasing the value of $\sigma$ results in increase in repetitive lines (Figure~\ref{fig:edge_result}(e)).
%
%Since the standard deviation of the second Gaussian filter is defined as $k\cdot \sigma$, adjusting $k$ could also change the line sketchiness similar to adjusting $\sigma$.
%
In addition, by adjusting other parameters (e.g., $\tau$), the XDoG filter is able to control the sensitivity on detected edges, which allows users to draw both strong and weak edges (Figure~\ref{fig:edge_result}(f)).
In Figure~\ref{fig:edge_result_fish}, we show the outline results for a simple cartoon image without heavy textures.
Figure~\ref{fig:shading_result} shows shading results of two examples, i.e., a still life and a portrait -- two popular reference choices for a pencil drawing.
By controlling the selection unit, users get results in different shading styles.  

\begin{figure}[t]
\centering
\begin{tabular}{c@{\hspace{0.005\linewidth}}c@{\hspace{0.005\linewidth}}c@{\hspace{0.005\linewidth}}c}

\includegraphics[width = .32\linewidth]{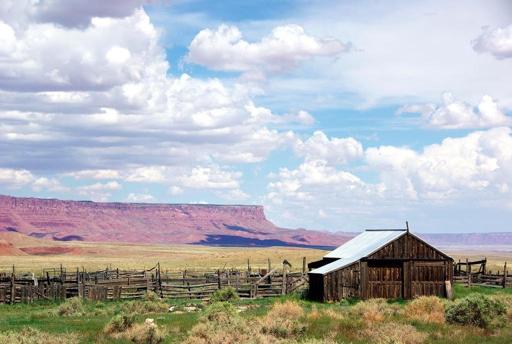} &
%\hspace{1pt}\vrule\hspace{1pt}
\includegraphics[width = .32\linewidth]{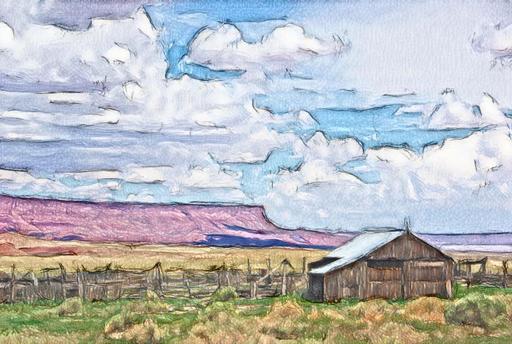} &
\includegraphics[width = .32\linewidth]{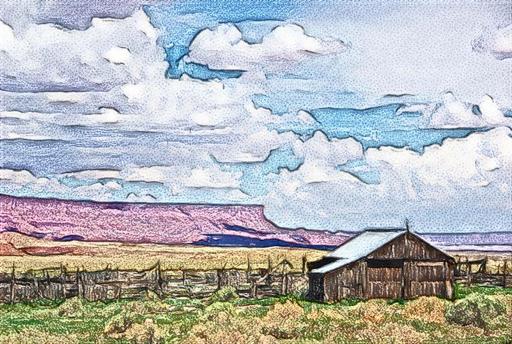} & \\

\includegraphics[width = .32\linewidth]{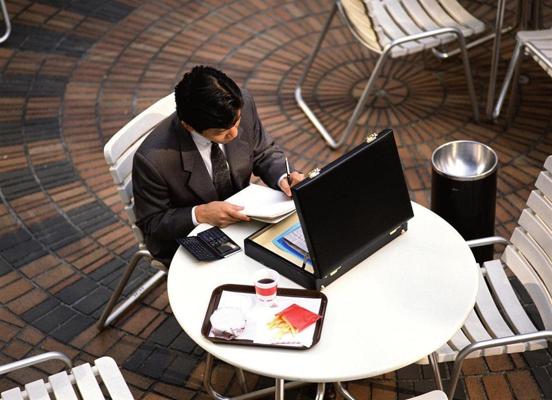} &
%\hspace{1pt}\vrule\hspace{1pt}
\includegraphics[width = .32\linewidth]{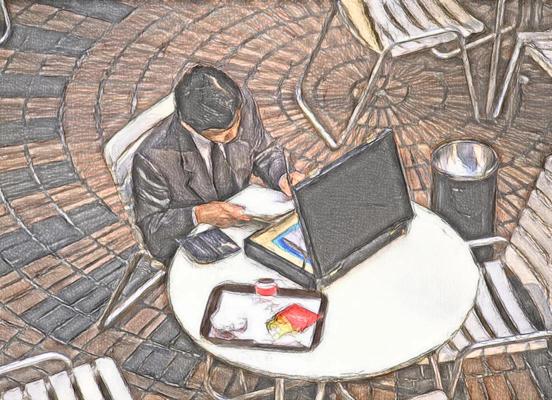} &
\includegraphics[width = .32\linewidth]{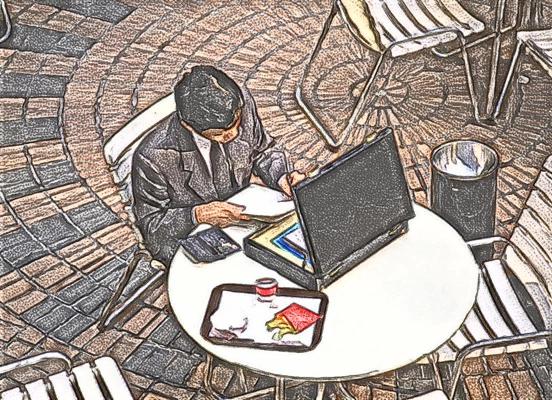} & \\

%\includegraphics[width = .32\linewidth]{figs/colorpencil/16--56.jpg} &
%\hspace{1pt}\vrule\hspace{1pt}
%\includegraphics[width = .32\linewidth]{figs/colorpencil/16--56_e0_s1_combo_color.jpg} &
%\includegraphics[width = .32\linewidth]{figs/colorpencil/16--56_e1_s2_combo_color.jpg} & \\

{(a) Input} & {(b) Ours: $L_1+S_2$} &  {(c) Ours: $L_2+S_4$} &\\

\end{tabular}
\vspace{0.05em}
\caption{Extension of our algorithm to color pencils in different outline and shading styles (\textbf{zoom in} for fine details).
Pencil outlines of all examples are generated by applying the XDoG and learned model on their boundary maps.}
\label{fig:colorpencil}
\end{figure}

\vspace{.5em}
\noindent\textbf{Color pencil drawings.}~The extension of our algorithm to color pencil drawing is quite straightforward, following the method of \cite{Hertzmann-2001-IA,Gatys2016-control}. 
We first convert the input image from $RGB$ to $LAB$ color space, then replace the $L$ channel with that of our generated gray-scale pencil drawing result, and finally map back to $RGB$ space.
Figure~\ref{fig:colorpencil} shows two color pencil results in different outline and shading styles.

\section{Conclusions}

In this work, we propose a photo-to-pencil translation method with flexible control over different drawing styles.
We design a two-branch network that learns separate filters for outline and shading generation respectively. 
To facilitate the network training, we introduce filtering/abstraction techniques into deep models that avoid the heavy burden of collecting paired data.
Our model enables multi-style synthesis in a single network to produce diverse results.
We demonstrate the effectiveness and flexibility of the proposed algorithm on different pencil outline and shading styles.

\vspace{-0.5em}
\paragraph{\bf Acknowledgment.} This work is supported in part by the NSF CAREER Grant \#1149783, and gifts from Adobe, Verisk, and NEC. YJL is supported by Adobe
and Snap Inc. Research Fellowship.

\clearpage

%\newpage
{\small

%\bibliography{reference,contour_tutorial}
}

\end{document}